\documentclass[11pt,letterpaper]{article}
\usepackage[margin=1in]{geometry}
\usepackage[T1]{fontenc}
\usepackage{times}
\usepackage[authoryear,round]{natbib}
\setcitestyle{citesep={;},aysep={,},yysep={;}}
\usepackage{amsmath,amssymb,amsthm,mathtools,bm,mathrsfs,booktabs}
\usepackage{graphicx,longtable,float,needspace}
\usepackage{hyperref}
\hypersetup{
  hidelinks,
  pdftitle={Semiparametric Efficiency for Entropic Optimal Transport under Structured Distribution Shift},
  pdfauthor={Yiming Ma, Weiwei Zhuang, Hang Liu}
}

\newtheorem{assumption}{Assumption}
\newtheorem{theorem}{Theorem}
\newtheorem{proposition}{Proposition}
\newtheorem{lemma}{Lemma}
\newtheorem{corollary}{Corollary}
\newcommand{\appref}[1]{\hyperref[app:#1]{Appendix~\ref*{app:#1}}}

\title{Semiparametric Efficiency for Entropic Optimal Transport
under Structured Distribution Shift}
\author{
  Yiming Ma\\[-0.1em]{\small\nolinkurl{mayiming@mail.ustc.edu.cn}}
  \and Weiwei Zhuang\\[-0.1em]{\small\nolinkurl{weizh@ustc.edu.cn}}
  \and Hang Liu\\[-0.1em]{\small\nolinkurl{hliu01@ustc.edu.cn}}
}
\date{}

\begin{document}
\raggedbottom
\maketitle
\begin{abstract}

Can structural information improve estimation of entropic optimal transport (EOT), and when is the gain zero? We study a fixed-dimensional exponential density ratio model (DRM) with a nonparametric baseline, compact support, and fixed positive regularization. At positive limiting sampling fractions, an exact projection identity quantifies the variance gain from the restriction. The pooled empirical-likelihood estimator attains the restricted bound and supports asymptotically valid confidence intervals when the bound is positive. Even under correct specification, the gain can be zero and can shrink as the tilt space expands. The same projection residual determines direction-specific first-order bias under local misspecification. When both samples grow and the target fraction vanishes, a zero first-order tilt derivative can leave source fluctuations and quadratic tilt error as the leading terms. Their balance gives a phase transition in the limiting distribution. A continuous example gives different convergence rates for two correctly specified bases at the same distribution pair. Controlled experiments assess the fixed-proportion theory. Natural-data comparisons distinguish variance reductions from improvements in reference RMSE driven by smaller reference bias.

\end{abstract}

\section{Introduction}
Entropic optimal transport (EOT) compares distributions through the geometry of their observations. The standard empirical estimator uses each sample to estimate its own marginal. A structural link lets both samples inform each marginal. The precision gain depends on which changes the link rules out and how sensitive the EOT cost is to those changes.

Let $F_0,F_1$ be the source and target distributions. The density ratio model (DRM) sets $r=dF_1/dF_0=\exp\{\theta_0^\top Q\}$ for a fixed vector $Q$ of basis functions that includes an intercept. The baseline $F_0$ is nonparametric. On compact support with fixed positive regularization, we ask how much precision the restriction makes possible and whether a pooled estimator can attain it.

First consider sample sizes $n_0,n_1$ with $n=n_0+n_1$ and sampling fractions $\rho_{gn}=n_g/n\to\rho_g\in(0,1)$ for $g=0,1$. The centered dual potential $\phi_g$ gives the first-order sensitivity of EOT to changes in $F_g$. Write $E_g$ for expectation under $F_g$ and define the contrast $\zeta=\phi_1-(\rho_1/\rho_0)\phi_0$. Project $\zeta$ onto the span of $Q$ in the weighted squared norm $E_1(\cdot^2/\omega)$, where $\omega=\rho_0+\rho_1r$, and call the residual $e$. For the $\sqrt n$-scaled error, $V_{\rm NP}$ is the efficiency bound in the unrestricted nonparametric (NP) model and $V_{\rm DRM}$ is the bound under the DRM. The exact gain is
\begin{equation}\label{eq:intro-gain}
 V_{\rm NP}-V_{\rm DRM}=\frac{\rho_0}{\rho_1}E_1(e^2/\omega).
\end{equation}
The gain is strict precisely when the DRM excludes some first-order variation in EOT, as measured by a nonzero residual. Enlarging a correctly specified tilt space admits more directions of change and can reduce the gain. The pooled empirical-likelihood estimator DRM-EOT attains the restricted bound.

When both sample sizes grow and the target fraction vanishes, the tilt directions can also affect the rate. If the first-order tilt derivative is zero and both the source variance and tilt curvature are nonzero, source fluctuations and quadratic tilt error compete. A continuous example in Section~\ref{sec:rare-target} gives different rates for two correctly specified bases at the same distribution pair.

At positive limiting fractions, the analysis gives feasible inference for the scalar cost and uses the projection residual to quantify bias along a local departure. Controlled synthetic and exact-model OfficeHome experiments assess the efficiency predictions. Natural OfficeHome and annual cash dividend proposal groups show how variance and reference root mean squared error (RMSE) change when the DRM serves as a working model.

\section{Related Work}
\paragraph{Regularized OT and its statistical properties.}
Entropic regularization enables transport computation by matrix scaling \citep{cuturi2013sinkhorn}. Sinkhorn divergences debias regularized transport and connect geometric and kernel comparisons \citep{feydy2019sinkhorn}. \citet{genevay2019sample} study their sample complexity. \citet{mena2019statistical} give bounds and a central limit theorem around expected empirical EOT under sub-Gaussian marginals. \citet{goldfeld2024regularized} derive differentiability, efficiency, and bootstrap results for regularized transport. Their delta method also applies to fitted marginals with the required process limit. We use the marginal derivative and stability of transport potentials \citep{nutz2023stability} to compute how the DRM changes the bound. \citet{goldfeld2024limit} establish second-order Hadamard differentiability of EOT potentials and the null limit for the Sinkhorn divergence. Our rate analysis concerns the scalar EOT cost with a product-marginal reference under a DRM.

\paragraph{Transport and domain shift.}
Transport has been used to align representations across domains \citep{courty2017adaptation} and to learn a predictor and a coupling jointly \citep{courty2017joint}. We fix the representations and study precision in estimating their EOT discrepancy. OfficeHome \citep{venkateswara2017officehome} supplies the representation geometry for our experiments.

\paragraph{Density ratios and structured information sharing.}
Density ratio estimation supports importance weighting under covariate shift \citep{sugiyama2007importance}. We use empirical likelihood for the DRM \citep{qin1998casecontrol} to fit both marginals. DRM inference also covers quantile and dominance functionals \citep{chen2013quantile,zhuang2019dominance} and the bootstrap \citep{zhuang2025bootstrap}. \citet{zhang2023optimal} study parameters, distribution functions, and quantiles when one sample dominates. In that regime the EOT tilt derivative can vanish. The second-order delta method for first-order degeneracy \citep{chen2019inference} gives the quadratic tilt term, and our expansion tracks it alongside the source fluctuation.

Related causal inference work uses Sinkhorn discrepancies between counterfactual outcomes \citep{agarwal2026sinkhorn} and EOT calibration for transported treatment effects \citep{burman2026transporting}. Here the joint DRM governs the allowed changes in two observed distributions, and the scalar EOT cost is the estimand.

\section{Structured EOT Estimation}
\subsection{Density-ratio shift and transport target}
Observe independent groups $X_1,\ldots,X_{n_0}\overset{\mathrm{iid}}{\sim} F_0$ and $Y_1,\ldots,Y_{n_1}\overset{\mathrm{iid}}{\sim} F_1$. The fixed-proportion results assume $\rho_{gn}=n_g/n\to\rho_g\in(0,1)$ without a rate restriction. Fitting uses $\rho_{gn}$, and the limiting efficiency bounds use $\rho_g$. Section~\ref{sec:rare-target} treats a vanishing target fraction separately. On the support $K\subset\mathbb R^d$, let $q:K\to\mathbb R^p$ be the $p$ nonintercept basis functions, with $Q=(1,q^\top)^\top$. The DRM with true parameter $\theta_0$ satisfies
\begin{equation}\label{eq:model}
 dF_1=r\,dF_0,\qquad r=\exp\{\theta_0^\top Q\},\qquad E_0r=1.
\end{equation}
Write $\theta_0=(\alpha_0,\beta_0^\top)^\top$, with normalizing intercept $\alpha_0$ and coefficients $\beta_0$ of $q$.
Let $\Pi(F_0,F_1)$ be the set of couplings with these marginals. For fixed regularization $\varepsilon>0$ and cost $c(x,y)=\|x-y\|^2/2$, define
\begin{equation}\label{eq:target}
 \Psi_\varepsilon(F_0,F_1)=\inf_{\pi\in\Pi(F_0,F_1)}\left\{\int c\,d\pi+\varepsilon\operatorname{KL}(\pi\Vert F_0\otimes F_1)\right\}.
\end{equation}
The KL term uses the product of the two marginals as its reference. This scalar cost can be positive for identical marginals, unlike the debiased Sinkhorn divergence. NP-EOT evaluates \eqref{eq:target} at the two empirical marginals $P_{0n}=n_0^{-1}\sum_i\delta_{X_i}$ and $P_{1n}=n_1^{-1}\sum_j\delta_{Y_j}$, where $\delta_z$ is unit mass at $z$.

\Needspace{4\baselineskip}
\begin{assumption}\label{ass:conditions}
The baseline support $K\subset\mathbb R^d$ is compact; $q$ is continuous on $K$ with fixed dimension; $\theta_0$ is interior; and $M=E_1(QQ^\top/\omega)$ is positive definite, where $\omega=\rho_0+\rho_1r$.
\end{assumption}
The positive tilt is bounded above and away from zero, so both distributions have common compact support. \appref{A} records the EOT regularity and empirical-process interface, using \citet[Appendix E, Lemmas E.22--E.24]{goldfeld2024regularized} and their delta-method results. All theory conditions on a fixed representation and basis.

\subsection{DRM-EOT estimator}
The DRM lets us estimate both marginals by reweighting the pooled sample. At the true ratio, let $H_n=\rho_{0n}F_0+\rho_{1n}F_1$ and $\omega_n=\rho_{0n}+\rho_{1n}r$. Since $dH_n=\omega_n\,dF_0$, the weights $1/\omega_n$ and $r/\omega_n$ recover $dF_0$ and $dF_1$ from the same mixture.

Pool observations as $Z_1,\ldots,Z_n$, retaining multiplicities. For $\theta\in\mathbb R^{p+1}$, set $r_\theta=\exp(\theta^\top Q)$ and $\omega_{\theta,n}=\rho_{0n}+\rho_{1n}r_\theta$. Fit
\begin{equation}\label{eq:likelihood}
 \widehat\theta=\arg\max_\theta\left\{\sum_{j=1}^{n_1}\theta^\top Q(Y_j)-\sum_{i=1}^n\log\omega_{\theta,n}(Z_i)\right\}.
\end{equation}
The density-ratio fit in \eqref{eq:likelihood} has a logistic interpretation \citep{qin1998casecontrol}. For a model pair with ratio $r_\theta$, Bayes' rule in an auxiliary mixture with fractions $\rho_{gn}$ gives target-group membership probability $\rho_{1n}r_\theta(z)/\omega_{\theta,n}(z)$ and log odds $\theta^\top Q(z)+\log(\rho_{1n}/\rho_{0n})$. The logistic log likelihood adds the constant $n_0\log\rho_{0n}+n_1\log\rho_{1n}$ to the criterion in \eqref{eq:likelihood}. This interpretation keeps the group sizes fixed in the sampling experiment.

Given an interior solution, define
\begin{align}\label{eq:fitted}
 \widehat F_0&=\sum_i\frac{\delta_{Z_i}}{n\omega_{\widehat\theta,n}(Z_i)},&
 \widehat F_1&=\sum_i\frac{r_{\widehat\theta}(Z_i)\delta_{Z_i}}{n\omega_{\widehat\theta,n}(Z_i)},\\
 \widehat\Psi_\varepsilon^{\rm DRM}&=\Psi_\varepsilon(\widehat F_0,\widehat F_1).\nonumber
\end{align}
Equation~\eqref{eq:fitted} uses the pooled support for both marginals, with weights determined by the fitted ratio. The intercept score normalizes both measures. The efficiency comparison must account for uncertainty from fitting these weights. If the fit fails, DRM-EOT returns NP-EOT. Under the assumptions, an interior solution exists with probability tending to one (\appref{D}).

\begin{center}
\fbox{\begin{minipage}{.94\linewidth}
\textbf{Algorithm 1: DRM-EOT}\\[2pt]
\textbf{Input:} source $X$, target $Y$, fixed basis $Q$, regularization $\varepsilon$.\\
1. Fit the exponential tilt by empirical likelihood, \eqref{eq:likelihood}.\\
2. Construct pooled fitted marginals, \eqref{eq:fitted}.\\
3. Compute entropic OT between the fitted marginals.\\
4. For optional inference under the fixed-proportion assumptions of Theorem~\ref{th:inference}, estimate analytic standard errors or run a stratified refit bootstrap. Refit the tilt and transport in every replicate, with NP-EOT fallback when the DRM fit fails.
\end{minipage}}
\end{center}
Both estimators use Sinkhorn iteration. Before compression, DRM-EOT can put up to $n$ support points in each marginal, so dense storage and work per iteration cost $O(n^2)$ versus $O(n_0n_1)$ for NP-EOT. The tilt fit uses $p+1$ parameters. \appref{H} reports timings and peak process memory after duplicate support points are combined. The fixed-proportion limits require numerical objective error $o_p(n^{-1/2})$, with analogous conditional accuracy in bootstrap replicates. Section~\ref{sec:rare-target} gives the accuracy requirements for a vanishing target fraction.

\section{Efficiency Geometry}
\subsection{Unrestricted influence function and projection}
The efficiency question is which EOT-relevant changes the DRM rules out. We first let the marginals vary separately. We then project the resulting gradient onto the DRM tangent space, which describes the local changes the model allows.
Let $(u,v)$ be optimal dual potentials for \eqref{eq:target} and define $\phi_0=u-E_0u$, $\phi_1=v-E_1v$. For mean-zero scores $s_g$ of local changes in $F_g$, the derivative is
\begin{equation}\label{eq:derivative}
 \dot\Psi_\varepsilon(s_0,s_1)=E_0(\phi_0s_0)+E_1(\phi_1s_1).
\end{equation}
In $\mathcal H=L_0^2(F_0)\oplus L_0^2(F_1)$, where $L_0^2$ denotes square-integrable mean-zero functions, the squared norm of $h=(h_0,h_1)$ is $\rho_0E_0h_0^2+\rho_1E_1h_1^2$. The unrestricted gradient is $(\phi_0/\rho_0,\phi_1/\rho_1)$, with squared norm
\begin{equation}\label{eq:np}
 V_{\rm NP}=E_0\phi_0^2/\rho_0+E_1\phi_1^2/\rho_1.
\end{equation}
The DRM couples the marginal changes through a shared baseline and a tilt in the span of $Q$. The part of the gradient affected by this restriction depends on the sampling-weighted difference between the potentials. Define the joint potential contrast $\zeta$:
\begin{equation}\label{eq:projection}
 \zeta=\phi_1-\frac{\rho_1}{\rho_0}\phi_0,\quad b^\star=M^{-1}E_1(Q\zeta/\omega),\quad e=\zeta-b^{\star\top}Q.
\end{equation}
Here $b^\star$ minimizes $E_1\{(\zeta-b^\top Q)^2/\omega\}$, so $E_1(Qe/\omega)=0$. We denote the weighted projection $b^{\star\top}Q$ by $\Pi_Q\zeta$.

\subsection{Canonical gradient and exact efficiency identity}
\begin{theorem}[Canonical gradient and variance reduction]\label{th:geometry}
Under Assumption~\ref{ass:conditions}, the DRM canonical gradient is
\begin{equation}\label{eq:gradient}
 g_0=\frac{\phi_0}{\rho_0}+\frac{re}{\omega},\qquad g_1=\frac{\phi_1}{\rho_1}-\frac{\rho_0}{\rho_1}\frac e\omega.
\end{equation}
Its efficiency bound is
\begin{equation}\label{eq:gain}
 V_{\rm DRM}:=\rho_0E_0g_0^2+\rho_1E_1g_1^2=V_{\rm NP}-\frac{\rho_0}{\rho_1}E_1(e^2/\omega).
\end{equation}
Thus $V_{\rm DRM}<V_{\rm NP}$ if and only if $e\ne0$ in $L^2(F_1)$; equality holds if and only if $\zeta\in\operatorname{span}(Q)$ $F_1$-almost surely.
\end{theorem}
The gain is positive when the potential contrast has a component outside the tilt span. Its size depends on the cost, distribution pair, and sampling fractions through the weighted residual. Even under correct specification, a contrast entirely in the tilt span gives zero gain. \appref{C} gives the proof outline and derivation.

The DRM canonical gradient represents the EOT derivative along the allowed paths. Its squared norm is the efficiency bound. For two correctly specified bases in the same experiment with $\operatorname{span}(Q_1)\subseteq\operatorname{span}(Q_2)$, the larger span admits additional tilt directions. The squared residual can only decrease, so $V_{\rm DRM}(Q_1)\le V_{\rm DRM}(Q_2)\le V_{\rm NP}$. For example, with $Q_1=(1,x)^\top$ and $Q_2=(1,x,x^2)^\top$ in Table~\ref{tab:simulation-main}, the quadratic term captures nearly all the potential contrast. The resulting gain relative to NP-EOT is negligible.

Sampling fractions also affect the contrast. When $F_0=F_1$ is symmetric about zero on the real line and $q(x)=x$, the centered potential $\phi$ is even. The gain becomes $(\rho_0-\rho_1)^2\operatorname{Var}(\phi)/(\rho_0\rho_1)$ and increases with imbalance in this example (\appref{C}). With identical marginals, a symmetric cost, and balanced fractions, $\zeta\equiv0$ for any tilt basis.

\Needspace{8\baselineskip}
\section{Estimation, Sensitivity, and Target Scarcity}
\subsection{Attainable efficiency and inference}
The bound gives the best first-order precision available to regular estimators in the DRM. Regularity means that the limiting distribution, centered at the moving EOT target, is stable along local DRM submodels.
\begin{theorem}[Efficient DRM-EOT estimation]\label{th:estimation}
Under Assumption~\ref{ass:conditions}, DRM-EOT is regular, asymptotically linear with the canonical gradient in \eqref{eq:gradient}, and semiparametrically efficient:
\begin{equation}\label{eq:linear}
 \sqrt n(\widehat\Psi_\varepsilon^{\rm DRM}-\Psi_\varepsilon)\Rightarrow N(0,V_{\rm DRM}).
\end{equation}
\end{theorem}
\appref{D} obtains the joint process limit of the fitted marginals, identifies the estimator's influence function with the canonical gradient in \eqref{eq:gradient}, and verifies regularity along local DRM submodels. The score and fitted-measure expansions use the actual sampling fractions, so no rate assumption on $\rho_{gn}\to\rho_g$ is needed.

\begin{theorem}[Feasible inference]\label{th:inference}
Under the same assumptions, the analytic estimator $\widehat V_{\rm DRM}$ in \appref{E} satisfies $\widehat V_{\rm DRM}\to_pV_{\rm DRM}$. A stratified refit bootstrap that resamples $n_g$ observations independently within each observed group and refits DRM-EOT has conditional limit $N(0,V_{\rm DRM})$ for $\sqrt n(\widehat\Psi_\varepsilon^*-\widehat\Psi_\varepsilon^{\rm DRM})$. If $V_{\rm DRM}>0$, Wald and centered basic-bootstrap confidence intervals are asymptotically valid.
\end{theorem}
Both intervals target the scalar $\Psi_\varepsilon(F_0,F_1)$ in \eqref{eq:target}. The analytic variance estimate sums the within-group empirical variances of the fitted influence functions, with weight $\rho_{gn}$ for group $g$. The Wald interval is $\widehat\Psi_\varepsilon^{\rm DRM}\pm\Phi^{-1}(1-\alpha/2)\sqrt{\widehat V_{\rm DRM}/n}$, with $\Phi$ the standard normal distribution function. The bootstrap keeps the group sizes fixed and repeats the full fit, including the density ratio. \appref{E} gives the variance and conditional-process formulas.

\subsection{Local misspecification sensitivity and the efficiency trade-off}
To assess local sensitivity, start at a correctly specified pair. Hold $F_0$ fixed and take the contiguous alternative $dF_{1,n}=(1+\eta\delta/\sqrt n)dF_1$, with bounded $\delta$ satisfying $E_1\delta=0$ and fixed $\eta$. Centering at the perturbed transport target gives limiting mean shift $\eta B_\delta$, where
\begin{equation}\label{eq:bias}
 B_\delta=\rho_1E_1(g_1\delta)-E_1(\phi_1\delta)=-\rho_0E_1(e\delta/\omega).
\end{equation}
Consequently,
\begin{equation}\label{eq:bound}
 |B_\delta|\le\sqrt{\rho_0\rho_1(V_{\rm NP}-V_{\rm DRM})E_1(\delta^2/\omega)}.
\end{equation}
The squared residual determines the variance gain. Its weighted inner product with $\delta$ determines the signed local bias. At a fixed weighted perturbation norm, a larger gain raises the Cauchy--Schwarz upper bound in \eqref{eq:bound}. Equality for a nonzero gain requires a direction proportional to $e$, which may violate $E_1\delta=0$. The bound can therefore be loose even within this local class. \appref{F} gives the calculation for contiguous departures.

\Needspace{27\baselineskip}
\subsection{Vanishing target fraction}\label{sec:rare-target}
With positive limiting fractions, the DRM can change the asymptotic variance. Under a correct DRM with a vanishing target fraction, the allowed tilt directions can also change the convergence rate. The rate depends on whether estimating the tilt affects the cost to first order.

Let $N=n_0$, $m=n_1$, and write $\beta=\beta_0$ at the true pair. For a distribution $F$, use $F(f)=\int f\,dF$. Write $F_1=R_{F_0,\beta}$, where $R_{F,\beta}(dx)=e^{\beta^\top q(x)}F(dx)/F(e^{\beta^\top q})$, and let $T(F,\beta)=\Psi_\varepsilon(F,R_{F,\beta})$. At the true $(F_0,\beta)$, put $q_c=q-E_1q$, $J=E_1(q_cq_c^\top)$, and
\[
a=\partial_\beta T(F_0,\beta)=E_1(\phi_1q_c),
\qquad \Gamma=\partial_\beta^2T(F_0,\beta).
\]
At $a=0$ the first-order tilt term disappears, and source variation can remain. The residual $e$ in Theorem~\ref{th:geometry} measures the variance gain at positive limiting fractions.

\begin{proposition}[Convergence when the target fraction vanishes]\label{th:rare-target}
Assume $N,m\to\infty$ with $m/N\to0$, the correctly specified model \eqref{eq:model}, compact support, fixed $\varepsilon>0$, and a fixed continuous basis with $J\succ0$. Let $\widehat T$ be the DRM-EOT estimate in \eqref{eq:fitted}, and write $T=T(F_0,\beta)$. Define $\psi=\phi_0+r\phi_1$, $\sigma^2=E_0\psi^2$, and the independent sums
\[
S_N=N^{-1/2}\sum_{i=1}^N\psi(X_i),\qquad
G_m=J^{-1}m^{-1/2}\sum_{j=1}^m q_c(Y_j).
\]
If $a\ne0$, then $\sqrt m(\widehat T-T)\Rightarrow \mathcal N(0,a^\top J^{-1}a)$. If $a=0$, $\sigma^2>0$, and $\Gamma\ne0$, then
\begin{equation}\label{eq:rare-expansion}
\widehat T-T=\frac{S_N}{\sqrt N}+\frac{G_m^\top\Gamma G_m}{2m}
             +o_p(N^{-1/2}+m^{-1}).
\end{equation}
\end{proposition}
The two fluctuations in \eqref{eq:rare-expansion} are of the same order at $m\asymp\sqrt N$. Let $S\sim\mathcal N(0,\sigma^2)$ and $G\sim\mathcal N_p(0,J^{-1})$ be independent. Under the proposition's orthogonality and nondegeneracy conditions,
\[
\begin{array}{ll}
m/\sqrt N\to\infty &
\sqrt N(\widehat T-T)\Rightarrow S,\\[2pt]
m/\sqrt N\to\kappa\in(0,\infty) &
\sqrt N(\widehat T-T)\Rightarrow S+G^\top\Gamma G/(2\kappa),\\[2pt]
m/\sqrt N\to0 &
m(\widehat T-T)\Rightarrow G^\top\Gamma G/2.
\end{array}
\]
The quadratic forms are uncentered. \appref{K} proves the expansion. These limits require numerical error relative to the exact plug-in value of $o_p(m^{-1/2})$ when $a\ne0$ and $o_p(N^{-1/2}+m^{-1})$ in the orthogonal case.

For the uniform law on $[-1,1]$ with $F_0=F_1$ and $\varepsilon=1$, the bases $Q_1=(1,x)^\top$ and $Q_2=(1,x,x^2)^\top$ both contain the true ratio through the zero tilt. The centered potential $\phi$ is even and strictly increasing in $|x|$. Hence $Q_1$ gives $a=0$ and satisfies the nondegeneracy conditions in \appref{K}. The quadratic direction in $Q_2$ has $\operatorname{Cov}(\phi(X),X^2)>0$ and restores the $m^{-1/2}$ error term. The two fixed models differ in the tilt directions they allow, which changes the rate at the same pair of distributions.

\section{Experiments}\label{sec:numerics}
The controlled studies examine the fixed-proportion results. The synthetic study compares predicted and observed variance gains and checks interval coverage. Exact tilts on OfficeHome representations let us compare empirical and oracle variance ratios across sample sizes. The natural OfficeHome domains and cash dividend proposal groups assess variance and reference RMSE under a working DRM. Section~\ref{sec:rare-target} illustrates the vanishing-target limits with a population example.

\subsection{Controlled synthetic validation}
The synthetic study compares two correctly specified bases with different predicted gains. We fit DRM-EOT with $Q_1(x)=(1,x)^\top$ and $Q_2(x)=(1,x,x^2)^\top$ for $F_0=F_1=\operatorname{Unif}[-1,1]$. We set $\varepsilon=1$ and $(n_0,n_1)=(80,320)$. The EOT target is $0.2799288$ by 800-node quadrature. Relative to NP-EOT, $Q_1$ has a $36.0\%$ theoretical gain and a $36.8\%$ Monte Carlo gain (Table~\ref{tab:simulation-main}). The potential contrast is even, so its nonconstant part remains in the residual under the smaller basis. The quadratic term in $Q_2$ approximates that part closely. Quadrature gives an estimated gain of $2.71\times10^{-7}$ on the $V$ scale. Wald and refit-bootstrap coverage are near nominal, with mild bootstrap conservatism for $Q_1$. \appref{G} gives the projection diagnostics and coverage details.
\begin{table}[H]
\centering\small
\caption{Synthetic validation at $n=400$, $(\rho_0,\rho_1)=(0.2,0.8)$, and $\varepsilon=1$, with $Q_1(x)=(1,x)^\top$ and $Q_2(x)=(1,x,x^2)^\top$. $V$ is the asymptotic variance of the $\sqrt n$-scaled error. MC $n\mathrm{Var}$ is $n$ times the Monte Carlo variance of the estimates. Variances and nominal 95\% Wald coverage use 2,000 repetitions. Bootstrap coverage uses 1,000 repetitions with 499 stratified refits each. Dashes mark coverage entries outside the evaluation.}\label{tab:simulation-main}
\begin{tabular}{lrrrrrr}\toprule
Method & Theory $V$ & MC $n\mathrm{Var}$ & Theory gain & MC gain & Wald cov. & Boot. cov.\\\midrule
NP-EOT & 0.06581 & 0.07030 & 0 & 0 & -- & --\\
DRM-EOT $Q_1$ & 0.04212 & 0.04444 & 36.0\% & 36.8\% & 96.05\% & 97.10\%\\
DRM-EOT $Q_2$ & 0.06581 & 0.07028 & $\approx0$ & $\approx0$ & 95.15\% & 95.50\%\\\bottomrule
\end{tabular}
\end{table}

\subsection{OfficeHome exact-model validation and natural-shift diagnostics}
OfficeHome contains 15,588 images and 65 classes across Art, Clipart, Product, and Real World domains \citep{venkateswara2017officehome}. We take penultimate features from a frozen ImageNet-pretrained ResNet-50 \citep{he2016resnet}. For each held-out target domain, we fit PCA on the other three domains. We retain the first $d$ principal components and normalize to obtain a unit vector $z$. The default is $d=16$, $q(z)=z$, $c(z,z')=\|z-z'\|^2/2$, and $\varepsilon=0.1$. The source and target sample sizes are $n_S=n_0$ and $n_T=n_1$. The analysis uses the fixed representation and its compact support. Image labels play no role in the fit.

For each base domain, the exact-model study uses the same frozen map fitted on the other three domains, with $d=16$ and unit normalization. On the resulting $N_K$ support points $\{z_i\}_{i=1}^{N_K}$, set $P_0(z_i)=1/N_K$ and $P_1(z_i)=P_0(z_i)\exp\{\alpha+\beta^\top z_i\}$ using the first PCA coordinate as the tilt direction. We calibrate the tilt to $\operatorname{ESS}(P_1)/N_K\in\{0.8,0.5\}$, where $\operatorname{ESS}(P_1)=1/\sum_iP_1(z_i)^2$ is the effective sample size of the target masses. We compute the population EOT and oracle bound on the finite support to numerical tolerance. The oracle ratio is below one at each evaluated geometry and allocation. At $(\rho_0,\rho_1)=(0.8,0.2)$, the four-geometry means are $0.929$ and $0.952$ at ESS ratios $0.8$ and $0.5$, respectively.

\begin{figure}[H]
\centering\includegraphics{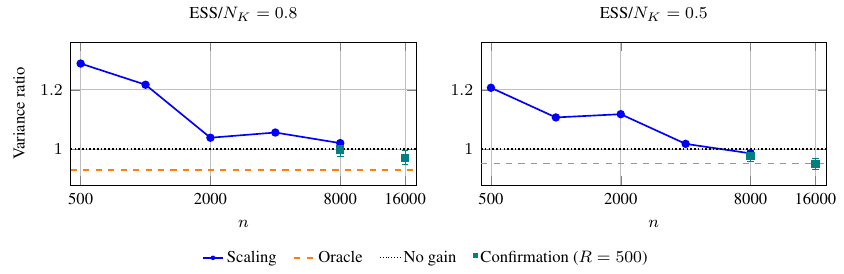}
\caption{Exact-DRM OfficeHome scaling at $(\rho_0,\rho_1)=(0.8,0.2)$. The curve averages DRM-EOT/NP-EOT variance ratios across four fixed geometries with 100 repetitions per cell. The horizontal axis uses a log scale. Points with 95\% pointwise Monte Carlo intervals use 500 confirmation repetitions per cell (Table~\ref{tab:tail}), including the 100 scaling draws at $n=8000$.}\label{fig:exact-scaling}
\end{figure}

We keep $(\rho_0,\rho_1)=(0.8,0.2)$ while increasing $n$ through $500,1000,2000,4000,8000$. The empirical variance ratios start above one and move toward the oracle levels below one (Figure~\ref{fig:exact-scaling}). The paths are nonmonotone. The smaller-sample ratios illustrate the difference between the asymptotic bound and the variance at a given sample size.

The confirmation run uses $R=500$ repetitions per geometry and cell. At ESS ratio $0.5$ and $n=16000$, the empirical variance ratio is $0.950$ (95\% Monte Carlo interval $[0.932,0.969]$), close to the oracle ratio of $0.952$. At ESS ratio $0.8$, the variance ratio is $0.971$ ($[0.948,0.995]$), above the oracle ratio of $0.929$. The interval for that ESS ratio at $n=8000$ includes one. The intervals quantify Monte Carlo uncertainty conditional on the four geometries. Table~\ref{tab:tail} reports empirical RMSE ratios alongside the variance ratios.

\begin{table}[H]
\centering\small
\caption{Exact-DRM confirmation at $(\rho_0,\rho_1)=(0.8,0.2)$ with 500 repetitions per geometry and cell. Ratios divide DRM-EOT by NP-EOT and are averaged across four geometries. ESS/$N_K$ measures concentration of the tilted masses. Intervals are 95\% pointwise Monte Carlo intervals.}\label{tab:tail}
\begin{tabular}{rrrrrr}\toprule
ESS/$N_K$ & $n$ & Var. ratio & MC interval & Oracle ratio & RMSE ratio\\\midrule
0.5 & 8000 & 0.978 & $[0.959,0.997]$ & 0.952 & 0.944\\
0.5 & 16000 & 0.950 & $[0.932,0.969]$ & 0.952 & 0.958\\
0.8 & 8000 & 0.999 & $[0.976,1.023]$ & 0.929 & 0.867\\
0.8 & 16000 & 0.971 & $[0.948,0.995]$ & 0.929 & 0.911\\\bottomrule
\end{tabular}
\end{table}

\Needspace{10\baselineskip}
\paragraph{Natural OfficeHome diagnostics.}
For observed OfficeHome domain pairs, we use the DRM as a working model. The fixed-geometry study varies $p=8,16,32,64$ at $d=64$. It retains every draw, with NP-EOT fallback on failed fits (\hyperref[sec:e1-fixed]{Appendix~\ref*{sec:e1-fixed}}). At $n_T=50$, $p=32$ gives the lowest mean of the task-level RMSE ratios. At $n_T=200$ and $800$, all four means exceed one. Mean variance ratios exceed one throughout this grid (Appendix Table~\ref{tab-e1-fixed-errors}). The baseline study uses $d=16$, $q(z)=z$, and paired successes. At $n_T=50$ its ratio of task-averaged RMSEs is $0.739$, and its mean task-level variance ratio is $1.123$ (Appendix Table~\ref{tab-natural-summary}).

\subsection{Annual cash dividend proposals}\label{sec:dividend}
We examine finite-sample EOT estimation for naturally observed groups under a working DRM. The groups consist of Chinese A-share firms whose recorded annual proposals specify a positive or explicitly zero pre-tax cash amount per share. Proposals for a fiscal year are disclosed in the next calendar year. We measure each firm's adjusted 60-session return, log volatility over the last 20 sessions, and log market capitalization at fiscal year-end, before disclosure. We fix the feature scaling on FY2018--FY2022 and use FY2023--FY2024 as later evaluation cohorts. Firms can appear in multiple years.

To study estimation with a small target sample, we draw $n_0=240$ positive-proposal and $n_1\in\{20,40\}$ zero-proposal observations with replacement within each fiscal year. NP-EOT and DRM-EOT use the same 600 draws per configuration. A linear model for group log odds motivates the working DRM, which uses an intercept and the three standardized features. Both estimators use half squared Euclidean cost and $\varepsilon=0.1$. The RMSE reference is NP-EOT on the two complete annual pools. Appendix~\ref{sec:dividend-pools} gives cohort rules and sizes, Appendix~\ref{sec:dividend-overlap} describes the fitted-score overlap diagnostic, and Appendix~\ref{sec:dividend-evaluation} gives the evaluation protocol and numerical checks.

Conditional on the annual pools, the 95\% pointwise Monte Carlo intervals lie below one for 12 of the 14 variance ratios and all 14 reference RMSE ratios (Table~\ref{tab:dividend-main}). Both FY2023 variance intervals include one. In FY2024, the variance ratios of $0.637$ and $0.618$ correspond to reductions of $36.3\%$ and $38.2\%$ at $n_1=20$ and $40$, respectively. The comparisons describe finite-sample performance under the working model.

\begin{table}[H]
\centering\small
\caption{Cash dividend proposal results with $n_0=240$, $\varepsilon=0.1$, and 600 paired draws with replacement per configuration. Ratios divide DRM-EOT by NP-EOT. Brackets give 95\% pointwise Monte Carlo intervals conditional on the annual empirical pools. FY2018--FY2022 determine feature scaling, which is fixed for FY2023--FY2024.}\label{tab:dividend-main}
\begin{tabular}{lrll}\toprule
Fiscal year & $n_1$ & Variance ratio [interval] & RMSE ratio [interval]\\\midrule

FY2018 & 20 & 0.660 [0.579, 0.752] & 0.643 [0.603, 0.684] \\
FY2018 & 40 & 0.741 [0.676, 0.809] & 0.688 [0.654, 0.723] \\
FY2019 & 20 & 0.448 [0.335, 0.681] & 0.611 [0.547, 0.695] \\
FY2019 & 40 & 0.263 [0.196, 0.365] & 0.472 [0.413, 0.545] \\
FY2020 & 20 & 0.644 [0.307, 0.876] & 0.754 [0.527, 0.892] \\
FY2020 & 40 & 0.563 [0.384, 0.692] & 0.697 [0.572, 0.783] \\
FY2021 & 20 & 0.669 [0.578, 0.751] & 0.717 [0.661, 0.767] \\
FY2021 & 40 & 0.737 [0.655, 0.810] & 0.780 [0.730, 0.821] \\
FY2022 & 20 & 0.660 [0.571, 0.739] & 0.769 [0.711, 0.816] \\
FY2022 & 40 & 0.621 [0.554, 0.695] & 0.743 [0.700, 0.788] \\
FY2023 & 20 & 0.875 [0.780, 1.004] & 0.774 [0.733, 0.823] \\
FY2023 & 40 & 0.957 [0.849, 1.079] & 0.809 [0.760, 0.863] \\
FY2024 & 20 & 0.637 [0.550, 0.746] & 0.698 [0.648, 0.750] \\
FY2024 & 40 & 0.618 [0.555, 0.692] & 0.706 [0.668, 0.751] \\

\bottomrule
\end{tabular}
\end{table}

\Needspace{8\baselineskip}
\section{Discussion and Limitations}
The gain from a density-ratio restriction depends on how the allowed changes in the marginals affect the EOT cost. At positive limiting sampling fractions, DRM-EOT attains the restricted bound, with an exact gain determined by the projection residual. When the target fraction vanishes, cancellation of the first-order tilt effect can change the convergence rate. In the dividend study, pooling lowered variance in several annual pools and lowered reference RMSE throughout the grid. In the natural OfficeHome baseline, DRM-EOT had lower reference bias and RMSE at small target sizes, with higher variance. The RMSE ratio exceeded one at larger target sizes.

The efficiency and rate results assume a correctly specified DRM on compact support, fixed $\varepsilon>0$, a fixed-dimensional continuous basis, and a fixed representation. Theorems~\ref{th:geometry}--\ref{th:inference} use positive limiting sampling fractions. The confidence intervals also require positive first-order variance. Proposition~\ref{th:rare-target} gives pointwise limits for a vanishing target fraction. Inference in that regime, fixed misspecification, shrinking regularization, and data-driven basis selection call for separate analyses. Unbounded support requires additional controls (\appref{I}). Penalizing the tilt may mitigate near-separation. It remains to determine which penalties preserve the efficiency result.

\section*{Reproducibility Statement}
Appendices A--F and K give assumptions and proofs. Appendices G, H and J document the experiments. The reviewer archive is named \texttt{anonymous\_code\_reviewer.zip}. Its top-level README gives separate commands for regenerating tables from saved estimates, refitting DRM-EOT, and checking derivatives for \appref{K}. Dividend refits require a separate company-year feature snapshot. The data preparation guide in the archive and \appref{J} document the authorized inputs, build command, and checksum.

\section*{AI Use Statement}
Generative AI tools assisted with the design of additional experiments, implementation and verification of experiment code, data formatting and analysis, and interpretation of results. They also assisted with manuscript and rebuttal drafting and editing, notation and reference checks, literature verification, and review of the mathematical proofs. The authors are responsible for the final text, mathematical claims, and reported results.


\clearpage

\appendix
\setcounter{figure}{0}
\setcounter{table}{0}
\setcounter{theorem}{0}
\setcounter{lemma}{0}
\setcounter{corollary}{0}
\def\theHfigure{appendix.\arabic{figure}}
\def\theHtable{appendix.\arabic{table}}
\def\theHtheorem{appendix.\arabic{theorem}}
\def\theHlemma{appendix.\arabic{lemma}}
\def\theHcorollary{appendix.\arabic{corollary}}
\section*{Appendices}
\addcontentsline{toc}{section}{Appendices}

\noindent Appendices~\ref{app:A}--\ref{app:F} use positive limiting sampling proportions. Appendix~\ref{app:A} states assumptions A1--A4 and the EOT regularity results. Appendices~\ref{app:B}--\ref{app:C} derive the tangent-space projection and efficiency bound. Appendix~\ref{app:D} proves attainment by the fitted estimator, Appendix~\ref{app:E} establishes feasible inference, and Appendix~\ref{app:F} derives the mean shift under local misspecification. Appendix~\ref{app:K} studies the same estimator when the target fraction vanishes.

\noindent Appendix Theorems~1 and~2 establish main-text Theorems~1 and~2. Appendix Theorems~3--4 establish main-text Theorem~3. Appendix~\ref{app:K} proves main-text Proposition~1.

\noindent Appendices~\ref{app:G}, \ref{app:H}, and~\ref{app:J} give the experiment protocols and full results. Appendix~\ref{app:I} discusses what an extension to unbounded support would require. Tables, figures, and theorems are numbered separately in the main text and appendices. Appendix equation tags include the section letter.

\section{Regularized EOT preliminaries}\label{app:A}
We describe the two-sample experiment, state the assumptions for efficiency and inference, and derive the unrestricted EOT gradient. Throughout Appendices A--F, the limiting sampling fractions are positive and the regularization and tilt basis are fixed. Appendix~\ref{app:K} treats a vanishing target fraction.

\hypertarget{statistical-experiment-and-target-functional}{%
\subsection{Statistical experiment and target functional}\label{statistical-experiment-and-target-functional}}

\hypertarget{fixed-two-sample-experiment}{%
\subsubsection{Fixed two-sample experiment}\label{fixed-two-sample-experiment}}

The experiment consists of two independent samples, \[
X_1,\ldots,X_{n_0}\overset{\mathrm{iid}}{\sim}F_0,
\qquad
Y_1,\ldots,Y_{n_1}\overset{\mathrm{iid}}{\sim}F_1,
\] The total sample size and the finite-sample and limiting group fractions are \[
n=n_0+n_1,
\qquad
\rho_{gn}=\frac{n_g}{n},
\qquad
\rho_{gn}\to\rho_g\in(0,1),
\qquad g=0,1,
\] with \[
\rho_{0n}+\rho_{1n}=1,
\qquad
\rho_0+\rho_1=1.
\]

Write $E_gf=F_gf=\int f\,dF_g$ and $P_{0n}=n_0^{-1}\sum_i\delta_{X_i}$, $P_{1n}=n_1^{-1}\sum_j\delta_{Y_j}$. The ambient dimension is $d$, and the basis has $p$ nonintercept functions. We treat the group sizes as fixed. The group fractions may converge without a rate condition of the form \[
\sqrt n(\rho_{gn}-\rho_g)\to0
\] because the finite-\(n\) empirical-likelihood and empirical-process expansions retain the exact fractions \(\rho_{0n},\rho_{1n}\). Define the finite-sample weight \[
\omega_n(x)
=
\rho_{0n}+\rho_{1n}r(x)
\] and the deterministic triangular-array mixture \[
H_n
=
\rho_{0n}F_0+\rho_{1n}F_1,
\qquad
dH_n(x)=\omega_n(x)\,dF_0(x).
\] The limiting quantities \[
\omega(x)=\rho_0+\rho_1r(x),
\qquad
H=\rho_0F_0+\rho_1F_1
\] enter after the finite-\(n\) expansions. This avoids introducing a first-order term by replacing \(\rho_{gn}\) with \(\rho_g\) in those expansions.

\hypertarget{density-ratio-model}{%
\subsubsection{Density ratio model}\label{density-ratio-model}}

Let \(q:K\to\mathbb R^p\) be a known basis and define \[
Q(x)=
\begin{pmatrix}
1\\q(x)
\end{pmatrix}
\in\mathbb R^{p+1}.
\]

Assume \[
\frac{dF_1}{dF_0}(x)
=
r(x)
=
r_{\theta_0}(x)
=
\exp\{\theta_0^\top Q(x)\},
\] where \[
\theta_0=
\begin{pmatrix}
\alpha_0\\\beta_0
\end{pmatrix}.
\] Normalization implies \[
E_0r(X)=1.
\]

Appendix~\ref{app:B} uses the equivalent normalized parameterization \[
r_{\beta,F_0}(x)
=
\frac{\exp\{\beta^\top q(x)\}}
{E_0[\exp\{\beta^\top q(X)\}]},
\] which expresses the intercept as a function of \((F_0,\beta)\).

\hypertarget{entropic-optimal-transport}{%
\subsubsection{Entropic optimal transport}\label{entropic-optimal-transport}}

For fixed \(\varepsilon>0\), the scalar EOT cost with ground cost \[
c(x,y)=\frac12\|x-y\|^2,
\] has value \[
\Psi_\varepsilon(F_0,F_1)
=
\operatorname{EOT}_\varepsilon(F_0,F_1)
\] by \[
\operatorname{EOT}_\varepsilon(F_0,F_1)
=
\inf_{\pi\in\Pi(F_0,F_1)}
\left\{
\int c(x,y)\,d\pi(x,y)
+
\varepsilon
\operatorname{KL}
\left(
\pi\,\Vert\,F_0\otimes F_1
\right)
\right\}.
\]

The efficiency and inference results hold with \(\varepsilon\) fixed.

\hypertarget{main-assumptions}{%
\subsection{Structural assumptions}\label{main-assumptions}}

Assumptions A1--A4 make the conditions in main-text Assumption 1 explicit. A1 specifies the fixed-fraction limit. A2 gives the compact domain used in the potential regularity argument. A3 gives a bounded change of measure between the two marginals, and A4 ensures the weighted projection coefficients in Appendix~\ref{app:C} are unique.

\hypertarget{assumption-a1-sampling-proportions}{%
\subsubsection{Assumption A1: sampling proportions}\label{assumption-a1-sampling-proportions}}

\[
\rho_{gn}\to\rho_g\in(0,1),
\qquad g=0,1.
\]

The convergence of \(\rho_{gn}-\rho_g\) requires no rate condition. At each finite \(n\), the proof retains \(\rho_{gn}\), \(\omega_n\), \(H_n\), \(W_n\), and the corresponding \(n\)-dependent influence functions until the final triangular-array limit.

\hypertarget{assumption-a2-compact-baseline-support}{%
\subsubsection{Assumption A2: compact baseline support}\label{assumption-a2-compact-baseline-support}}

There exists a compact set \(K\subset\mathbb R^d\) such that \[
\operatorname{supp}(F_0)=K.
\]

\hypertarget{assumption-a3-correct-drm-and-bounded-basis}{%
\subsubsection{Assumption A3: correct fixed-dimensional DRM}\label{assumption-a3-correct-drm-and-bounded-basis}}

The DRM is correctly specified with fixed-dimensional basis \(q\) at an interior parameter \(\theta_0\), and each component of \(q\) is continuous on \(K\). Hence \(q\) and \(Q\) are bounded, while the positive continuous tilt satisfies
\[
0<\inf_{x\in K}r(x)\le \sup_{x\in K}r(x)<\infty.
\]
Consequently, \(F_0\) and \(F_1\) are mutually absolutely continuous and
\[
\operatorname{supp}(F_1)=\operatorname{supp}(F_0)=K.
\]
In particular, \(r\), \(1/r\), and \[
\omega(x):=\rho_0+\rho_1r(x)
\] are bounded on \(K\), with \(\omega\) bounded away from zero.

\hypertarget{assumption-a4-drm-nondegeneracy}{%
\subsubsection{Assumption A4: DRM nondegeneracy}\label{assumption-a4-drm-nondegeneracy}}

The matrix \[
M
=
E_1\left[
\frac{Q(X)Q(X)^\top}{\omega(X)}
\right]
\] is positive definite.

Equivalently, the augmented basis \(Q\) is linearly independent in the weighted \(L^2(F_1)\) space. The subsequent lemmas construct the smooth potential class, empirical-process envelopes, and local parameter neighborhoods on the common compact support. They give the measurability conventions where needed.

\hypertarget{eot-pathwise-derivative-in-the-unrestricted-model}{%
\subsection{EOT pathwise derivative in the unrestricted model}\label{eot-pathwise-derivative-in-the-unrestricted-model}}

\setcounter{lemma}{0}
\begin{lemma}[Regularity class for fixed-\(\varepsilon\) EOT on compact support]\label{lem:eot-regularity}

Under A2 and the fixed quadratic-cost EOT definition in Section~A.1, choose a closed ball \(B\) containing \(K\) in its interior and an integer \(s>d/2\), \(s\ge2\). There is a sufficiently large bounded \(C^s(B)\) ball whose restrictions to \(K\), enlarged to include the constant function $1$ and separately centered potentials, form a class \(\mathscr F_\varepsilon\) with the following properties:
\begin{enumerate}
\item EOT potentials normalized by $u(x_\circ)=0$ at a fixed $x_\circ\in K$, for every pair of marginals in \(\mathcal P(K)^2\) belong to \(\mathscr F_\varepsilon\);
\item the uniform log covering number satisfies $\log N(\eta,\mathscr F_\varepsilon,\|\cdot\|_{\infty,K})=O(\eta^{-d/s})$, so its entropy integral is finite, and \(\mathscr F_\varepsilon\) is uniformly bounded and Donsker;
\item regarding signed perturbations as elements of \(\ell^\infty(\mathscr F_\varepsilon)^2\), the EOT functional is tangentially Hadamard differentiable on the relevant feasible tangent set, with derivative
\[
(h_0,h_1)\longmapsto h_0(u)+h_1(v).
\]
\end{enumerate}
The Gaussian-process support relevant below is a closed linear subspace of the feasible tangent cone. Thus the same differentiability statement applies to the empirical-process and bootstrap delta methods.

\end{lemma}
\begin{proof}

The Schr\"odinger equations give uniform oscillation bounds for normalized quadratic-cost potentials. Differentiation on \(B\) gives derivative bounds of every fixed order. The standard smooth-class entropy bound \citep{vanderVaartWellner1996} for a bounded \(C^s(B)\) ball is of order at most \(\eta^{-d/s}\). Since \(s>d/2\), the entropy integral is finite and the restricted class is Donsker.

\citet[Appendix E, Lemma E.22]{goldfeld2024regularized} gives regular versions of the potentials, Lemma E.23 gives local Lipschitz continuity and the directional derivative, and Lemma E.24 gives the Donsker property. The two-marginal argument in their proof of Theorem 7(ii), together with Proposition 1 and Corollary 1, supplies the differentiability needed for the delta method and bootstrap when the derivative is linear on the Gaussian support. We apply this argument on the convex domain \(\mathcal P(K)^2\), where both marginals remain supported on \(K\). Rescaling \(x\) by \(\varepsilon^{-1/2}\) matches their unit-regularization convention, and all measures on $K$ share a finite sub-Gaussian envelope. Enlarging the smooth ball to contain the restrictions of their potential class transfers the Lipschitz and differentiability bounds to the norm used here. This smoothness argument concerns the transport potentials, and the DRM basis needs only the continuity in A3.

Define the feasible tangent cone at the reference pair as the norm closure in $\ell^\infty(\mathscr F_\varepsilon)^2$ of the cone of nonnegative multiples of admissible perturbations from that pair. It is closed by construction and convex because $\mathcal P(K)^2$ is convex. Any weak limit of these differences is supported on that cone. The support of a centered Gaussian limit is a closed linear subspace.

\end{proof}

Let \((u,v)\) be optimal dual potentials for \[
\operatorname{EOT}_\varepsilon(F_0,F_1).
\] They are unique up to the transformation \[
(u,v)\mapsto(u+a,v-a).
\]

Center each potential under its corresponding marginal to obtain \[
\phi_0(x)=u(x)-E_0u(X),
\qquad
\phi_1(x)=v(x)-E_1v(X).
\] These representatives give the same derivative as the original potential pair when integrated against zero-mass perturbations.

For scores \[
s_0\in L_0^2(F_0),
\qquad
s_1\in L_0^2(F_1),
\] the pathwise derivative is \[
\dot\Psi_\varepsilon(s_0,s_1)
=
E_0[\phi_0s_0]
+
E_1[\phi_1s_1].
\tag{A.1}
\]

The fixed limiting fractions determine the inner product on the score space \[
\mathcal H
=
L_0^2(F_0)\oplus L_0^2(F_1)
\] through \[
\langle a,b\rangle_{\mathcal H}
=
\rho_0E_0[a_0b_0]
+
\rho_1E_1[a_1b_1].
\tag{A.2}
\]

The notation $g_g$ in the main text abbreviates $g_{g,\mathrm{DRM}}$ used below. Equations (A.1)--(A.2) give the unrestricted canonical gradient \[
{
g_{\mathrm{NP}}
=
\left(
\frac{\phi_0}{\rho_0},
\frac{\phi_1}{\rho_1}
\right).
}
\tag{A.3}
\] Its squared norm is \[
{
V_{\mathrm{NP}}
=
\frac{E_0[\phi_0^2]}{\rho_0}
+
\frac{E_1[\phi_1^2]}{\rho_1}.
}
\tag{A.4}
\]

\section{DRM tangent space}\label{app:B}
We first derive the scores of local changes allowed by the DRM. We then find their orthogonal complement in the weighted score space of (A.2). Appendix~\ref{app:C} uses this complement to compute the part of the unrestricted gradient removed by the DRM.

\hypertarget{drm-tangent-space}{%
\subsection{DRM tangent space}\label{drm-tangent-space}}

\setcounter{lemma}{1}
\begin{lemma}[Tangent space of the two-sample DRM]

Under A1--A4, the tangent space of the two-sample DRM at \((F_0,F_1)\) is \[
{
\mathcal T_{\mathrm{DRM}}
=
\overline{
\left\{
\left(
h,\,
h-E_1h+a^\top q_c
\right):
h\in L_0^2(F_0),\;
a\in\mathbb R^p
\right\}
},
}
\tag{B.1}
\] where \[
q_c(x)=q(x)-E_1q(X).
\]

\end{lemma}
\begin{proof}

Consider a regular path that perturbs the baseline distribution and tilt parameter jointly, \[
dF_{0,t}(x)
=
\{1+t h(x)+o(t)\}\,dF_0(x),
\qquad
E_0h=0,
\] and \[
\beta_t=\beta_0+ta.
\]

The log normalizing constant is \[
A(F_0,\beta)
=
\log E_0e^{\beta^\top q(X)}.
\] Its derivative along the path is \[
\dot A
=
\frac{
E_0[e^{\beta_0^\top q}h]
+
E_0[e^{\beta_0^\top q}a^\top q]
}{
E_0e^{\beta_0^\top q}
}
=
E_1h+a^\top E_1q.
\tag{B.2}
\]

The first-sample score depends only on the baseline perturbation, giving \[
s_0=h.
\]

The target log density also changes with the tilt and its normalizer, \[
\log dF_{1,t}
=
\log dF_{0,t}
+
\beta_t^\top q
-
A(F_{0,t},\beta_t),
\] so the second-sample score is \[
\begin{aligned}
s_1
&=
h+a^\top q-\dot A\\
&=
h-E_1h+a^\top(q-E_1q).
\end{aligned}
\] The derivative of the normalizer centers the target score under its own marginal. Thus every regular DRM score has the form in (B.1).

Conversely, for bounded \(h\) with \(E_0h=0\) and arbitrary \(a\), the paths above generate precisely the displayed score. The upper bound on \(r\) over the common compact support lets bounded mean-zero functions approximate the baseline scores in \(L^2\) under both marginals. Taking the closure gives the full tangent space.

\end{proof}
\hypertarget{orthogonal-complement-of-the-drm-tangent-space}{%
\subsection{Orthogonal complement of the DRM tangent space}\label{orthogonal-complement-of-the-drm-tangent-space}}

\setcounter{lemma}{2}
\begin{lemma}[Exact orthogonal complement]

Under A1--A4, \[
{
\mathcal T_{\mathrm{DRM}}^\perp
=
\left\{
\left(
-\frac{\rho_1}{\rho_0}r u,\,
u
\right):
u\in L^2(F_1),\; E_1u=0,\;
E_1[uq_c]=0
\right\}.
}
\tag{B.3}
\]

\end{lemma}
\begin{proof}

Let \((u_0,u_1)\in\mathcal H\) be orthogonal to every element of (B.1). Then for all admissible \(h,a\), \[
\rho_0E_0[u_0h]
+
\rho_1E_1\left[
u_1
\{h-E_1h+a^\top q_c\}
\right]
=0.
\tag{B.4}
\]

Varying the tilt alone, by setting \(h=0\), gives \[
E_1[u_1q_c]=0.
\tag{B.5}
\] Since \(u_1\in L_0^2(F_1)\), \[
E_1u_1=0.
\tag{B.6}
\]

Varying the baseline alone, by setting \(a=0\), leaves the terms involving the baseline score in (B.4). By (B.6), these satisfy \[
\rho_0E_0[u_0h]
+
\rho_1E_1[u_1h]
=0.
\] Since \(dF_1=r\,dF_0\), \[
E_0[
\{\rho_0u_0+\rho_1ru_1\}h
]=0
\] for every \(h\in L_0^2(F_0)\). Hence \[
\rho_0u_0+\rho_1ru_1
\] is \(F_0\)-a.s. constant. Its \(F_0\)-mean is \[
\rho_0E_0u_0+\rho_1E_1u_1=0,
\] so the constant is zero. Therefore \[
u_0=-\frac{\rho_1}{\rho_0}ru_1.
\] Together with (B.5)--(B.6), this gives the form in (B.3). Conversely, each pair in (B.3) belongs to $\mathcal H$ because $r$ is bounded, and substitution into (B.4) verifies orthogonality.

\end{proof}

\section{Proof of the efficiency identity}\label{app:C}
The unrestricted gradient from Appendix~\ref{app:A} and the orthogonal complement from Appendix~\ref{app:B} determine the DRM efficiency bound. The required projection reduces to a weighted least-squares problem for the contrast between the two EOT potentials.

\hypertarget{canonical-gradient-and-exact-efficiency-gain}{%
\subsection{Canonical gradient and exact efficiency gain}\label{canonical-gradient-and-exact-efficiency-gain}}

Recall the mixture weight \[
\omega(x)=\rho_0+\rho_1r(x),
\tag{C.1}
\] and the unrestricted EOT contrast \[
\zeta(x)
=
\phi_1(x)
-
\frac{\rho_1}{\rho_0}\phi_0(x).
\tag{C.2}
\]

The weighted least-squares coefficients for the contrast are \[
b^\star
=
\arg\min_{b\in\mathbb R^{p+1}}
E_1
\left[
\frac{
\{\zeta(X)-b^\top Q(X)\}^2
}{\omega(X)}
\right].
\tag{C.3}
\] Assumption A4 makes the coefficient matrix invertible, giving \[
{
b^\star
=
M^{-1}
E_1\left[
\frac{Q\zeta}{\omega}
\right].
}
\tag{C.4}
\]

The weighted projection residual is \[
e(x)
=
\zeta(x)-b^{\star\top}Q(x).
\tag{C.5}
\] The weighted normal equations are \[
{
E_1\left[
\frac{Qe}{\omega}
\right]=0.
}
\tag{C.6}
\]

\setcounter{theorem}{0}
\begin{theorem}[Restricted canonical gradient and efficiency identity]

Under A1--A4 and the EOT differentiability conclusion of Lemma~\ref{lem:eot-regularity}, the canonical gradient of \[
\Psi_\varepsilon(F_0,F_1)
\] in the DRM is \[
{
g_{0,\mathrm{DRM}}(x)
=
\frac{\phi_0(x)}{\rho_0}
+
\frac{r(x)e(x)}{\omega(x)},
}
\tag{C.7}
\] \[
{
g_{1,\mathrm{DRM}}(x)
=
\frac{\phi_1(x)}{\rho_1}
-
\frac{\rho_0}{\rho_1}
\frac{e(x)}{\omega(x)}.
}
\tag{C.8}
\]

The corresponding semiparametric efficiency bound is \[
V_{\mathrm{DRM}}
=
\rho_0E_0[g_{0,\mathrm{DRM}}^2]
+
\rho_1E_1[g_{1,\mathrm{DRM}}^2],
\tag{C.9}
\] and satisfies the exact identity \[
{
V_{\mathrm{DRM}}
=
V_{\mathrm{NP}}
-
\frac{\rho_0}{\rho_1}
E_1\left[
\frac{e(X)^2}{\omega(X)}
\right].
}
\tag{C.10}
\]

Equivalently, \[
{
V_{\mathrm{DRM}}
=
V_{\mathrm{NP}}
-
\frac{\rho_0}{\rho_1}
\inf_b
E_1\left[
\frac{
\{\zeta(X)-b^\top Q(X)\}^2
}{\omega(X)}
\right].
}
\tag{C.11}
\]

Consequently, \[
{
V_{\mathrm{DRM}}\le V_{\mathrm{NP}}.
}
\tag{C.12}
\]

Moreover, \[
{
V_{\mathrm{DRM}}=V_{\mathrm{NP}}
\iff
\zeta\in\operatorname{span}(Q)
\quad F_1\text{-a.s.},
}
\tag{C.13}
\] and otherwise \[
V_{\mathrm{DRM}}<V_{\mathrm{NP}}.
\tag{C.14}
\]

\end{theorem}

\paragraph{Proof outline for main-text Theorem 1.}
Appendix~\ref{app:B} derives the scores $(h,h-E_1h+a^\top(q-E_1q))$, with $E_0h=0$, and identifies their orthogonal complement as the set of $k(w)=(-\rho_1rw/\rho_0,w)$ with $E_1(Qw)=0$. We show that the projection of the unrestricted gradient onto this complement is $k(\rho_0e/(\rho_1\omega))$. Subtracting this component gives the DRM canonical gradient. The orthogonal decomposition then yields the exact variance gain as the squared norm of the removed component.
\begin{proof}

For \[
u\in L_0^2(F_1)
\] satisfying \[
E_1[uq_c]=0,
\] write the generic orthogonal direction from Lemma 3 as \[
k(u)
=
\left(
-\frac{\rho_1}{\rho_0}ru,\,
u
\right).
\]

Using (A.3), \[
\begin{aligned}
\langle g_{\mathrm{NP}},k(v)\rangle
&=
\rho_0E_0\left[
\frac{\phi_0}{\rho_0}
\left(-\frac{\rho_1}{\rho_0}rv\right)
\right]
+
\rho_1E_1\left[
\frac{\phi_1}{\rho_1}v
\right]\\
&=
E_1[\zeta v].
\end{aligned}
\tag{C.15}
\]

Also, \[
\begin{aligned}
\langle k(u),k(v)\rangle
&=
\rho_0E_0
\left[
\frac{\rho_1^2}{\rho_0^2}r^2uv
\right]
+
\rho_1E_1[uv]\\
&=
\frac{\rho_1}{\rho_0}
E_1[\omega uv].
\end{aligned}
\tag{C.16}
\]

The projection \(k(u^{\star})\) of \(g_{\mathrm{NP}}\) onto \(\mathcal T_{\mathrm{DRM}}^\perp\) therefore satisfies \[
E_1[\zeta v]
=
\frac{\rho_1}{\rho_0}
E_1[\omega u^{\star}v]
\tag{C.17}
\] for all \(v\) orthogonal to \(1\) and \(q_c\) under \(F_1\).

By (C.3)--(C.6), \[
u^{\star}
=
\frac{\rho_0}{\rho_1}
\frac{e}{\omega}.
\tag{C.18}
\] Hence \[
E_1u^{\star}=0,
\qquad
E_1(u^{\star}q_c)=0,
\]
because the intercept and centered nonintercept components of (C.6) give
\(E_1(e/\omega)=0\) and \(E_1(q_ce/\omega)=0\), respectively. Thus \(u^{\star}\)
is admissible in the orthogonal complement. Its corresponding direction is \[
k(u^{\star})
=
\left(
-\frac{re}{\omega},
\frac{\rho_0}{\rho_1}\frac{e}{\omega}
\right).
\tag{C.19}
\]

Subtracting (C.19) from \(g_{\mathrm{NP}}\) gives (C.7)--(C.8).

The projected components satisfy \[
g_{\mathrm{NP}}
=
g_{\mathrm{DRM}}+k(u^{\star}),
\qquad
g_{\mathrm{DRM}}\perp k(u^{\star}),
\] so \[
V_{\mathrm{NP}}
=
V_{\mathrm{DRM}}
+
\|k(u^{\star})\|_{\mathcal H}^2.
\tag{C.20}
\]

The squared norm of the removed component is \[
\begin{aligned}
\|k(u^{\star})\|_{\mathcal H}^2
&=
\rho_0E_0
\left[
\frac{r^2e^2}{\omega^2}
\right]
+
\rho_1E_1
\left[
\frac{\rho_0^2}{\rho_1^2}
\frac{e^2}{\omega^2}
\right]\\
&=
\frac{\rho_0}{\rho_1}
E_1\left[
\frac{e^2}{\omega}
\right],
\end{aligned}
\] which proves (C.10). Since \(\omega>0\), equality occurs exactly when \(e=0\) \(F_1\)-a.s., i.e.~when \(\zeta\in\operatorname{span}(Q)\). 

\end{proof}
\hypertarget{interpretation}{%
\subsubsection*{Interpretation}\label{interpretation}}

The gain is strict exactly when the EOT contrast has a nonzero weighted projection residual. The gradient component associated with this residual is \[
g_{\mathrm{NP}}-g_{\mathrm{DRM}}
=
\left(
-\frac{re}{\omega},
\frac{\rho_0}{\rho_1}\frac{e}{\omega}
\right),
\] and \[
V_{\mathrm{NP}}-V_{\mathrm{DRM}}
=
\left\|
g_{\mathrm{NP}}-g_{\mathrm{DRM}}
\right\|_{\mathcal H}^2.
\]

\hypertarget{corollaries-and-structural-implications}{%
\subsection{Corollaries and structural implications}\label{corollaries-and-structural-implications}}

\setcounter{corollary}{0}
\begin{corollary}[Explicit matrix form of the efficiency gain]

Let \[
M
=
E_1\left[
\frac{QQ^\top}{\omega}
\right],
\qquad
m
=
E_1\left[
\frac{Q\zeta}{\omega}
\right].
\] Then \[
b^\star=M^{-1}m
\] and \[
E_1\left[
\frac{e^2}{\omega}
\right]
=
E_1\left[
\frac{\zeta^2}{\omega}
\right]
-
m^\top M^{-1}m.
\] Therefore \[
{
V_{\mathrm{DRM}}
=
V_{\mathrm{NP}}
-
\frac{\rho_0}{\rho_1}
\left\{
E_1\left[\frac{\zeta^2}{\omega}\right]
-
m^\top M^{-1}m
\right\}.
}
\tag{C.21}
\]

\end{corollary}
\setcounter{corollary}{1}
\begin{corollary}[Nested basis spaces]

Suppose two correctly specified DRM basis spaces satisfy \[
\operatorname{span}(Q_1)
\subset
\operatorname{span}(Q_2).
\] Then \[
{
V_{\mathrm{DRM}}(Q_1)
\le
V_{\mathrm{DRM}}(Q_2)
\le
V_{\mathrm{NP}}.
}
\tag{C.22}
\]

\end{corollary}
\hypertarget{interpretation-1}{%
\subsubsection*{Interpretation}\label{interpretation-1}}

The comparison fixes the distribution pair and sampling fractions in the same two-sample experiment. Both bases correctly specify the density ratio at that pair. Enlarging the basis can bring more of the EOT contrast into the tilt span, reducing the residual and hence the efficiency gain. The larger span also accommodates more density ratios, which can improve fit at model points where the smaller basis is misspecified.

\setcounter{corollary}{2}
\begin{corollary}[Zero-gain characterization]

There is no first-order gain from the DRM exactly when \[
\phi_1
-
\frac{\rho_1}{\rho_0}\phi_0
\] is representable by the augmented DRM basis.

This gives a natural theoretical explanation for zero-gain examples such as suitably parameterized location-tilt families where the EOT potential contrast lies entirely in the tilt span.

\end{corollary}

\subsection{Full symmetric example}
Suppose $F_0=F_1=F$ is symmetric about zero on a compact subset of the real line, $q(x)=x$, and $\operatorname{Var}_F(X)>0$. Exchange of the marginals and reflection both preserve the cost and the model point. By uniqueness of the normalized transport potentials, the centered potentials satisfy $\phi_0=\phi_1=\phi$, with $\phi$ even and centered. Thus $E_F(X\phi)=0$, $r=\omega=1$, and $\zeta=(1-\rho_1/\rho_0)\phi$ is orthogonal to both $1$ and $x$. The projection therefore has $b^\star=0$, the residual is $e=\zeta$, and the restricted gradients are
\[
g_0=\phi/\rho_0+e=2\phi,\qquad g_1=\phi/\rho_1-(\rho_0/\rho_1)e=2\phi.
\]
Consequently $V_{\rm DRM}=4\operatorname{Var}_F(\phi)$ and
\[
 V_{\rm NP}-V_{\rm DRM}=\left(\frac1{\rho_0}+\frac1{\rho_1}-4\right)\operatorname{Var}_F(\phi)
 =\frac{(\rho_0-\rho_1)^2}{\rho_0\rho_1}\operatorname{Var}_F(\phi).
\]
For this symmetric population and a nonconstant potential, the variance gain is positive exactly when the sampling fractions differ. The displayed dependence on imbalance is specific to this distribution pair and basis.

\section{Efficient DRM-EOT estimator}\label{app:D}
Appendix~\ref{app:C} derives the efficiency bound. To show that the DRM-EOT estimator attains it, we expand the fitted tilt using the actual sampling fractions and obtain a joint process limit for the fitted marginals. Applying the EOT delta method gives a scalar expansion whose influence functions converge to the DRM canonical gradient. Section~\ref{sec:drm-local-regularity} verifies regularity by centering at the moving EOT target along local DRM submodels.

\hypertarget{drm-empirical-likelihood-estimation}{%
\subsection{DRM empirical-likelihood estimation}\label{drm-empirical-likelihood-estimation}}

\hypertarget{dual-empirical-likelihood}{%
\subsubsection{Dual empirical likelihood}\label{dual-empirical-likelihood}}

Let $\Theta=\mathbb R^{p+1}$ be the parameter space of the dual likelihood. The pooled sample below is indexed by sample position, so repeated values contribute once per occurrence. \[
\mathcal Z_n
=
\{X_1,\ldots,X_{n_0},Y_1,\ldots,Y_{n_1}\}.
\]

For \[
r_\theta(x)=e^{\theta^\top Q(x)},
\qquad
\omega_{\theta,n}(x)
=
\rho_{0n}+\rho_{1n}r_\theta(x),
\] the two-sample dual log empirical likelihood is \[
\ell_n(\theta)
=
\sum_{j=1}^{n_1}
\theta^\top Q(Y_j)
-
\sum_{Z\in\mathcal Z_n}
\log\omega_{\theta,n}(Z).
\tag{D.1}
\]

Let \[
\widehat\theta
=
\arg\max_\theta\ell_n(\theta).
\]

The implied baseline probabilities are \[
\widehat p(Z)
=
\frac{1}{n\,\omega_{\widehat\theta,n}(Z)}.
\tag{D.2}
\]

Hence \[
{
\widehat F_0
=
\sum_{Z\in\mathcal Z_n}
\widehat p(Z)\delta_Z,
}
\tag{D.3}
\] and \[
{
\widehat F_1
=
\sum_{Z\in\mathcal Z_n}
r_{\widehat\theta}(Z)
\widehat p(Z)\delta_Z.
}
\tag{D.4}
\]

The proposed EOT estimator is \[
{
\widehat\Psi_\varepsilon^{\mathrm{DRM}}
=
\Psi_\varepsilon(\widehat F_0,\widehat F_1).
}
\tag{D.5}
\]
Write $\mathbb P_n=\rho_{0n}P_{0n}+\rho_{1n}P_{1n}$, $\widehat r=r_{\widehat\theta}$ and $\widehat\omega=\omega_{\widehat\theta,n}$. At an interior solution, the intercept score gives $\mathbb P_n(\widehat r/\widehat\omega)=1$. The identity $\rho_{0n}/\widehat\omega+\rho_{1n}\widehat r/\widehat\omega=1$ gives $\mathbb P_n(1/\widehat\omega)=1$, so both fitted measures are probabilities. We use these fitted-measure formulas when a finite interior maximum exists. Otherwise we use NP-EOT for the transport estimate and set $\widehat\theta=0$ as an auxiliary convention. Lemma 4 shows that this fallback event has probability tending to zero.

\hypertarget{fixed-sample-expansion-of-the-drm-tilt-estimator}{%
\subsection{Fixed-sample expansion of the DRM tilt estimator}\label{fixed-sample-expansion-of-the-drm-tilt-estimator}}

We retain the actual sample fractions in the finite-\(n\) score and information matrix. At these fractions the group score means cancel exactly, giving the centered representation used in the tilt expansion. Define \[
\omega_n(x)
=
\rho_{0n}+\rho_{1n}r(x),
\qquad
\pi_n(x)
=
\frac{\rho_{1n}r(x)}{\omega_n(x)}.
\tag{D.6}
\]

The per-observation score contributions at \(\theta_0\) are \[
s_{0,n}(x)
=
-\pi_n(x)Q(x),
\tag{D.7}
\] and \[
s_{1,n}(x)
=
\{1-\pi_n(x)\}Q(x)
=
\frac{\rho_{0n}}{\omega_n(x)}Q(x).
\tag{D.8}
\]

Define \[
M_n
=
E_1\left[
\frac{QQ^\top}{\omega_n}
\right],
\qquad
W_n
=
\rho_{0n}\rho_{1n}M_n.
\tag{D.9}
\] By A1--A4, \[
M_n\to M,
\qquad
W_n\to
W:=\rho_0\rho_1M,
\] and \(W_n\) is nonsingular for all sufficiently large \(n\).

Let \[
\mathbb G_{0n}f
=
\frac1{\sqrt{n_0}}
\sum_{i=1}^{n_0}\{f(X_i)-E_0f\},
\] \[
\mathbb G_{1n}f
=
\frac1{\sqrt{n_1}}
\sum_{j=1}^{n_1}\{f(Y_j)-E_1f\},
\] and \[
U_n
=
\sqrt{\rho_{0n}}\,
\mathbb G_{0n}s_{0,n}
+
\sqrt{\rho_{1n}}\,
\mathbb G_{1n}s_{1,n}.
\tag{D.10}
\]

The representation in (D.10) equals the normalized \emph{raw} total score because \[
\rho_{0n}E_0s_{0,n}
+
\rho_{1n}E_1s_{1,n}
=
0
\tag{D.11}
\] exactly for every \(n\).

\setcounter{lemma}{3}
\begin{lemma}[Fixed two-sample MELE expansion]

Under A1--A4, the maximum empirical-likelihood estimator (MELE) exists and is unique with probability tending to one. Because \(\theta_0\) is interior, choose a compact neighborhood
\[
\Theta_0\Subset\operatorname{int}(\Theta)
\]
of \(\theta_0\). Then \[
{
\sqrt n(\widehat\theta-\theta_0)
=
W_n^{-1}U_n
+
o_p(1).
}
\tag{D.12}
\]

Moreover, \[
U_n\Rightarrow U,
\tag{D.13}
\] where \[
U
=
\sqrt{\rho_0}\,\mathbb G_0s_0
+
\sqrt{\rho_1}\,\mathbb G_1s_1,
\] \(\mathbb G_0,\mathbb G_1\) are independent Brownian bridges, and \[
s_0(x)
=
-\frac{\rho_1r(x)}{\omega(x)}Q(x),
\qquad
s_1(x)
=
\frac{\rho_0}{\omega(x)}Q(x).
\]

Consequently, \[
{
\sqrt n(\widehat\theta-\theta_0)
\Rightarrow
N(0,W^{-1}-S),
}
\tag{D.14}
\] where \[
S
=
\frac1{\rho_0\rho_1}
e_0e_0^\top,
\qquad
e_0=(1,0,\ldots,0)^\top.
\tag{D.15}
\]

\end{lemma}
\begin{proof}

Continuity of \(Q\) on compact \(K\) and compactness of \(\Theta_0\) imply
\[
\sup_{\theta\in\Theta_0,x\in K}
\{\|Q(x)\|+r_\theta(x)+r_\theta(x)^{-1}\}<\infty.
\]
By A1, \(\rho_{1n}\) is eventually bounded away from zero and one, so \(\omega_{\theta,n}^{-1}\) and the first two tilt derivatives are uniformly bounded on \(\Theta_0\times K\). These are local consequences of A1--A3, not additional model assumptions.

Differentiate the dual log empirical likelihood in (D.1). At \(\theta_0\), \[
\frac1{\sqrt n}\dot\ell_n(\theta_0)
=
\frac1{\sqrt n}
\left\{
\sum_{i=1}^{n_0}s_{0,n}(X_i)
+
\sum_{j=1}^{n_1}s_{1,n}(Y_j)
\right\}
=
U_n,
\tag{D.16}
\] where the last equality follows from (D.11).

For \(\theta\in\Theta_0\), \[
-\frac1n\ddot\ell_n(\theta)
=
\mathbb P_n
\left[
\pi_{\theta,n}(1-\pi_{\theta,n})QQ^\top
\right],
\] with \[
\pi_{\theta,n}
=
\frac{\rho_{1n}r_\theta}
{\rho_{0n}+\rho_{1n}r_\theta}.
\] The class of Hessian entries is uniformly bounded and finite-dimensional-Lipschitz in \(\theta\). A uniform law of large numbers therefore yields local uniform convergence of the normalized Hessian, and at the truth its deterministic finite-\(n\) limit is \(W_n\). Consistency and existence can be established locally as follows. On a sufficiently small fixed ball about \(\theta_0\), the negative normalized Hessian is bounded below by \(\lambda I\) with probability tending to one for some \(\lambda>0\). The normalized score at \(\theta_0\) is \(O_p(n^{-1/2})\). On the sphere of radius \(a\) in this ball, Taylor's inequality gives \[
n^{-1}\{\ell_n(\theta_0+h)-\ell_n(\theta_0)\}
\le \|h\|\|n^{-1}\dot\ell_n(\theta_0)\|-\lambda\|h\|^2/2<0.
\] The maximum over the closed ball is therefore interior, and concavity of the unrestricted dual likelihood makes it a global maximum. Shrinking \(a\) proves consistency; local strict concavity gives uniqueness. Thus consistency and A4 imply that, with probability tending to one, the relevant root of the score equation lies in this neighborhood and the Hessian is nonsingular.

Taylor expansion of \[
\dot\ell_n(\widehat\theta)=0
\] around \(\theta_0\) gives \[
0
=
U_n
-
\left\{
W_n+o_p(1)
\right\}
\sqrt n(\widehat\theta-\theta_0),
\] which proves (D.12).

Because \(s_{g,n}\to s_g\) uniformly on \(K\), the ordinary two-sample triangular-array CLT gives (D.13).

For completeness, put \[
a_n
=
E_1\left[\frac{Q}{\omega_n}\right].
\] Then \[
E_0s_{0,n}=-\rho_{1n}a_n,
\qquad
E_1s_{1,n}=\rho_{0n}a_n.
\tag{D.17}
\] The covariance of \(U_n\) is therefore \[
B_n
=
\rho_{0n}\operatorname{Var}_0(s_{0,n})
+
\rho_{1n}\operatorname{Var}_1(s_{1,n})
=
W_n
-
\rho_{0n}\rho_{1n}a_na_n^\top.
\tag{D.18}
\] Since the first coordinate of \(Q\) is one, \[
W_ne_0
=
\rho_{0n}\rho_{1n}a_n.
\tag{D.19}
\] Hence \[
W_n^{-1}B_nW_n^{-1}
=
W_n^{-1}
-
\frac1{\rho_{0n}\rho_{1n}}
e_0e_0^\top.
\tag{D.20}
\] Taking limits proves (D.14). 

\end{proof}
\hypertarget{remark-9.1.-why-no-rate-for-rho_gn-is-needed}{%
\subsubsection*{\texorpdfstring{Remark D.2.1. Why no rate for \(\rho_{gn}\) is needed}{Remark D.2.1. Why no rate for \textbackslash rho\_\{gn\} is needed}}\label{remark-9.1.-why-no-rate-for-rho_gn-is-needed}}

All finite-\(n\) score, Hessian, mixture, and projection quantities use \(\rho_{gn}\) exactly. The passage from \(\rho_{gn}\) to \(\rho_g\) occurs only after centered triangular-array representations have been obtained. Therefore an \(O(n^{-1/2})\) rate for the sample proportions is unnecessary.

\hypertarget{uniform-drm-empirical-process-theorem}{%
\subsection{Uniform DRM empirical-process theorem}\label{uniform-drm-empirical-process-theorem}}

The tilt expansion controls a finite-dimensional parameter. Applying the EOT delta method also requires a joint limit for both fitted marginals, uniformly over the EOT potential class. We use pointwise measurable or separable versions of the empirical-process classes here and in Appendix~\ref{app:E} for statements in outer probability.

For any bounded measurable \(f\), \[
\widehat F_0f
=
\mathbb P_n
\left[
\frac{f}{\omega_{\widehat\theta,n}}
\right],
\qquad
\widehat F_1f
=
\mathbb P_n
\left[
\frac{r_{\widehat\theta}f}
{\omega_{\widehat\theta,n}}
\right],
\tag{D.21}
\] where \[
\mathbb P_nf
=
\rho_{0n}P_{0n}f+\rho_{1n}P_{1n}f.
\]

At the truth, \[
H_n\left(\frac f{\omega_n}\right)=F_0f,
\qquad
H_n\left(\frac{rf}{\omega_n}\right)=F_1f
\tag{D.22}
\] for every \(n\).

For \(\theta\) near \(\theta_0\), \[
\frac{\partial}{\partial\theta}
\frac1{\omega_{\theta,n}}
=
-\frac{\rho_{1n}r_\theta Q}
{\omega_{\theta,n}^2},
\tag{D.23}
\] and \[
\frac{\partial}{\partial\theta}
\frac{r_\theta}{\omega_{\theta,n}}
=
\frac{\rho_{0n}r_\theta Q}
{\omega_{\theta,n}^2}.
\tag{D.24}
\]

Define \[
c_{f,n}
=
E_1\left[
\frac{fQ}{\omega_n}
\right],
\qquad
c_f
=
E_1\left[
\frac{fQ}{\omega}
\right].
\tag{D.25}
\]

Expanding (D.21) produces products of the test functions with DRM weights and their derivatives. Lemma 5 controls these product classes so that Lemma 6 can linearize both fitted marginals uniformly over the EOT potential class.

\setcounter{lemma}{4}
\begin{lemma}[Bounded DRM multipliers preserve the EOT Donsker class]

Under A1--A3 and the class construction of Lemma~\ref{lem:eot-regularity}, choose the compact neighborhood
\[
\Theta_0\Subset\operatorname{int}(\Theta)
\]
introduced in Lemma 4 and a compact interval \[
\mathcal R\Subset(0,1)
\] containing \(\rho_{1n}\) and \(\rho_1\) for all sufficiently large \(n\). Define \[
\omega_{\theta,\rho}(x)
=
1-\rho+\rho r_\theta(x).
\]

The classes \[
\mathscr G_0
=
\left\{
\frac{f}{\omega_{\theta,\rho}}:
f\in\mathscr F_\varepsilon,\,
\theta\in\Theta_0,\,
\rho\in\mathcal R
\right\},
\tag{D.26}
\] \[
\mathscr G_1
=
\left\{
\frac{r_\theta f}{\omega_{\theta,\rho}}:
f\in\mathscr F_\varepsilon,\,
\theta\in\Theta_0,\,
\rho\in\mathcal R
\right\},
\tag{D.27}
\] together with the coordinatewise first- and second-\(\theta\)-derivative classes generated by (D.23)--(D.24), are \(F_0\)- and \(F_1\)-Donsker and possess uniformly bounded envelopes. The same holds jointly after adjoining the coordinate classes of $-\pi_{\theta,\rho}Q$, $(1-\pi_{\theta,\rho})Q$, and $\pi_{\theta,\rho}(1-\pi_{\theta,\rho})QQ^\top$, where $\pi_{\theta,\rho}=\rho r_\theta/\omega_{\theta,\rho}$.

\end{lemma}
\begin{proof}

By A2--A3 and compactness of \(\Theta_0\times\mathcal R\), \[
r_\theta,\quad
\omega_{\theta,\rho}^{-1},\quad
Q,
\] and all first two derivatives in \(\theta\) of \[
\omega_{\theta,\rho}^{-1},
\qquad
r_\theta\omega_{\theta,\rho}^{-1}
\] are uniformly bounded on \[
K\times\Theta_0\times\mathcal R.
\tag{D.28}
\]

The finite-dimensional multiplier classes \[
\mathscr M_0
=
\{\omega_{\theta,\rho}^{-1}\},
\qquad
\mathscr M_1
=
\{r_\theta\omega_{\theta,\rho}^{-1}\},
\] and their derivative analogues are uniformly bounded Euclidean classes: by the mean-value theorem their \(L^2(P)\) covering numbers are polynomial in \(1/\eta\), uniformly for \(P=F_0,F_1\). Hence they are Donsker.

By Lemma~\ref{lem:eot-regularity}, \(\mathscr F_\varepsilon\) is Donsker and uniformly bounded on \(K\). The map \[
(f,m)\mapsto fm
\] is Lipschitz on bounded rectangles. Standard Donsker permanence for bounded Lipschitz transformations therefore implies that the product classes \[
\mathscr F_\varepsilon\mathscr M_0,
\qquad
\mathscr F_\varepsilon\mathscr M_1,
\] and the finitely many derivative product classes are Donsker. The score and Hessian coordinate classes are also bounded and Lipschitz in the finite-dimensional parameters $(\theta,\rho)$; adjoining them preserves the joint Donsker property. 

\end{proof}
\setcounter{lemma}{5}
\begin{lemma}[Joint uniform linearization and UCLT for the fitted DRM measures]

Under A1--A4 and the technical conclusions of Lemmas~\ref{lem:eot-regularity} and 5, uniformly over \(f\in\mathscr F_\varepsilon\), \[
{
\sqrt n(\widehat F_0f-F_0f)
=
\sqrt{\rho_{0n}}\,
\mathbb G_{0n}\left(\frac f{\omega_n}\right)
+
\sqrt{\rho_{1n}}\,
\mathbb G_{1n}\left(\frac f{\omega_n}\right)
-
\rho_{1n}c_{f,n}^\top W_n^{-1}U_n
+
o_p(1),
}
\tag{D.29}
\] and \[
{
\sqrt n(\widehat F_1f-F_1f)
=
\sqrt{\rho_{0n}}\,
\mathbb G_{0n}\left(\frac {rf}{\omega_n}\right)
+
\sqrt{\rho_{1n}}\,
\mathbb G_{1n}\left(\frac {rf}{\omega_n}\right)
+
\rho_{0n}c_{f,n}^\top W_n^{-1}U_n
+
o_p(1).
}
\tag{D.30}
\]

Consequently, \[
{
\sqrt n
\left(
\widehat F_0-F_0,\,
\widehat F_1-F_1
\right)
\rightsquigarrow
(\mathbb Z_0,\mathbb Z_1)
}
\tag{D.31}
\] in \[
\ell^\infty(\mathscr F_\varepsilon)
\times
\ell^\infty(\mathscr F_\varepsilon),
\] where \[
\mathbb Z_0(f)
=
\sqrt{\rho_0}\,
\mathbb G_0\left(\frac f\omega\right)
+
\sqrt{\rho_1}\,
\mathbb G_1\left(\frac f\omega\right)
-
\rho_1c_f^\top W^{-1}U,
\tag{D.32}
\] and \[
\mathbb Z_1(f)
=
\sqrt{\rho_0}\,
\mathbb G_0\left(\frac {rf}\omega\right)
+
\sqrt{\rho_1}\,
\mathbb G_1\left(\frac {rf}\omega\right)
+
\rho_0c_f^\top W^{-1}U.
\tag{D.33}
\]

\end{lemma}
\begin{proof}

Consider first \[
\widehat F_0f
=
\mathbb P_n
\left[
f\,\omega_{\widehat\theta,n}^{-1}
\right].
\] A second-order Taylor expansion in \(\theta\) gives \[
\begin{aligned}
\widehat F_0f-F_0f
&=
(\mathbb P_n-H_n)
\left(\frac f{\omega_n}\right)\\
&\quad
-
\rho_{1n}
E_1\left[
\frac{fQ}{\omega_n}
\right]^\top
(\widehat\theta-\theta_0)
+
R_{0n}(f)
+
R_{0n}^{\mathrm{emp}}(f).
\end{aligned}
\tag{D.34}
\] Here \(R_{0n}(f)\) is the deterministic second-order Taylor remainder and \(R_{0n}^{\mathrm{emp}}(f)\) is the empirical-process effect of replacing the derivative at \(\theta_0\) by a derivative at an intermediate random parameter.

By Lemma 5, the derivative and second-derivative classes are bounded Donsker classes. Since \[
\|\widehat\theta-\theta_0\|=O_p(n^{-1/2}),
\] uniform boundedness of the second derivative gives \[
\sup_{f\in\mathscr F_\varepsilon}
\sqrt n\,|R_{0n}(f)|
=
O_p\left(
\sqrt n\|\widehat\theta-\theta_0\|^2
\right)
=o_p(1).
\tag{D.35}
\] Stochastic equicontinuity of the derivative Donsker class gives \[
\sup_{f\in\mathscr F_\varepsilon}
\sqrt n\,|R_{0n}^{\mathrm{emp}}(f)|
=o_p(1).
\tag{D.36}
\]

Because \[
\sqrt n(\mathbb P_n-H_n)h
=
\sqrt{\rho_{0n}}\mathbb G_{0n}h
+
\sqrt{\rho_{1n}}\mathbb G_{1n}h,
\tag{D.37}
\] substituting Lemma 4 into (D.34) yields (D.29).

The same argument applied to \[
r_{\widehat\theta}f/\omega_{\widehat\theta,n}
\] and derivative (D.24) yields (D.30).

For weak convergence, Lemma 5 gives joint asymptotic tightness of all empirical processes appearing in (D.29)--(D.30). The multipliers \[
\omega_n^{-1},\quad
r\omega_n^{-1},\quad
s_{g,n},
\quad\text{and}\quad
c_{\cdot,n}
\] converge uniformly to their limiting counterparts because \(\rho_{gn}\to\rho_g\). Thus asymptotic equicontinuity and Slutsky's theorem in \(\ell^\infty(\mathscr F_\varepsilon)\) yield (D.31)--(D.33). 

\end{proof}
\hypertarget{eot-delta-method-for-drm-empirical-measures}{%
\subsection{EOT delta method for DRM empirical measures}\label{eot-delta-method-for-drm-empirical-measures}}

The regularized-OT inference framework of \citet{goldfeld2024regularized} applies to fitted measures with a uniform limit in the EOT potential class. Lemma 6 establishes this limit for the DRM marginals. Lemma~\ref{lem:eot-regularity} gives the tangential differentiability of the EOT functional on the same class.

Hadamard differentiability therefore gives \[
{
\widehat\Psi_\varepsilon^{\mathrm{DRM}}
-
\Psi_\varepsilon
=
(\widehat F_0-F_0)\phi_0
+
(\widehat F_1-F_1)\phi_1
+
o_p(n^{-1/2}).
}
\tag{D.38}
\]

Section~\ref{efficiency-of-the-drm-el-eot-estimator} identifies the influence functions in (D.38) with the finite-fraction versions of the DRM canonical gradient.

\hypertarget{efficiency-of-the-drm-el-eot-estimator}{%
\subsection{Efficiency of the DRM-EOT estimator}\label{efficiency-of-the-drm-el-eot-estimator}}

The influence functions inherit the finite-sample sampling fractions from the fitted marginal expansion. Define the corresponding potential contrast \[
\zeta_n(x)
=
\phi_1(x)
-
\frac{\rho_{1n}}{\rho_{0n}}\phi_0(x),
\tag{D.39}
\] \[
b_n
=
M_n^{-1}
E_1\left[
\frac{Q\zeta_n}{\omega_n}
\right],
\tag{D.40}
\] and \[
e_n(x)
=
\zeta_n(x)-b_n^\top Q(x).
\tag{D.41}
\]

The finite-\(n\) weighted normal equations are \[
E_1\left[
\frac{Qe_n}{\omega_n}
\right]=0.
\tag{D.42}
\] Since \(M_n\to M\) and all multipliers converge uniformly, \[
b_n\to b^\star,
\qquad
e_n\to e
\quad\text{uniformly on }K.
\tag{D.43}
\]

Define the \(n\)-dependent influence functions \[
g_{0,n}(x)
=
\frac{\phi_0(x)}{\rho_{0n}}
+
\frac{r(x)e_n(x)}{\omega_n(x)},
\tag{D.44}
\] and \[
g_{1,n}(x)
=
\frac{\phi_1(x)}{\rho_{1n}}
-
\frac{\rho_{0n}}{\rho_{1n}}
\frac{e_n(x)}{\omega_n(x)}.
\tag{D.45}
\] The intercept normal equation in (D.42) implies \[
E_0g_{0,n}=0,
\qquad
E_1g_{1,n}=0
\tag{D.46}
\] for every \(n\). Moreover, \[
\|g_{g,n}-g_{g,\mathrm{DRM}}\|_{\infty,K}\to0.
\tag{D.47}
\]

\setcounter{theorem}{1}
\begin{theorem}[Asymptotic linearity and semiparametric efficiency]

Under A1--A4 and the technical conclusions established in Lemmas~\ref{lem:eot-regularity} and 4--6, \[
{
\sqrt n
\left(
\widehat\Psi_\varepsilon^{\mathrm{DRM}}
-
\Psi_\varepsilon
\right)
=
\frac1{\sqrt n}
\left\{
\sum_{i=1}^{n_0}g_{0,n}(X_i)
+
\sum_{j=1}^{n_1}g_{1,n}(Y_j)
\right\}
+
o_p(1).
}
\tag{D.48}
\]

Consequently, \[
{
\sqrt n
\left(
\widehat\Psi_\varepsilon^{\mathrm{DRM}}
-
\Psi_\varepsilon
\right)
\Rightarrow
N(0,V_{\mathrm{DRM}}),
}
\tag{D.49}
\] and the DRM empirical-likelihood plug-in estimator is semiparametrically efficient in the correctly specified DRM.

\end{theorem}
\begin{proof}

Set \[
c_{0,n}
=
E_1\left[
\frac{\phi_0Q}{\omega_n}
\right],
\qquad
c_{1,n}
=
E_1\left[
\frac{\phi_1Q}{\omega_n}
\right].
\] Then \[
\rho_{0n}c_{1,n}
-
\rho_{1n}c_{0,n}
=
\rho_{0n}
E_1\left[
\frac{Q\zeta_n}{\omega_n}
\right].
\tag{D.50}
\] Since \[
W_n=\rho_{0n}\rho_{1n}M_n,
\] we have the exact identity \[
{
(\rho_{0n}c_{1,n}-\rho_{1n}c_{0,n})^\top
W_n^{-1}
=
\frac1{\rho_{1n}}b_n^\top.
}
\tag{D.51}
\]

Insert (D.29)--(D.30) into the EOT expansion (D.38) and use (D.51). The contribution of a baseline observation is \[
\widetilde g_{0,n}(x)
=
\frac{\phi_0(x)+r(x)\phi_1(x)-r(x)b_n^\top Q(x)}
{\omega_n(x)}.
\tag{D.52}
\] Using \[
e_n
=
\phi_1
-
\frac{\rho_{1n}}{\rho_{0n}}\phi_0
-
b_n^\top Q,
\] simple algebra yields \[
\widetilde g_{0,n}
=
\frac{\phi_0}{\rho_{0n}}
+
\frac{re_n}{\omega_n}
=
g_{0,n}.
\tag{D.53}
\]

Likewise, the tilted-sample contribution is \[
\widetilde g_{1,n}(x)
=
\frac{
\phi_0(x)+r(x)\phi_1(x)
+
(\rho_{0n}/\rho_{1n})b_n^\top Q(x)
}
{\omega_n(x)},
\tag{D.54}
\] and \[
\widetilde g_{1,n}
=
\frac{\phi_1}{\rho_{1n}}
-
\frac{\rho_{0n}}{\rho_{1n}}
\frac{e_n}{\omega_n}
=
g_{1,n}.
\tag{D.55}
\] This proves (D.48).

The functions \(g_{g,n}\) are uniformly bounded on \(K\), centered groupwise by (D.46), and converge uniformly to \(g_{g,\mathrm{DRM}}\). The Lindeberg condition is therefore automatic, and the triangular-array two-sample CLT yields (D.49).

Finally, Theorem 1 identifies \[
(g_{0,\mathrm{DRM}},g_{1,\mathrm{DRM}})
\] as the canonical gradient of the DRM experiment. The local-regularity argument following the technical derivation verifies that the expansion is stable along differentiable-in-quadratic-mean DRM submodels and is centered at the corresponding moving target. Thus the limiting influence function in (D.48) is the canonical gradient in the regular semiparametric sense, and the plug-in estimator attains the efficiency bound. 

\end{proof}
\subsection{Local regularity of the efficient estimator}\label{sec:drm-local-regularity}
Section~\ref{efficiency-of-the-drm-el-eot-estimator} establishes the CLT at the true distribution pair. Regularity requires the same centered limit along local DRM submodels. Let $\mathsf F_t=(F_{0,t},F_{1,t})$ be a one-dimensional differentiable-in-quadratic-mean DRM submodel on $K$, with score $\xi=(\xi_0,\xi_1)$. For a fixed scalar $a$, take $t=a/\sqrt n$. The fixed two-sample experiment is locally asymptotically normal with information $\|\xi\|_{\mathcal H}^2$. Contiguity transfers the remainder in (D.48) to these alternatives. The leading sum is jointly asymptotically normal with the central sequence, and their covariance is
\[
 \rho_0E_0(g_{0,\mathrm{DRM}}\xi_0)+\rho_1E_1(g_{1,\mathrm{DRM}}\xi_1)
 =\dot\Psi_\varepsilon(\xi_0,\xi_1).
\]
Le Cam's third lemma therefore shifts the leading sum by $a\dot\Psi_\varepsilon(\xi_0,\xi_1)$, which equals the first-order change in $\sqrt n\Psi_\varepsilon(\mathsf F_{a/\sqrt n})$. After centering at the moving target, the limit remains $N(0,V_{\rm DRM})$. On compact $K$, differentiability in quadratic mean implies the corresponding first-order signed-measure expansion on the uniformly bounded potential class, so (A.1) gives the required target derivative. Applying this argument to each direction of a finite-dimensional submodel proves regularity and, together with Theorem 1, semiparametric efficiency.

\section{Variance estimation and bootstrap}\label{app:E}
This appendix develops two routes to inference for the scalar EOT value under the fixed-proportion assumptions of Theorem~2. Section~\ref{feasible-analytic-inference} estimates the variance of the efficient influence function and studentizes the EOT estimator. Section~\ref{stratified-bootstrap} establishes the conditional limit for the stratified refit bootstrap. Both routes use a positive limiting variance for the confidence-interval conclusions.

\hypertarget{feasible-analytic-inference}{%
\subsection{Feasible analytic inference}\label{feasible-analytic-inference}}

We estimate the groupwise variances of the influence functions in Appendix~\ref{app:D}. The plug-in calculation starts with the fitted EOT potentials. Let \[
(\widehat u,\widehat v)
\] be EOT dual potentials for \[
(\widehat F_0,\widehat F_1).
\] Normalize them by \[
\widehat\phi_0
=
\widehat u-\widehat F_0\widehat u,
\qquad
\widehat\phi_1
=
\widehat v-\widehat F_1\widehat v.
\tag{E.1}
\]

Define \[
\widehat r
=
r_{\widehat\theta},
\qquad
\widehat\omega
=
\rho_{0n}+\rho_{1n}\widehat r,
\tag{E.2}
\] \[
\widehat \zeta
=
\widehat\phi_1
-
\frac{\rho_{1n}}{\rho_{0n}}
\widehat\phi_0,
\tag{E.3}
\] \[
\widehat M
=
\widehat F_1
\left[
\frac{QQ^\top}{\widehat\omega}
\right],
\qquad
\widehat m
=
\widehat F_1
\left[
\frac{Q\widehat \zeta}{\widehat\omega}
\right],
\tag{E.4}
\] \[
\widehat b
=
\widehat M^{-1}\widehat m,
\qquad
\widehat e
=
\widehat \zeta-\widehat b^\top Q.
\tag{E.5}
\]

On the exceptional event that the tilt fit fails or $\widehat M$ is singular, set the variance estimate to zero. By Lemmas 4 and 8, this event has probability tending to zero and does not affect consistency. On the complementary event, the estimated efficient influence functions are \[
\widehat g_0
=
\frac{\widehat\phi_0}{\rho_{0n}}
+
\frac{\widehat r\widehat e}{\widehat\omega},
\tag{E.6}
\] and \[
\widehat g_1
=
\frac{\widehat\phi_1}{\rho_{1n}}
-
\frac{\rho_{0n}}{\rho_{1n}}
\frac{\widehat e}{\widehat\omega}.
\tag{E.7}
\]

Define \[
{
\widehat V_{\mathrm{DRM}}
=
\rho_{0n}
P_{0n}
\left[
\{\widehat g_0-P_{0n}\widehat g_0\}^2
\right]
+
\rho_{1n}
P_{1n}
\left[
\{\widehat g_1-P_{1n}\widehat g_1\}^2
\right].
}
\tag{E.8}
\]

\Needspace{15\baselineskip}
\setcounter{lemma}{6}
\begin{lemma}[Uniform stability of fixed-\(\varepsilon\) EOT potentials on compact support]\label{lem:potential-stability}

Let \[
\mu_m\Rightarrow\mu,
\qquad
\nu_m\Rightarrow\nu
\] be probability measures supported on a common compact set \(K\), with \(\operatorname{supp}(\mu)=\operatorname{supp}(\nu)=K\), and let \[
(u_m,v_m)
\] be optimal fixed-\(\varepsilon\) EOT potentials for \((\mu_m,\nu_m)\). Extend the potentials to all of \(K\) through the Schrödinger equations and normalize them by \[
u_m(x_0)=0
\] for a fixed \(x_0\in K\). Let \((u,v)\) be the similarly normalized potentials for \((\mu,\nu)\). Then \[
{
\|u_m-u\|_{\infty,K}
+
\|v_m-v\|_{\infty,K}
\to0.
}
\tag{E.9}
\]
\end{lemma}

\begin{proof}

The Schrödinger equations can be written on all of \(K\) as \[
u_m(x)
=
-\varepsilon
\log
\int_K
\exp\left\{
\frac{v_m(y)-c(x,y)}{\varepsilon}
\right\}
d\nu_m(y),
\tag{E.10}
\] \[
v_m(y)
=
-\varepsilon
\log
\int_K
\exp\left\{
\frac{u_m(x)-c(x,y)}{\varepsilon}
\right\}
d\mu_m(x).
\tag{E.11}
\]

For \(x,x'\in K\), the log-sum-exp map gives \[
|u_m(x)-u_m(x')|
\le
\sup_{y\in K}
|c(x,y)-c(x',y)|.
\tag{E.12}
\] The right-hand side is a common modulus of continuity because \(c\) is uniformly continuous on \(K\times K\). Since \(u_m(x_0)=0\), (E.12) also gives a uniform bound on \(\|u_m\|_\infty\).

Equation (E.11), the uniform bound on \(u_m\), and boundedness of \(c\) on \(K\times K\) imply a uniform bound on \(\|v_m\|_\infty\). The same log-sum-exp argument gives equicontinuity of \(v_m\). Hence \(\{u_m\}\) and \(\{v_m\}\) are uniformly bounded and equicontinuous.

By Arzelà--Ascoli, every subsequence contains a further subsequence converging uniformly to some pair \((u^\circ,v^\circ)\). Along that subsequence, the integrands in (E.10)--(E.11) converge uniformly and are continuous and uniformly bounded. Weak convergence of \(\mu_m,\nu_m\) therefore permits passage to the limit in both equations. Thus \((u^\circ,v^\circ)\) solves the Schrödinger system for \((\mu,\nu)\).

Fixed-\(\varepsilon\) EOT potentials are unique up to the transformation \[
(u,v)\mapsto(u+a,v-a).
\] The normalization \(u(x_0)=0\) removes this degree of freedom, so \[
(u^\circ,v^\circ)=(u,v).
\] Every convergent subsequence has the same limit; hence the full sequence converges uniformly. 

\end{proof}
\setcounter{lemma}{7}
\begin{lemma}[Uniform consistency of the fitted centered potentials and influence functions]

Under A1--A4 and the local regularity conclusions established in Lemmas 4--6, \[
\widehat F_0\Rightarrow F_0,
\qquad
\widehat F_1\Rightarrow F_1
\quad\text{in probability}.
\tag{E.13}
\] After normalization, \[
{
\|\widehat\phi_0-\phi_0\|_{\infty,K}
+
\|\widehat\phi_1-\phi_1\|_{\infty,K}
\to_p0.
}
\tag{E.14}
\] Consequently, \[
\widehat M\to_pM,
\qquad
\widehat m\to_pm,
\qquad
\widehat b\to_pb^\star,
\tag{E.15}
\] and \[
{
\|\widehat g_g-g_{g,\mathrm{DRM}}\|_{\infty,K}
\to_p0,
\qquad g=0,1.
}
\tag{E.16}
\]

\end{lemma}
\begin{proof}

Fix \(f\in C(K)\). From (D.21), \[
\widehat F_0f-F_0f
=
(\mathbb P_n-H_n)
\left(\frac f{\omega_n}\right)
+
o_p(1),
\] because \(\widehat\theta\to_p\theta_0\) and the multiplier is uniformly smooth in \(\theta\). The first term is \(o_p(1)\) by the ordinary two-sample law of large numbers. Hence \[
\widehat F_0f\to_pF_0f.
\] The same argument gives \[
\widehat F_1f\to_pF_1f.
\] A countable convergence-determining subclass of \(C(K)\), together with compactness of \(K\), yields (E.13).

By the subsequence principle, apply Lemma~\ref{lem:potential-stability} pathwise along arbitrary almost-surely convergent subsequences. For potentials normalized at \(x_0\), \[
\|\widehat u-u\|_\infty
+
\|\widehat v-v\|_\infty
\to_p0.
\tag{E.17}
\] Centering preserves uniform convergence because, for example, \[
\begin{aligned}
\left|
\widehat F_0\widehat u-F_0u
\right|
&\le
\|\widehat u-u\|_\infty
+
|(\widehat F_0-F_0)u|\\
&\to_p0.
\end{aligned}
\] Thus (E.14) follows.

Uniform consistency of \(\widehat r,\widehat\omega,\widehat \zeta\), weak consistency of \(\widehat F_1\), and bounded continuity of \(Q\) imply (E.15). A4 and continuity of matrix inversion give \[
\widehat b\to_pb^\star.
\] Substitution into (E.6)--(E.7) yields (E.16). 

\end{proof}
\setcounter{theorem}{2}
\begin{theorem}[Consistent analytic variance estimation]

Under A1--A4, \[
{
\widehat V_{\mathrm{DRM}}
\xrightarrow p
V_{\mathrm{DRM}}.
}
\tag{E.18}
\] If \(V_{\mathrm{DRM}}>0\), then \[
{
\frac{
\sqrt n
(\widehat\Psi_\varepsilon^{\mathrm{DRM}}-\Psi_\varepsilon)
}{
\sqrt{\widehat V_{\mathrm{DRM}}}
}
\Rightarrow
N(0,1).
}
\tag{E.19}
\]

\end{theorem}
\begin{proof}

By Lemma 8, \[
\|\widehat g_g-g_{g,\mathrm{DRM}}\|_\infty\to_p0.
\] Hence \[
P_{gn}\widehat g_g^k
-
P_{gn}g_{g,\mathrm{DRM}}^k
\to_p0,
\qquad k=1,2.
\tag{E.20}
\] The ordinary law of large numbers gives \[
P_{gn}g_{g,\mathrm{DRM}}^k
\to_p
E_gg_{g,\mathrm{DRM}}^k.
\tag{E.21}
\] Since \[
E_gg_{g,\mathrm{DRM}}=0
\] and \(\rho_{gn}\to\rho_g\), substituting (E.20)--(E.21) into (E.8) proves (E.18). Theorem 2, the stated condition \(V_{\mathrm{DRM}}>0\), and Slutsky's theorem give (E.19). 

\end{proof}
\hypertarget{stratified-bootstrap}{%
\subsection{Stratified refit bootstrap}\label{stratified-bootstrap}}

Refitting within each replicate carries the sampling variation of the tilt into the EOT estimate. For each replicate, draw independently within each group. \[
X_1^*,\ldots,X_{n_0}^*
\overset{\mathrm{iid}}{\sim}P_{0n},
\] \[
Y_1^*,\ldots,Y_{n_1}^*
\overset{\mathrm{iid}}{\sim}P_{1n}.
\] Let \[
P_{0n}^*,\qquad P_{1n}^*
\] be the corresponding bootstrap empirical measures and define \[
\mathbb G_{0n}^*
=
\sqrt{n_0}(P_{0n}^*-P_{0n}),
\qquad
\mathbb G_{1n}^*
=
\sqrt{n_1}(P_{1n}^*-P_{1n}).
\tag{E.22}
\]

Recompute the full DRM estimator: \[
\widehat\theta^*
\longrightarrow
(\widehat F_0^*,\widehat F_1^*)
\longrightarrow
\widehat\Psi_\varepsilon^*
=
\Psi_\varepsilon(\widehat F_0^*,\widehat F_1^*).
\]

In this subsection, $\widehat\Psi_\varepsilon=\widehat\Psi_\varepsilon^{\mathrm{DRM}}$, and a superscript $*$ denotes its fully refitted bootstrap analogue. For conditional statements, \(o_{P^*}(1)\) means convergence to zero in bootstrap probability, in outer probability with respect to the original data.

\setcounter{lemma}{8}
\begin{lemma}[Conditional bootstrap expansion of the DRM MELE]

Define \[
\widehat\pi_n(x)
=
\frac{\rho_{1n}\widehat r(x)}
{\widehat\omega(x)},
\] \[
\widehat s_{0,n}
=
-\widehat\pi_nQ,
\qquad
\widehat s_{1,n}
=
(1-\widehat\pi_n)Q,
\tag{E.23}
\] and \[
\widehat W_n
=
\mathbb P_n
\left[
\widehat\pi_n(1-\widehat\pi_n)QQ^\top
\right].
\tag{E.24}
\] Then \[
\widehat W_n\to_pW,
\tag{E.25}
\] and, conditionally on the data, \[
{
\sqrt n(\widehat\theta^*-\widehat\theta)
=
\widehat W_n^{-1}
\widehat U_n^*
+
o_{P^*}(1),
}
\tag{E.26}
\] where \[
\widehat U_n^*
=
\sqrt{\rho_{0n}}\,
\mathbb G_{0n}^*\widehat s_{0,n}
+
\sqrt{\rho_{1n}}\,
\mathbb G_{1n}^*\widehat s_{1,n}.
\tag{E.27}
\]

\end{lemma}
\begin{proof}

The conditional bootstrap score at \(\widehat\theta\) has mean zero because the original score equation is satisfied: \[
\rho_{0n}P_{0n}\widehat s_{0,n}
+
\rho_{1n}P_{1n}\widehat s_{1,n}
=0.
\tag{E.28}
\] Taylor-expand the bootstrap score equation \[
\dot\ell_n^*(\widehat\theta^*)=0
\] around \(\widehat\theta\). Use the compact neighborhood \(\Theta_0\Subset\operatorname{int}(\Theta)\) constructed in Lemma 4. Uniform boundedness of the first two derivatives there, consistency of \(\widehat\theta\), and the conditional law of large numbers give convergence of the bootstrap Hessian to \(\widehat W_n\), uniformly on an \(o_p(1)\) neighborhood. The conditional bootstrap empirical-process theorem for the pointwise measurable/separable bounded Donsker classes of Lemma 5 gives \[
\widehat U_n^*=O_{P^*}(1)
\] in outer probability. The conditional score at \(\widehat\theta\) is \(O_{P^*}(n^{-1/2})\) after division by \(n\). The same sphere argument as in Lemma 4, with the conditional Hessian lower bound, gives an interior bootstrap maximum with conditional probability tending to one and \(\widehat\theta^*-\widehat\theta=O_{P^*}(n^{-1/2})\). Taylor expansion then yields (E.26). On the exceptional event use the same fallback convention as for the original estimator. 

\end{proof}
To pass from the conditional tilt expansion to a joint limit for the fitted marginals, the next lemma replaces fitted coefficients uniformly over the EOT potential class. Section~\ref{sec:bootstrap-coefficient-convergence} establishes the needed uniform convergence.

\setcounter{lemma}{9}
\begin{lemma}[Conditional bootstrap UCLT for the fitted DRM measures]

Define \[
\widehat c_{f,n}
=
\widehat F_1
\left[
\frac{fQ}{\widehat\omega}
\right].
\tag{E.29}
\] Uniformly over \(f\in\mathscr F_\varepsilon\), \[
{
\begin{aligned}
\sqrt n(\widehat F_0^*-\widehat F_0)f
&=
\sqrt{\rho_{0n}}\,
\mathbb G_{0n}^*
\left(\frac f{\widehat\omega}\right)
+
\sqrt{\rho_{1n}}\,
\mathbb G_{1n}^*
\left(\frac f{\widehat\omega}\right)\\
&\quad
-
\rho_{1n}
\widehat c_{f,n}^\top
\widehat W_n^{-1}
\widehat U_n^*
+
o_{P^*}(1),
\end{aligned}
}
\tag{E.30}
\] and \[
{
\begin{aligned}
\sqrt n(\widehat F_1^*-\widehat F_1)f
&=
\sqrt{\rho_{0n}}\,
\mathbb G_{0n}^*
\left(\frac{\widehat r f}{\widehat\omega}\right)
+
\sqrt{\rho_{1n}}\,
\mathbb G_{1n}^*
\left(\frac{\widehat r f}{\widehat\omega}\right)\\
&\quad
+
\rho_{0n}
\widehat c_{f,n}^\top
\widehat W_n^{-1}
\widehat U_n^*
+
o_{P^*}(1).
\end{aligned}
}
\tag{E.31}
\]

Consequently, \[
{
\sqrt n
\left(
\widehat F_0^*-\widehat F_0,\,
\widehat F_1^*-\widehat F_1
\right)
\rightsquigarrow_p
(\mathbb Z_0,\mathbb Z_1)
}
\tag{E.32}
\] in \[
\ell^\infty(\mathscr F_\varepsilon)
\times
\ell^\infty(\mathscr F_\varepsilon).
\]

\end{lemma}
\begin{proof}

Conditionally on the data, expand \[
\mathbb P_n^*
\left[
\frac f{\omega_{\widehat\theta^*,n}}
\right]
-
\mathbb P_n
\left[
\frac f{\omega_{\widehat\theta,n}}
\right]
\] around \(\widehat\theta\). The derivative is \[
-\rho_{1n}
\widehat F_1
\left[
\frac{fQ}{\widehat\omega}
\right]
=
-\rho_{1n}\widehat c_{f,n},
\] and the analogous derivative for the tilted fitted measure is \[
\rho_{0n}\widehat c_{f,n}.
\]

Lemma 5 provides a fixed, uniformly bounded Donsker envelope containing all random multiplier classes with probability tending to one. The standard bootstrap empirical-process theorem therefore gives, conditionally, \[
(\mathbb G_{0n}^*,\mathbb G_{1n}^*)
\rightsquigarrow_p
(\mathbb G_0,\mathbb G_1)
\] jointly on every class appearing in (E.30)--(E.31).

Conditional stochastic equicontinuity permits replacement of the random multipliers \[
\widehat r,\widehat\omega^{-1},
\widehat s_{g,n},
\widehat c_{\cdot,n},
\widehat W_n
\] by their population counterparts because they converge uniformly or in operator norm in probability. The conditional second-order Taylor remainder is \[
O_{P^*}
\left(
\sqrt n\|\widehat\theta^*-\widehat\theta\|^2
\right)
=o_{P^*}(1)
\] in outer probability by Lemma 9. This proves (E.30)--(E.31), and the conditional continuous-mapping theorem gives (E.32). 

\end{proof}
\setcounter{theorem}{3}
\begin{theorem}[Bootstrap consistency]

Under A1--A4, \[
{
\mathcal L^*
\left[
\sqrt n
\left(
\widehat\Psi_\varepsilon^*
-
\widehat\Psi_\varepsilon
\right)
\right]
\rightsquigarrow_p
N(0,V_{\mathrm{DRM}}).
}
\tag{E.33}
\] Let $\Phi$ be the standard normal distribution function. If \(V_{\mathrm{DRM}}>0\), the corresponding uniform distribution-function convergence is \[
{
\sup_{t\in\mathbb R}
\left|
P^*
\left\{
\sqrt n
(\widehat\Psi_\varepsilon^*-\widehat\Psi_\varepsilon)
\le t
\right\}
-
\Phi
\left(
\frac{t}{\sqrt{V_{\mathrm{DRM}}}}
\right)
\right|
\to_p0.
}
\tag{E.34}
\]

\end{theorem}
\begin{proof}

Lemma 10 supplies the conditional UCLT in the same EOT potential class used in the population delta method. The bootstrap functional delta method for fixed-\(\varepsilon\) EOT therefore yields \[
\begin{aligned}
\sqrt n
(\widehat\Psi_\varepsilon^*-\widehat\Psi_\varepsilon)
&=
\sqrt n
(\widehat F_0^*-\widehat F_0)\phi_0\\
&\quad+
\sqrt n
(\widehat F_1^*-\widehat F_1)\phi_1
+
o_{P^*}(1)
\end{aligned}
\tag{E.35}
\] in outer probability.

Combining (E.30)--(E.31) with the same algebra used in Theorem 2 gives \[
{
\sqrt n
(\widehat\Psi_\varepsilon^*-\widehat\Psi_\varepsilon)
=
\sqrt{\rho_{0n}}\,
\mathbb G_{0n}^*g_{0,n}
+
\sqrt{\rho_{1n}}\,
\mathbb G_{1n}^*g_{1,n}
+
o_{P^*}(1).
}
\tag{E.36}
\]

The functions \(g_{g,n}\) are uniformly bounded, groupwise centered, and converge uniformly to \(g_{g,\mathrm{DRM}}\). Conditional bootstrap CLTs therefore imply that the right-hand side of (E.36) converges in probability to \[
N(0,V_{\mathrm{DRM}}).
\] This proves (E.33). Under the stated condition \(V_{\mathrm{DRM}}>0\), the Gaussian limit is continuous and nondegenerate, so the uniform distribution-function statement (E.34) follows. 

\end{proof}
\hypertarget{remark-14.1.-relation-to-existing-bootstrap-results}{%
\subsubsection*{Remark E.2.1. Relation to existing bootstrap results}\label{remark-14.1.-relation-to-existing-bootstrap-results}}

The fixed-group resampling scheme agrees with the stratified refit bootstrap used in current DRM empirical-likelihood theory \citep{zhuang2025bootstrap}, which establishes bootstrap validity for the DRM parameter estimator and distribution-function process. Lemmas 9--10 provide the additional step specific to the present problem: conditional weak convergence on the smooth EOT-potential indexing class. The regularized-OT bootstrap framework of \citet{goldfeld2024regularized} then transfers this process limit through the EOT functional.

\subsection{Uniform coefficient convergence in the bootstrap}\label{sec:bootstrap-coefficient-convergence}
The class $\mathscr F_\varepsilon$ is totally bounded in the uniform norm on $K$. Multiplying it by a fixed bounded continuous coordinate of $Q/\omega$ preserves total boundedness. A finite-net argument, weak consistency of $\widehat F_1$, and uniform consistency of the random multipliers therefore imply
\[
 \sup_{f\in\mathscr F_\varepsilon}\|\widehat c_{f,n}-c_f\|\to_p0.
\]
This gives the uniform coefficient replacement used in Lemma 10. The joint limiting Gaussian process has a closed linear support contained in the feasible tangent cone of $\mathcal P(K)^2$. The bootstrap delta method applies on this subspace.

\section{Local Misspecification Sensitivity}\label{app:F}
The mean shift under local misspecification measures how much the estimator's response to a target perturbation differs from the change in the true EOT value. The weighted projection residual from Appendix~\ref{app:C} determines this difference for each contiguous departure.

\hypertarget{local-misspecification-and-the-efficiencyrobustness-trade-off}{%
\subsection{Mean Shift under Contiguous Target Departures}\label{local-misspecification-and-the-efficiencyrobustness-trade-off}}

For sufficiently large \(n\), consider the following contiguous departure from the correctly specified DRM at $(F_0,F_1)$. \[
dF_{1,n}(x)
=
\left\{
1+\frac{\eta}{\sqrt n}\delta(x)
\right\}
dF_1(x),
\tag{F.1}
\] where \[
E_1\delta=0
\] and \(\delta\) is bounded. The mean-zero condition keeps the perturbed law normalized.

Write $\Psi_{\varepsilon,n}=\Psi_\varepsilon(F_0,F_{1,n})$. On the root-\(n\) scale, the first-order change in the true EOT target is \[
\eta E_1[\phi_1\delta].
\tag{F.2}
\]

Under Le Cam's local asymptotic theory, the DRM-efficient estimator responds according to \[
\eta\rho_1E_1[g_{1,\mathrm{DRM}}\delta].
\tag{F.3}
\]

Subtracting the change in the target from the estimator's response gives the following bias coefficient per unit of \(\eta\). \[
\begin{aligned}
B_\delta
&=
\rho_1E_1[g_{1,\mathrm{DRM}}\delta]
-
E_1[\phi_1\delta]\\
&=
-\rho_0
E_1\left[
\frac{e\delta}{\omega}
\right].
\end{aligned}
\tag{F.4}
\]

Hence the local limit is \[
{
\sqrt n
\left(
\widehat\Psi_\varepsilon^{\mathrm{DRM}}
-
\Psi_{\varepsilon,n}
\right)
\Rightarrow
N\left(
-\eta\rho_0E_1\left[\frac{e\delta}{\omega}\right],
V_{\mathrm{DRM}}
\right).
}
\tag{F.5}
\]

The bias coefficient in (F.4) depends on the weighted pairing of the residual \(e\) with the perturbation. The asymptotic variance gain depends on the weighted squared norm of that same residual. \[
V_{\mathrm{NP}}-V_{\mathrm{DRM}}
=
\frac{\rho_0}{\rho_1}
E_1\left[
\frac{e^2}{\omega}
\right]
\tag{F.6}
\]

By Cauchy--Schwarz, \[
{
|B_\delta|
\le
\sqrt{
\rho_0\rho_1
\{V_{\mathrm{NP}}-V_{\mathrm{DRM}}\}
}
\left\{
E_1\left[
\frac{\delta^2}{\omega}
\right]
\right\}^{1/2}.
}
\tag{F.7}
\]

For a fixed value of $E_1(\delta^2/\omega)$, the upper bound in (F.7) increases with the variance gain. Equality in Cauchy--Schwarz requires $\delta$ to be proportional to $e$, and that direction is admissible only if it satisfies $E_1\delta=0$. The bound can therefore be loose over admissible mean-zero perturbations. Ordering worst-case sensitivity across nested models would require additional analysis. For a given departure, (F.4) gives the signed bias coefficient, which can vanish even when the variance gain is positive. The calculation concerns contiguous departures from a correctly specified model. Fixed misspecification requires a separate analysis.

\section{Additional synthetic experiments}\label{app:G}
At fixed positive sampling proportions, we compare the asymptotic variance with its Monte Carlo counterpart, check Wald and stratified refit bootstrap coverage, and trace the mean shift along a specified local departure. These experiments use the fixed-proportion regime of Appendices~\ref{app:A}--\ref{app:F}.

The common design uses $K=[-1,1]$, $F_0=F_1=\operatorname{Unif}[-1,1]$, $c(x,y)=(x-y)^2/2$, and $\varepsilon=1$. Both $Q_1=(1,x)^\top$ and $Q_2=(1,x,x^2)^\top$ correctly specify the density ratio $r\equiv1$. Population quantities use 800-node Gauss--Legendre quadrature; the baseline EOT target is $0.2799288123$. Simulations use seed 20260912. Within each bootstrap sample, we resample separately within the two groups, preserve the observed group sizes, refit the DRM, and use the basic interval based on the centered root-$n$ statistic.

At this allocation $(\rho_0,\rho_1)=(0.2,0.8)$, the contrast $\zeta$ is a nonconstant even function. The linear coordinate of $Q_1$ is orthogonal to it under the weighted inner product. The quadratic coordinate of $Q_2$ leaves only a small residual. The archived 800-node quadrature gives $V_{\rm NP}-V_{\rm DRM}(Q_2)=2.7109506\times10^{-7}$, corresponding to $E_1(e^2/\omega)=1.0843802\times10^{-6}$ and a relative variance reduction of $4.1192309\times10^{-6}$. The residual gain is a quadrature approximation to the variance difference.

\begin{table}[ht]
\centering
\caption{Fixed-proportion synthetic experiment. The quadrature variance is compared with the scaled Monte Carlo variance, with Wald coverage reported for the two DRM fits.}\label{tab:ncheck}
\begin{tabular}{rrlrrr}
\toprule
$n$ & Model & Theory $V$ & MC $n\operatorname{Var}$ & Relative error & Wald coverage\\
\midrule
200 & DRM-EOT $Q_1$ & 0.04212 & 0.04837 & 14.85\% & 96.50\%\\
200 & DRM-EOT $Q_2$ & 0.06581 & 0.07569 & 15.00\% & 93.17\%\\
200 & NP-EOT & 0.06581 & 0.07567 & 14.99\% & --\\
400 & DRM-EOT $Q_1$ & 0.04212 & 0.04444 & 5.52\% & 96.05\%\\
400 & DRM-EOT $Q_2$ & 0.06581 & 0.07028 & 6.79\% & 95.15\%\\
400 & NP-EOT & 0.06581 & 0.07030 & 6.82\% & --\\
800 & DRM-EOT $Q_1$ & 0.04212 & 0.04377 & 3.92\% & 95.67\%\\
800 & DRM-EOT $Q_2$ & 0.06581 & 0.06730 & 2.26\% & 94.67\%\\
800 & NP-EOT & 0.06581 & 0.06729 & 2.25\% & --\\
\bottomrule
\end{tabular}
\par\smallskip\footnotesize
Notes: The $n=200$ and $n=800$ rows use 600 replications; $n=400$ reuses the 2,000-replication main experiment. The group proportions are $(0.2,0.8)$ throughout.
\end{table}

Under both DRM fits and NP-EOT, the scaled Monte Carlo variance moves closer to the first-order bound from $n=200$ to $n=800$.

\begin{figure}[ht]
\centering
\includegraphics[width=.76\textwidth]{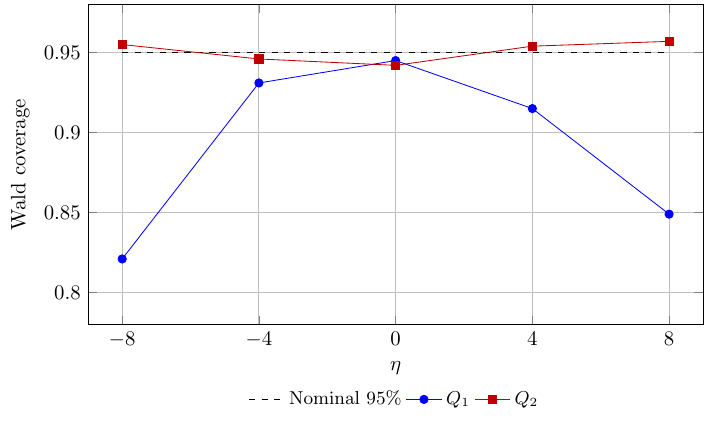}
\caption{Wald coverage under the quadratic contiguous departure.  The restrictive $Q_1$ fit loses coverage as $|\eta|$ increases, while $Q_2$ has near-nominal coverage in this direction.}\label{fig:tradeoff-coverage}
\end{figure}

For local misspecification, we use $\delta(x)=3x^2/2-1/2$ and $\eta\in\{-8,-4,0,4,8\}$, with $n=400$ ($n_0=80$, $n_1=320$) and 1,000 replications per value. The population targets are $0.2615983$, $0.2707571$, $0.2799288$, $0.2891134$, and $0.2983108$, respectively. Quadrature gives $B_\delta(Q_1)=0.0275345$. For $Q_2$, write $\delta=c^\top Q_2$ with $c=(-1/2,0,3/2)^\top$. The projection normal equations give $B_\delta(Q_2)=-\rho_0 c^\top E_1(Q_2e/\omega)=0$ exactly. The small quadrature residual is roundoff. This zero concerns the first-order local mean shift in the specified direction. At $\eta=0$, the mean scaled errors in the simulation are 0.0251 for $Q_1$ and 0.0225 for $Q_2$. All numerical fits succeeded.

\subsection{Local mean shift and baseline bootstrap coverage}
\begin{figure}[H]\centering
\includegraphics[width=.76\textwidth]{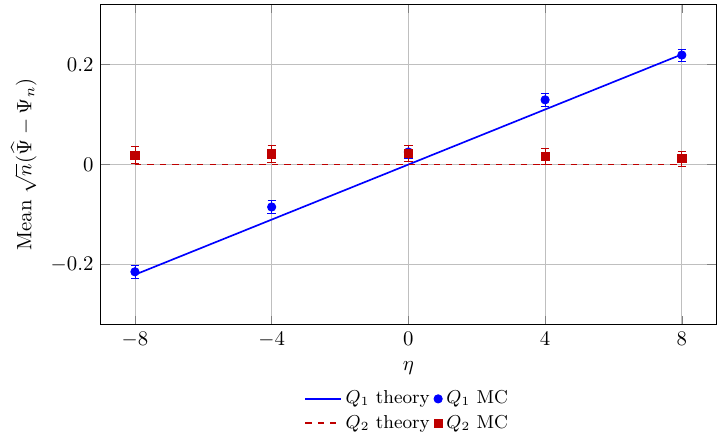}
\caption{Quadratic local departure in the fixed-proportion synthetic design. Points show the mean scaled error with 95\% pointwise Monte Carlo error bars. Lines show $\eta B_\delta$.}\label{fig:tradeoff-bias}
\end{figure}

For the quadratic perturbation $\delta(x)=3x^2/2-1/2$, the first-order mean shift is $\eta B_\delta$, with $B_\delta(Q_1)$ estimated at $0.0275345$ by quadrature and $B_\delta(Q_2)=0$ by projection orthogonality. Figure~\ref{fig:tradeoff-bias} compares the directional prediction with the average scaled errors in the finite-sample experiment.
\begin{table}[H]\centering\small
\caption{Coverage and mean length of the 95\% confidence intervals under the correctly specified synthetic baseline. Wald summaries use 2,000 repetitions; bootstrap summaries use 1,000 repetitions, each with 499 stratified refit bootstrap replicates.}
\begin{tabular}{lrrrr}\toprule
Method & Wald coverage & Wald length & Bootstrap coverage & Bootstrap length\\\midrule
DRM-EOT $Q_1$ & .9605 & .04284 & .9710 & .04398\\
DRM-EOT $Q_2$ & .9515 & .05216 & .9550 & .05296\\\bottomrule
\end{tabular}\end{table}
\begin{table}[H]\centering\small
\caption{Complete local-departure results, 1,000 repetitions per $\eta$. Mean denotes scaled estimation error; SE is its Monte Carlo standard error.}
\begin{tabular}{rrrrrrr}\toprule
$\eta$ & $Q_1$ mean & SE & Wald cov. & $Q_2$ mean & SE & Wald cov.\\\midrule
-8 & -.21453 & .00667 & .821 & .01848 & .00862 & .955\\
-4 & -.08487 & .00666 & .931 & .02116 & .00853 & .946\\
0 & .02511 & .00697 & .945 & .02254 & .00849 & .942\\
4 & .12938 & .00665 & .915 & .01620 & .00825 & .954\\
8 & .21904 & .00625 & .849 & .01186 & .00775 & .957\\\bottomrule
\end{tabular}\end{table}

\clearpage

\section{OfficeHome setup and full results}\label{app:H}
The exact-model experiments evaluate the predicted efficiency gain on fixed OfficeHome feature geometries. The natural experiments use the DRM as a working model for EOT estimation between observed domain pools.

\paragraph{Protocol index.}
\begin{itemize}\setlength{\itemsep}{0pt}
\item \textbf{Exact-model scaling.} A prescribed tilt on each $d=16$ base geometry gives an exact DRM. We sample with replacement at fixed group proportions, use $R=100$ repetitions per cell, and average the variance ratios across geometries (Table~\ref{tab-scaling-full}).
\item \textbf{Confirmation.} The same exact-model design uses $R=500$ repetitions per cell. At $n=8000$ these include the scaling draws. The intervals resample paired estimates within each geometry (Table~\ref{tab-tail-domains}).
\item \textbf{Natural baseline.} For $d=16$ and $q(z)=z$, we draw $R=100$ subsamples without replacement per task and size. The primary analysis uses paired successes and takes the ratio of task-averaged RMSEs (Tables~\ref{tab-natural-summary}--\ref{tab-natural-full}).
\item \textbf{Fixed-geometry diagnostic.} For $d=64$ and $p=8,16,32,64$, each basis uses the same subsamples without replacement. The primary analysis includes all draws with NP-EOT fallback and averages the task-level RMSE ratios (Section~\ref{sec:e1-fixed}).
\end{itemize}

\subsection{Representation and sampling protocol}
We use the 15,588-image OfficeHome dataset, which contains 65 categories and four domains: Art (A), Clipart (C), Product (P), and Real World (R). Frozen ImageNet ResNet-50 penultimate representations are extracted once. For each target domain, PCA is fit using only the other three domains. The stored PCA-64 representation is truncated to its first $d$ coordinates and L2-normalized again. This is equivalent to truncating the original scores before normalization. The natural $d=16$ analysis and the original joint comparisons at $d=32,64$ share the PCA construction. The joint comparison changes the cost geometry and the tilt dimension together. Labels do not enter fitting, EOT, or sample selection. After projection and L2 normalization, the observed and semi-synthetic representations lie on the unit sphere, providing a bounded representation domain consistent with the compact-support setting used in the theory. The theory conditions on the fixed representation map and does not account for representation-learning or PCA uncertainty.

The natural baseline uses all 12 directed tasks, $n_S=1000$, $n_T\in\{50,100,200,400,800\}$, and $R=100$. Each repetition samples without replacement within each group. All resampling comparisons condition on the fixed feature pools. Given those pools, the reference is the fixed EOT cost between the two complete empirical marginals under the same map and regularization. Bias and RMSE describe performance for this empirical target. If the pools are treated as random, the reference and subsamples share the same observations and are dependent unconditionally. The main natural results use $\varepsilon=0.1$. The regularization grid is $0.05,0.1,0.2,0.5$.

The bias, variance, and RMSE summaries in Tables~\ref{tab-natural-summary}--\ref{tab-complexity-full} and~\ref{tab-epsilon-full} use paired successes, meaning draws on which both procedures succeed and yield finite estimates. Fit success rates are calculated over all scheduled draws. Table~\ref{tab-fallback-operational} also reports the results with NP-EOT fallback for cells with failed DRM fits. The fixed-geometry study in Section~\ref{sec:e1-fixed} uses all draws with NP-EOT fallback as its primary comparison.

\paragraph{Exact-model sampling.}
For each semi-synthetic base domain, we use the frozen PCA fit on the other three domains. We retain the base domain's points, truncate to $d=16$, and normalize the truncated vectors to unit length. The $N_K$ resulting support points $z_i$ have masses $P_0(\{z_i\})=1/N_K$ and $P_1(\{z_i\})=P_0(\{z_i\})\exp(\alpha+\tau z_{i1})$. The tilt direction is fixed at the first retained PCA coordinate. Calibration targets ESS/$N_K=0.8,0.5$, where effective sample size (ESS) is the reciprocal sum of squared population target masses. ESS/$N_K$ measures mass concentration on the fixed support, and $\rho$ gives the allocation of observations between the groups. Exact-identity errors are of floating-point order. The target is the finite-support population EOT, computed to solver tolerance. Sampling is i.i.d. with replacement. Multinomial counts combine repeated support points without changing marginal masses. The oracle bound uses population potentials and the weighted projection in main-text Theorem 1. Every reported oracle ratio is below one.

Fixed-proportion scaling uses $\rho=(0.8,0.2)$, $n=500,1000,2000,4000,8000$, and $R=100$. The confirmation run uses $n=8000,16000$ and $R=500$ per base domain and ESS cell. At $n=8000$, it extends each scaling cell by 400 repetitions. The archived replication indices, seeds, and paired NP-EOT and DRM-EOT estimates match on the first 100 repetitions in all eight geometry-by-ESS cells. The comparison conditions on the four evaluated geometries.

\subsection{Numerical accuracy and Monte Carlo uncertainty}
The cost is $\|z-z'\|^2/2$, and the reported scalar EOT objective includes the product-marginal KL term. DRM empirical likelihood uses an unpenalized exponential tilt with intercept. The natural complexity runs use score tolerance $10^{-8}$ and Sinkhorn tolerance $10^{-9}$. Execution manifests record iteration limits and convergence checks. These finite stopping tolerances define the numerical protocol. The asymptotic $o_p(n^{-1/2})$ condition concerns objective error as sample size grows.

\paragraph{Monte Carlo intervals.}
For each confirmation cell, the intervals use the paired finite estimates on draws where both methods succeed. Each of 2,000 resamples draws pairs with replacement within each fixed geometry, keeping the two method estimates from a repetition together. For an interval on the mean ratio, each resample first computes the geometry-level variance ratios, then their unweighted mean. The 2.5\% and 97.5\% quantiles give the pointwise 95\% Monte Carlo interval. These intervals quantify simulation uncertainty conditional on the four geometries. Main-text Theorem 3, proved by Theorems 3--4 in this supplement, concerns refitting the estimator to resampled observations. Reduction is confirmed only when the upper Monte Carlo endpoint is below one. Domain-specific intervals can include or exceed one even when the interval for the mean ratio lies below one. Execution seeds are 20260916 for OfficeHome and 20260912 for the controlled synthetic experiment.

\paragraph{Aggregation.}
All method ratios divide DRM-EOT by NP-EOT. The main natural and epsilon-sensitivity RMSE ratios divide the two methods' unweighted mean RMSEs across tasks. The dimension diagnostics and fixed-geometry comparison average the within-task RMSE ratios. Table~\ref{tab-natural-summary} reports both natural RMSE aggregations. Variance ratios are averaged across fixed tasks or geometries, as appropriate. Absolute reference-bias summaries average absolute task-level biases.
\paragraph{Computational cost audit.}
We benchmark the unchanged exact-model implementation on the Art geometry ($N_K=2427$, $d=16$, ESS/$N_K=0.5$), with $\rho=(0.8,0.2)$ and $\varepsilon=0.1$, using an Apple M1 with 16 GB RAM, CPU float64, and one OpenBLAS 0.3.31 thread. Each size uses three fresh sequential processes with the original seed derivation and solver settings. The table reports median milliseconds and median process-peak MiB ($2^{20}$ bytes). Transport times exclude kernel construction; DRM total includes kernel construction, tilt fitting, fitted masses, transport, and analytic variance. Loading, representation processing, calibration, and sampling are excluded from timing. Peak memory is the operating system's process-lifetime high-water mark, including setup and the sequential NP/DRM pipeline with its shared kernel; it is not method-specific memory or single-matrix storage.
\begin{center}
\begingroup\small
\begin{tabular}{rrrrrrr}
\toprule
$n$ & NP transport & Tilt fit & DRM transport & Kernel & DRM total & Peak MiB\\
\midrule
500 & 1.69 & 0.58 & 6.72 & 3.83 & 10.32 & 157.0\\
2000 & 14.07 & 0.90 & 58.83 & 15.82 & 76.78 & 195.6\\
8000 & 69.65 & 1.30 & 165.29 & 52.05 & 219.04 & 292.6\\
16000 & 105.89 & 1.23 & 172.39 & 56.54 & 229.49 & 284.5\\
\bottomrule
\end{tabular}
\endgroup
\end{center}
Duplicate-support compression is retained: the shared kernel has dimension $U\times U$, where $U$ ranges over $448$--$458$, $1347$--$1378$, $2326$--$2334$, and $2417$--$2420$ at the four sizes. NP uses the positive-mass source/target submatrix; DRM uses the pooled support. These timings therefore do not measure expanded $n\times n$ computations. Without compression, the DRM/NP matrix-entry ratio is $1/(\rho_0\rho_1)=6.25$ at this allocation and $1050^2/(1000\cdot50)=22.05$ in the natural target-scarce setting. One float64 matrix with $m$ entries occupies $8m$ bytes, distinct from process peak memory. One refit-bootstrap replicate at $n=2000$ takes $39.74$ ms including kernel construction; $B$ replicates repeat the estimator work roughly $B$ times, apart from reusable preprocessing, rather than following an exact linear runtime law.

\subsection{Natural OfficeHome results}
\begin{figure}[t]
\centering\includegraphics{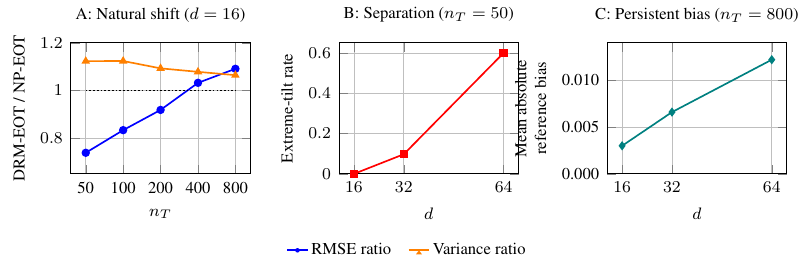}
\caption{Natural OfficeHome comparisons under a working DRM, conditional on the observed feature pools. A: paired-success DRM-EOT/NP-EOT ratio of task-averaged reference RMSE and mean task-level variance ratio ($d=16$). B: extreme-tilt frequency $\widehat{\Pr}\!\left(\max_i|\widehat\theta^\top Q(Z_i)|>30\right)$ over all scheduled draws at $n_T=50$. C: mean absolute DRM reference bias at $n_T=800$ on paired successes. In B and C, $d$ changes both representation geometry and the dimension of the tilt basis.}\label{fig:natural}
\end{figure}

The paired-success natural baseline at $d=16$ gives ratios of task-averaged reference RMSE of $0.739,\allowbreak 0.834,\allowbreak 0.918,\allowbreak 1.032,\allowbreak 1.091$ (Figure~\ref{fig:natural}). DRM-EOT improves reference RMSE on $12/12$, $12/12$, and $9/12$ tasks at the three smallest target sizes. The mean task-level variance ratios are $1.123,\allowbreak 1.124,\allowbreak 1.093,\allowbreak 1.078,\allowbreak 1.064$, all above one. At $n_T=50$, mean absolute reference bias falls from $0.0219$ to $0.0078$. The pooled fit improves reference RMSE at small target sizes through lower reference bias. At larger target sizes the RMSE ratio exceeds one.

\subsection{Representation geometry, tilt dimension, and misspecification}
In the original dimension comparison, $d=16,32,64$ changes the features entering the transport cost and the number of basis functions in $q(z)=z$. The comparison is a joint sensitivity diagnostic. At $n_T=50$, extreme-tilt frequency rises from $0$ to $0.098$ to $0.600$, mean tilt norm from $10.3$ to $35.35$ to $853.8$, and fit success falls from $1.000$ to $0.999$ to $0.958$. These patterns are consistent with near-separation; the threshold is diagnostic. At $d=64$ and $n_T=50$, applying NP-EOT fallback changes the RMSE ratio from $0.718$ to $0.725$ and the variance ratio from $1.525$ to $1.536$. At $n_T=800$, extreme tilts nearly disappear, yet mean absolute DRM reference bias rises from $0.0030$ to $0.0066$ to $0.0122$. Persistent reference bias is consistent with misspecification. Task-mean RMSE ratios at $n_T=50$ are $0.741,0.662,0.718$, so higher dimension can also lower small-sample RMSE. These joint comparisons do not isolate the effect of tilt dimension.

At $n_T=50$, the RMSE ratios for $\varepsilon=0.05,0.10,0.20,0.50$ are $0.566,0.739,0.938,1.001$, respectively. In this grid the two estimators have more similar RMSE as regularization increases.

Tables~\ref{tab-population-full}--\ref{tab-tail-domains} give the exact-model populations, oracle comparisons, scaling results and confirmation intervals. Tables~\ref{tab-natural-summary}--\ref{tab-epsilon-full} give natural-model diagnostics, including the fallback comparison in Table~\ref{tab-fallback-operational}. Section~\ref{sec:e1-fixed} reports the separate fixed-geometry protocol and full results.

\Needspace{10\baselineskip}\subsection{Exact finite-support populations}
\begingroup\small
\begin{longtable}{lrrrrrr}
\caption{Exact finite-support populations. The target is the population EOT functional; $r_{\min}=\min_i P_1(\{z_i\})/P_0(\{z_i\})$ and $r_{\max}=\max_i P_1(\{z_i\})/P_0(\{z_i\})$ are the density-ratio extrema.}\label{tab-population-full}\\
\toprule
Base & ESS/$N_K$ & $N_K$ & $\tau$ & Target & $r_{\min}$ & $r_{\max}$\\\midrule
\endfirsthead
\toprule
Base & ESS/$N_K$ & $N_K$ & $\tau$ & Target & $r_{\min}$ & $r_{\max}$\\\midrule
\endhead
\bottomrule\endfoot
A & 0.8 & 2427 & 1.399 & 0.53428 & 0.385 & 3.836\\
A & 0.5 & 2427 & 2.472 & 0.58532 & 0.160 & 9.247\\
C & 0.8 & 4365 & 1.657 & 0.51728 & 0.192 & 2.572\\
C & 0.5 & 4365 & 3.638 & 0.55697 & 0.020 & 5.979\\
P & 0.8 & 4439 & 1.525 & 0.55091 & 0.302 & 3.353\\
P & 0.5 & 4439 & 2.871 & 0.60719 & 0.087 & 8.097\\
R & 0.8 & 4357 & 1.415 & 0.53899 & 0.428 & 4.332\\
R & 0.5 & 4357 & 2.417 & 0.59516 & 0.205 & 10.704\\
\end{longtable}\endgroup

\Needspace{10\baselineskip}\subsection{Oracle bounds by geometry and allocation}
\begingroup\small
\begin{longtable}{lrrrrrr}
\caption{Oracle asymptotic variances for the exact OfficeHome models, by fixed geometry and group allocation. The ratio is DRM-EOT variance divided by NP-EOT variance.}\label{tab-oracle-full}\\
\toprule
Base & ESS/$N_K$ & $n_0$ & $n_1$ & $V_{\rm NP}$ & $V_{\rm DRM}$ & Ratio\\\midrule
\endfirsthead
\toprule
Base & ESS/$N_K$ & $n_0$ & $n_1$ & $V_{\rm NP}$ & $V_{\rm DRM}$ & Ratio\\\midrule
\endhead
\bottomrule\endfoot
A & 0.8 & 1000 & 50 & 0.33797 & 0.31611 & 0.935\\
C & 0.8 & 1000 & 50 & 0.22795 & 0.19244 & 0.844\\
P & 0.8 & 1000 & 50 & 0.40768 & 0.36802 & 0.903\\
R & 0.8 & 1000 & 50 & 0.44525 & 0.41675 & 0.936\\
A & 0.8 & 1000 & 100 & 0.18024 & 0.16936 & 0.940\\
C & 0.8 & 1000 & 100 & 0.13258 & 0.11436 & 0.863\\
P & 0.8 & 1000 & 100 & 0.21990 & 0.19985 & 0.909\\
R & 0.8 & 1000 & 100 & 0.23617 & 0.22214 & 0.941\\
A & 0.8 & 1000 & 200 & 0.10182 & 0.09633 & 0.946\\
C & 0.8 & 1000 & 200 & 0.08669 & 0.07699 & 0.888\\
P & 0.8 & 1000 & 200 & 0.12687 & 0.11646 & 0.918\\
R & 0.8 & 1000 & 200 & 0.13203 & 0.12510 & 0.948\\
A & 0.8 & 1000 & 400 & 0.06348 & 0.06052 & 0.953\\
C & 0.8 & 1000 & 400 & 0.06734 & 0.06165 & 0.916\\
P & 0.8 & 1000 & 400 & 0.08209 & 0.07619 & 0.928\\
R & 0.8 & 1000 & 400 & 0.08076 & 0.07713 & 0.955\\
A & 0.8 & 1000 & 800 & 0.04606 & 0.04406 & 0.957\\
C & 0.8 & 1000 & 800 & 0.06485 & 0.06073 & 0.936\\
P & 0.8 & 1000 & 800 & 0.06317 & 0.05896 & 0.933\\
R & 0.8 & 1000 & 800 & 0.05673 & 0.05433 & 0.958\\
A & 0.5 & 1000 & 50 & 0.62094 & 0.59341 & 0.956\\
C & 0.5 & 1000 & 50 & 0.24166 & 0.21611 & 0.894\\
P & 0.5 & 1000 & 50 & 0.64724 & 0.60341 & 0.932\\
R & 0.5 & 1000 & 50 & 0.79777 & 0.76213 & 0.955\\
A & 0.5 & 1000 & 100 & 0.33447 & 0.32054 & 0.958\\
C & 0.5 & 1000 & 100 & 0.15052 & 0.13727 & 0.912\\
P & 0.5 & 1000 & 100 & 0.35368 & 0.33113 & 0.936\\
R & 0.5 & 1000 & 100 & 0.42641 & 0.40857 & 0.958\\
A & 0.5 & 1000 & 200 & 0.19249 & 0.18519 & 0.962\\
C & 0.5 & 1000 & 200 & 0.10822 & 0.10095 & 0.933\\
P & 0.5 & 1000 & 200 & 0.20890 & 0.19672 & 0.942\\
R & 0.5 & 1000 & 200 & 0.24189 & 0.23269 & 0.962\\
A & 0.5 & 1000 & 400 & 0.12401 & 0.11979 & 0.966\\
C & 0.5 & 1000 & 400 & 0.09360 & 0.08904 & 0.951\\
P & 0.5 & 1000 & 400 & 0.14050 & 0.13310 & 0.947\\
R & 0.5 & 1000 & 400 & 0.15196 & 0.14671 & 0.965\\
A & 0.5 & 1000 & 800 & 0.09480 & 0.09174 & 0.968\\
C & 0.5 & 1000 & 800 & 0.09934 & 0.09568 & 0.963\\
P & 0.5 & 1000 & 800 & 0.11430 & 0.10857 & 0.950\\
R & 0.5 & 1000 & 800 & 0.11164 & 0.10784 & 0.966\\
\end{longtable}\endgroup

\Needspace{10\baselineskip}\subsection{Full fixed-proportion scaling}
\begingroup\small
\begin{longtable}{lrrrrrr}
\caption{Exact-model OfficeHome scaling with 100 paired repetitions per geometry and cell. The reported DRM-EOT/NP-EOT ratios average the geometry-specific ratios over four fixed geometries. The scaled variances are compared with their corresponding oracle bounds.}\label{tab-scaling-full}\\
\toprule
ESS/$N_K$ & $n$ & Var. ratio & Oracle & $n\mathrm{Var}_{\rm NP}/V_{\rm NP}$ & $n\mathrm{Var}_{\rm DRM}/V_{\rm DRM}$ & RMSE ratio\\\midrule
\endfirsthead
\toprule
ESS/$N_K$ & $n$ & Var. ratio & Oracle & $n\mathrm{Var}_{\rm NP}/V_{\rm NP}$ & $n\mathrm{Var}_{\rm DRM}/V_{\rm DRM}$ & RMSE ratio\\\midrule
\endhead
\bottomrule\endfoot
0.5 & 500 & 1.207 & 0.952 & 0.880 & 1.119 & 0.825\\
0.5 & 1000 & 1.107 & 0.952 & 0.910 & 1.052 & 0.900\\
0.5 & 2000 & 1.118 & 0.952 & 0.991 & 1.161 & 0.933\\
0.5 & 4000 & 1.018 & 0.952 & 1.102 & 1.179 & 0.924\\
0.5 & 8000 & 0.986 & 0.952 & 0.925 & 0.961 & 0.917\\
0.8 & 500 & 1.288 & 0.929 & 0.921 & 1.272 & 0.707\\
0.8 & 1000 & 1.217 & 0.929 & 0.884 & 1.157 & 0.810\\
0.8 & 2000 & 1.039 & 0.929 & 0.937 & 1.044 & 0.824\\
0.8 & 4000 & 1.056 & 0.929 & 1.167 & 1.321 & 0.826\\
0.8 & 8000 & 1.021 & 0.929 & 1.000 & 1.100 & 0.876\\
\end{longtable}\endgroup

\Needspace{10\baselineskip}\subsection{Domain-specific high-R confirmation}
\begingroup\small
\begin{longtable}{lrrrrrr}
\caption{Exact-model OfficeHome confirmation by fixed geometry. The empirical DRM-EOT/NP-EOT ratios use paired finite estimates from the 500 scheduled repetitions. Intervals are 95\% pointwise Monte Carlo intervals within geometry.}\label{tab-tail-domains}\\
\toprule
Base & ESS/$N_K$ & $n$ & Var. ratio & 95\% MC interval & Oracle & RMSE ratio\\\midrule
\endfirsthead
\toprule
Base & ESS/$N_K$ & $n$ & Var. ratio & 95\% MC interval & Oracle & RMSE ratio\\\midrule
\endhead
\bottomrule\endfoot
A & 0.8 & 8000 & 1.050 & $[1.011,1.093]$ & 0.949 & 0.835\\
C & 0.8 & 8000 & 0.959 & $[0.904,1.017]$ & 0.897 & 0.903\\
P & 0.8 & 8000 & 0.962 & $[0.914,1.013]$ & 0.921 & 0.860\\
R & 0.8 & 8000 & 1.025 & $[0.980,1.071]$ & 0.950 & 0.869\\
A & 0.8 & 16000 & 0.995 & $[0.948,1.043]$ & 0.949 & 0.893\\
C & 0.8 & 16000 & 0.963 & $[0.907,1.021]$ & 0.897 & 0.954\\
P & 0.8 & 16000 & 0.952 & $[0.911,0.995]$ & 0.921 & 0.885\\
R & 0.8 & 16000 & 0.973 & $[0.938,1.013]$ & 0.950 & 0.911\\
A & 0.5 & 8000 & 0.991 & $[0.960,1.021]$ & 0.963 & 0.946\\
C & 0.5 & 8000 & 0.983 & $[0.943,1.022]$ & 0.939 & 0.971\\
P & 0.5 & 8000 & 0.949 & $[0.910,0.989]$ & 0.944 & 0.891\\
R & 0.5 & 8000 & 0.989 & $[0.954,1.024]$ & 0.963 & 0.967\\
A & 0.5 & 16000 & 0.987 & $[0.952,1.020]$ & 0.963 & 0.965\\
C & 0.5 & 16000 & 0.910 & $[0.872,0.949]$ & 0.939 & 0.944\\
P & 0.5 & 16000 & 0.936 & $[0.898,0.976]$ & 0.944 & 0.953\\
R & 0.5 & 16000 & 0.968 & $[0.936,1.001]$ & 0.963 & 0.969\\
\end{longtable}\endgroup

\Needspace{15\baselineskip}\subsection{Natural result aggregation}
\begingroup\small
\begin{longtable}{lrrrrr}
\caption{Natural OfficeHome baseline at $d=16$, using finite paired successes in the 12 directed tasks. DRM-EOT/NP-EOT ratios condition on the fixed feature pools. The two RMSE ratio columns report the ratio of task-averaged reference RMSEs and the mean of the within-task ratios.}\label{tab-natural-summary}\\
\toprule
$n_T$ & Mean NP RMSE & Mean DRM RMSE & Ratio of means & Mean of ratios & Mean var. ratio\\\midrule
\endfirsthead
\toprule
$n_T$ & Mean NP RMSE & Mean DRM RMSE & Ratio of means & Mean of ratios & Mean var. ratio\\\midrule
\endhead
\bottomrule\endfoot
50 & 0.0289 & 0.0213 & 0.739 & 0.741 & 1.123\\
100 & 0.0185 & 0.0154 & 0.834 & 0.833 & 1.124\\
200 & 0.0121 & 0.0111 & 0.918 & 0.921 & 1.093\\
400 & 0.0084 & 0.0087 & 1.032 & 1.041 & 1.078\\
800 & 0.0067 & 0.0073 & 1.091 & 1.106 & 1.064\\
\end{longtable}\endgroup

\Needspace{10\baselineskip}\subsection{Complete natural task-level results}
\begingroup\small\renewcommand{\arraystretch}{0.94}\setlength{\LTpre}{6pt}\setlength{\LTpost}{6pt}
\begin{longtable}{lrrrrrrr}
\caption{Natural OfficeHome at $d=16$, by directed task and target sample size. Bias and RMSE use the full-pool reference. The DRM-EOT/NP-EOT variance ratio uses paired successes, as do the error summaries. Success is relative to all scheduled draws.}\label{tab-natural-full}\\
\toprule
Task & $n_T$ & NP bias & DRM bias & NP RMSE & DRM RMSE & Var. ratio & Success\\\midrule
\endfirsthead
\toprule
Task & $n_T$ & NP bias & DRM bias & NP RMSE & DRM RMSE & Var. ratio & Success\\\midrule
\endhead
\bottomrule\endfoot
C$\to$A & 50 & 0.0122 & 0.0043 & 0.0256 & 0.0237 & 1.069 & 1.000\\
P$\to$A & 50 & 0.0207 & 0.0101 & 0.0268 & 0.0207 & 1.129 & 1.000\\
R$\to$A & 50 & 0.0227 & -0.0007 & 0.0272 & 0.0155 & 1.068 & 1.000\\
A$\to$C & 50 & 0.0142 & 0.0053 & 0.0230 & 0.0201 & 1.158 & 1.000\\
P$\to$C & 50 & 0.0249 & 0.0124 & 0.0303 & 0.0233 & 1.295 & 1.000\\
R$\to$C & 50 & 0.0234 & 0.0177 & 0.0311 & 0.0279 & 1.112 & 1.000\\
A$\to$P & 50 & 0.0271 & 0.0148 & 0.0346 & 0.0264 & 1.032 & 1.000\\
C$\to$P & 50 & 0.0197 & 0.0021 & 0.0275 & 0.0193 & 1.010 & 1.000\\
R$\to$P & 50 & 0.0281 & 0.0024 & 0.0333 & 0.0197 & 1.189 & 1.000\\
A$\to$R & 50 & 0.0195 & -0.0021 & 0.0241 & 0.0150 & 1.093 & 1.000\\
C$\to$R & 50 & 0.0230 & 0.0137 & 0.0316 & 0.0266 & 1.098 & 1.000\\
P$\to$R & 50 & 0.0278 & 0.0081 & 0.0311 & 0.0175 & 1.226 & 1.000\\
C$\to$A & 100 & 0.0064 & 0.0015 & 0.0175 & 0.0170 & 1.084 & 1.000\\
P$\to$A & 100 & 0.0100 & 0.0022 & 0.0169 & 0.0144 & 1.100 & 1.000\\
R$\to$A & 100 & 0.0131 & -0.0044 & 0.0163 & 0.0112 & 1.122 & 1.000\\
A$\to$C & 100 & 0.0087 & 0.0038 & 0.0174 & 0.0162 & 1.088 & 1.000\\
P$\to$C & 100 & 0.0147 & 0.0072 & 0.0208 & 0.0170 & 1.087 & 1.000\\
R$\to$C & 100 & 0.0118 & 0.0100 & 0.0203 & 0.0199 & 1.077 & 1.000\\
A$\to$P & 100 & 0.0123 & 0.0030 & 0.0191 & 0.0156 & 1.097 & 1.000\\
C$\to$P & 100 & 0.0125 & -0.0003 & 0.0209 & 0.0170 & 1.031 & 1.000\\
R$\to$P & 100 & 0.0155 & -0.0028 & 0.0198 & 0.0140 & 1.258 & 1.000\\
A$\to$R & 100 & 0.0110 & -0.0061 & 0.0154 & 0.0130 & 1.145 & 1.000\\
C$\to$R & 100 & 0.0076 & 0.0016 & 0.0188 & 0.0177 & 1.052 & 1.000\\
P$\to$R & 100 & 0.0155 & 0.0011 & 0.0185 & 0.0117 & 1.343 & 1.000\\
C$\to$A & 200 & 0.0040 & 0.0000 & 0.0117 & 0.0113 & 1.053 & 1.000\\
P$\to$A & 200 & 0.0060 & -0.0000 & 0.0107 & 0.0097 & 1.211 & 1.000\\
R$\to$A & 200 & 0.0070 & -0.0070 & 0.0098 & 0.0101 & 1.135 & 1.000\\
A$\to$C & 200 & 0.0083 & 0.0048 & 0.0130 & 0.0116 & 1.125 & 1.000\\
P$\to$C & 200 & 0.0065 & 0.0028 & 0.0114 & 0.0104 & 1.152 & 1.000\\
R$\to$C & 200 & 0.0069 & 0.0080 & 0.0134 & 0.0144 & 1.097 & 1.000\\
A$\to$P & 200 & 0.0058 & -0.0012 & 0.0117 & 0.0105 & 1.063 & 1.000\\
C$\to$P & 200 & 0.0076 & -0.0025 & 0.0141 & 0.0127 & 1.089 & 1.000\\
R$\to$P & 200 & 0.0094 & -0.0041 & 0.0123 & 0.0096 & 1.180 & 1.000\\
A$\to$R & 200 & 0.0058 & -0.0074 & 0.0103 & 0.0111 & 0.924 & 1.000\\
C$\to$R & 200 & 0.0076 & 0.0038 & 0.0146 & 0.0131 & 1.018 & 1.000\\
P$\to$R & 200 & 0.0088 & -0.0026 & 0.0118 & 0.0085 & 1.069 & 1.000\\
C$\to$A & 400 & 0.0030 & -0.0000 & 0.0098 & 0.0094 & 1.021 & 1.000\\
P$\to$A & 400 & 0.0023 & -0.0031 & 0.0078 & 0.0084 & 1.109 & 1.000\\
R$\to$A & 400 & 0.0044 & -0.0067 & 0.0076 & 0.0094 & 1.111 & 1.000\\
A$\to$C & 400 & 0.0027 & -0.0004 & 0.0086 & 0.0082 & 1.010 & 1.000\\
P$\to$C & 400 & 0.0052 & 0.0035 & 0.0095 & 0.0092 & 1.141 & 1.000\\
R$\to$C & 400 & 0.0040 & 0.0057 & 0.0095 & 0.0103 & 1.006 & 1.000\\
A$\to$P & 400 & 0.0046 & -0.0008 & 0.0083 & 0.0071 & 1.041 & 1.000\\
C$\to$P & 400 & 0.0036 & -0.0043 & 0.0087 & 0.0092 & 1.074 & 1.000\\
R$\to$P & 400 & 0.0050 & -0.0044 & 0.0077 & 0.0076 & 1.106 & 1.000\\
A$\to$R & 400 & 0.0034 & -0.0066 & 0.0065 & 0.0087 & 1.050 & 1.000\\
C$\to$R & 400 & 0.0037 & 0.0005 & 0.0095 & 0.0089 & 1.048 & 1.000\\
P$\to$R & 400 & 0.0051 & -0.0044 & 0.0078 & 0.0079 & 1.225 & 1.000\\
C$\to$A & 800 & 0.0019 & -0.0007 & 0.0088 & 0.0087 & 1.005 & 1.000\\
P$\to$A & 800 & 0.0018 & -0.0028 & 0.0064 & 0.0068 & 1.006 & 1.000\\
R$\to$A & 800 & 0.0033 & -0.0052 & 0.0054 & 0.0070 & 1.217 & 1.000\\
A$\to$C & 800 & 0.0014 & -0.0014 & 0.0070 & 0.0071 & 1.053 & 1.000\\
P$\to$C & 800 & 0.0025 & 0.0014 & 0.0070 & 0.0070 & 1.092 & 1.000\\
R$\to$C & 800 & 0.0029 & 0.0050 & 0.0072 & 0.0083 & 0.988 & 1.000\\
A$\to$P & 800 & 0.0018 & -0.0025 & 0.0060 & 0.0064 & 1.059 & 1.000\\
C$\to$P & 800 & 0.0013 & -0.0053 & 0.0069 & 0.0088 & 1.063 & 1.000\\
R$\to$P & 800 & 0.0033 & -0.0028 & 0.0061 & 0.0059 & 1.003 & 1.000\\
A$\to$R & 800 & 0.0023 & -0.0054 & 0.0049 & 0.0071 & 1.129 & 1.000\\
C$\to$R & 800 & 0.0022 & -0.0001 & 0.0083 & 0.0081 & 1.020 & 1.000\\
P$\to$R & 800 & 0.0038 & -0.0039 & 0.0058 & 0.0061 & 1.136 & 1.000\\
\end{longtable}\endgroup

\Needspace{10\baselineskip}\subsection{Complete complexity diagnostics}
\begingroup\small\renewcommand{\arraystretch}{0.94}\setlength{\LTpre}{6pt}\setlength{\LTpost}{6pt}
\begin{longtable}{lrrrrrrr}
\caption{Natural OfficeHome with joint changes to geometry and tilt dimension. DRM-EOT/NP-EOT error and variance ratios average within-task ratios on paired successes. Success and extreme-tilt rates use all scheduled draws. The mean tilt norm uses successful DRM fits.}\label{tab-complexity-full}\\
\toprule
$d$ & $n_T$ & RMSE ratio & Var. ratio & Success & Extreme tilt & $|\mathrm{bias}|$ & $\|\hat\theta\|$\\\midrule
\endfirsthead
\toprule
$d$ & $n_T$ & RMSE ratio & Var. ratio & Success & Extreme tilt & $|\mathrm{bias}|$ & $\|\hat\theta\|$\\\midrule
\endhead
\bottomrule\endfoot
16 & 50 & 0.741 & 1.123 & 1.000 & 0.000 & 0.0078 & 10.30\\
16 & 100 & 0.833 & 1.124 & 1.000 & 0.000 & 0.0037 & 9.35\\
16 & 200 & 0.921 & 1.093 & 1.000 & 0.000 & 0.0037 & 8.89\\
16 & 800 & 1.106 & 1.064 & 1.000 & 0.000 & 0.0030 & 8.43\\
32 & 50 & 0.662 & 1.255 & 0.999 & 0.098 & 0.0076 & 35.35\\
32 & 100 & 0.787 & 1.150 & 1.000 & 0.012 & 0.0062 & 18.11\\
32 & 200 & 1.037 & 1.153 & 1.000 & 0.001 & 0.0067 & 16.05\\
32 & 800 & 1.637 & 1.080 & 1.000 & 0.000 & 0.0066 & 14.32\\
64 & 50 & 0.718 & 1.525 & 0.958 & 0.600 & 0.0122 & 853.80\\
64 & 100 & 0.757 & 1.326 & 0.993 & 0.313 & 0.0067 & 152.57\\
64 & 200 & 1.219 & 1.312 & 1.000 & 0.049 & 0.0099 & 31.76\\
64 & 800 & 2.546 & 1.177 & 1.000 & 0.000 & 0.0122 & 24.13\\
\end{longtable}\endgroup

\begingroup\small\setlength{\tabcolsep}{3pt}
\begin{longtable}{rrrrrrrr}
\caption{Natural OfficeHome fallback diagnostic for cells with DRM fit failures. Paired columns use successful pairs. Fallback columns use all repetitions within the fixed feature pools and substitute matched NP-EOT estimates for failed DRM fits. Ratios average the task-level DRM-EOT/NP-EOT ratios.}\label{tab-fallback-operational}\\
\toprule
$d$ & $n_T$ & Success & \multicolumn{2}{c}{RMSE ratio} & \multicolumn{2}{c}{Variance ratio} & Extreme tilt\\
 & & & Paired & Fallback & Paired & Fallback & \\
\midrule
\endfirsthead
\toprule
$d$ & $n_T$ & Success & \multicolumn{2}{c}{RMSE ratio} & \multicolumn{2}{c}{Variance ratio} & Extreme tilt\\
 & & & Paired & Fallback & Paired & Fallback & \\
\midrule
\endhead
\bottomrule\endfoot
32 & 50 & 0.9992 & 0.662090 & 0.662144 & 1.254926 & 1.255011 & 0.0975\\
64 & 50 & 0.9583 & 0.717718 & 0.725298 & 1.524860 & 1.536201 & 0.6000\\
64 & 100 & 0.9933 & 0.757109 & 0.759524 & 1.325702 & 1.330941 & 0.3133\\
\end{longtable}
\endgroup
Across all 14,400 complexity repetitions, every NP-EOT estimate was valid and 59 DRM fits failed; the fallback evaluation retains every repetition and uses its corresponding NP-EOT estimate for a failed DRM fit. The fallback-inclusive results are nearly identical to the paired-success diagnostics, so conditioning on successful fits does not drive the reported small-target RMSE improvement. Variance remains higher than NP-EOT in these cells.

\Needspace{10\baselineskip}\subsection{Full regularization sensitivity}
\begingroup\small\renewcommand{\arraystretch}{0.94}\setlength{\LTpre}{6pt}\setlength{\LTpost}{6pt}
\begin{longtable}{lrrrrrr}
\caption{Natural OfficeHome at $d=16$, conditional on the fixed feature pools. Error summaries use paired successes. RMSE ratios divide task-averaged DRM-EOT RMSE by task-averaged NP-EOT RMSE. Variance ratios average the within-task DRM-EOT/NP-EOT ratios.}\label{tab-epsilon-full}\\
\toprule
$\varepsilon$ & $n_T$ & RMSE ratio & Better tasks & Var. ratio & $|\mathrm{bias}|$ & Success\\\midrule
\endfirsthead
\toprule
$\varepsilon$ & $n_T$ & RMSE ratio & Better tasks & Var. ratio & $|\mathrm{bias}|$ & Success\\\midrule
\endhead
\bottomrule\endfoot
0.05 & 50 & 0.566 & 12/12 & 1.334 & 0.0158 & 1.000\\
0.05 & 100 & 0.629 & 12/12 & 1.299 & 0.0082 & 1.000\\
0.05 & 200 & 0.728 & 12/12 & 1.226 & 0.0069 & 1.000\\
0.05 & 400 & 0.897 & 9/12 & 1.177 & 0.0061 & 1.000\\
0.05 & 800 & 1.071 & 7/12 & 1.164 & 0.0055 & 1.000\\
0.10 & 50 & 0.739 & 12/12 & 1.123 & 0.0078 & 1.000\\
0.10 & 100 & 0.834 & 12/12 & 1.124 & 0.0037 & 1.000\\
0.10 & 200 & 0.918 & 9/12 & 1.093 & 0.0037 & 1.000\\
0.10 & 400 & 1.032 & 6/12 & 1.078 & 0.0034 & 1.000\\
0.10 & 800 & 1.091 & 4/12 & 1.064 & 0.0030 & 1.000\\
0.20 & 50 & 0.938 & 10/12 & 1.008 & 0.0038 & 1.000\\
0.20 & 100 & 0.974 & 8/12 & 1.018 & 0.0018 & 1.000\\
0.20 & 200 & 0.988 & 8/12 & 1.022 & 0.0017 & 1.000\\
0.20 & 400 & 1.015 & 5/12 & 1.030 & 0.0011 & 1.000\\
0.20 & 800 & 1.019 & 5/12 & 1.017 & 0.0010 & 1.000\\
0.50 & 50 & 1.001 & 6/12 & 0.996 & 0.0022 & 1.000\\
0.50 & 100 & 1.003 & 6/12 & 0.998 & 0.0011 & 1.000\\
0.50 & 200 & 1.004 & 6/12 & 1.006 & 0.0010 & 1.000\\
0.50 & 400 & 1.007 & 5/12 & 1.010 & 0.0004 & 1.000\\
0.50 & 800 & 1.005 & 5/12 & 1.004 & 0.0004 & 1.000\\
\end{longtable}\endgroup

\Needspace{8\baselineskip}
\subsection{Varying the tilt basis at fixed transport geometry}\label{sec:e1-fixed}
For each of the 12 directed OfficeHome tasks we keep the target-excluded PCA-64 representation, full-vector L2 normalization, cost $\|z-z'\|^2/2$, and $\varepsilon=0.1$ fixed. The unpenalized tilt uses $Q_p(z)=(1,z_1,\ldots,z_p)^\top$ for $p=8,16,32,64$. We do not renormalize the truncated basis or use it in the transport cost. The reference is unrestricted EOT between the two complete empirical feature pools at the same fixed cost. Within each task and target size we take 100 draws, subsampling without replacement within each domain, with $n_S=1000$ and $n_T=50,200,800$. A draw reuses its samples and NP-EOT estimate across all four bases. Sampling follows the archived 64-dimensional experiment with base seed 20260916.

The fallback policy evaluates all draws. It uses DRM-EOT when the fit and transport computation succeed and the matched NP-EOT otherwise. The run retained all 14,400 estimates over 3,600 distinct draws; the 53 fallbacks are three fits at $(n_T,p)=(50,32)$ and 50 at $(50,64)$ with runaway parameters. No NP-EOT estimate failed. All 3,600 common $p=64$ draws match the original $d=64$ archive to numerical tolerance, including the 50 fallback cases. The largest absolute difference in the successful DRM estimates or the fallback estimates is $4.8\times 10^{-15}$. Fit and complete-estimate success rates coincide. The extreme-tilt indicator flags finite $\max_i|\widehat{\theta}^{\top}Q_p(z_i)|>30$, including finite last iterates on failed fits.

For each task, compute RMSE, sample variance, and the absolute mean error relative to the full-sample empirical reference. We average the task ratios or the task absolute biases with equal weights. Brackets give 95\% percentile Monte Carlo intervals from 2,000 resamples of repetition indices within each fixed task and target size, pairing all four bases. These intervals quantify simulation uncertainty conditional on the fixed pools and task list. They are distinct from the stratified refit bootstrap for inference in Theorem~3. The companion CSVs also report signed bias and the paired-success and common-success comparisons.

\begingroup\small\setlength{\tabcolsep}{3pt}
\begin{longtable}{rrllll}
\caption{Natural OfficeHome with fixed PCA-64 geometry, using all draws with NP-EOT fallback. The table averages the within-task DRM-EOT/NP-EOT ratios across 12 tasks. Absolute bias uses the full-pool reference. Brackets give 95\% pointwise Monte Carlo intervals conditional on the pools and tasks.}\label{tab-e1-fixed-errors}\\
\toprule
$n_T$ & $p$ & RMSE ratio & Variance ratio & DRM $|\mathrm{bias}|$ & NP $|\mathrm{bias}|$\\\midrule
\endfirsthead
\toprule
$n_T$ & $p$ & RMSE ratio & Variance ratio & DRM $|\mathrm{bias}|$ & NP $|\mathrm{bias}|$\\\midrule
\endhead
\bottomrule\endfoot
50 & 8 & 1.470 [1.414, 1.536] & 2.159 [2.017, 2.328] & 0.0379 [0.0369, 0.0389] & 0.0269 [0.0263, 0.0276]\\
50 & 16 & 0.830 [0.792, 0.869] & 1.806 [1.710, 1.923] & 0.0169 [0.0160, 0.0178] & 0.0269 [0.0263, 0.0276]\\
50 & 32 & 0.617 [0.592, 0.641] & 1.568 [1.488, 1.659] & 0.0083 [0.0074, 0.0092] & 0.0269 [0.0263, 0.0276]\\
50 & 64 & 0.725 [0.711, 0.738] & 1.536 [1.469, 1.617] & 0.0126 [0.0118, 0.0134] & 0.0269 [0.0263, 0.0276]\\
200 & 8 & 3.726 [3.571, 3.917] & 1.818 [1.704, 1.963] & 0.0396 [0.0391, 0.0401] & 0.0086 [0.0083, 0.0090]\\
200 & 16 & 2.227 [2.133, 2.342] & 1.513 [1.438, 1.609] & 0.0233 [0.0228, 0.0237] & 0.0086 [0.0083, 0.0090]\\
200 & 32 & 1.628 [1.558, 1.717] & 1.351 [1.292, 1.424] & 0.0157 [0.0153, 0.0161] & 0.0086 [0.0083, 0.0090]\\
200 & 64 & 1.219 [1.167, 1.286] & 1.312 [1.265, 1.366] & 0.0099 [0.0095, 0.0104] & 0.0086 [0.0083, 0.0090]\\
800 & 8 & 6.611 [6.364, 6.977] & 1.336 [1.266, 1.425] & 0.0341 [0.0338, 0.0344] & 0.0028 [0.0025, 0.0030]\\
800 & 16 & 4.076 [3.919, 4.305] & 1.229 [1.178, 1.290] & 0.0207 [0.0205, 0.0210] & 0.0028 [0.0025, 0.0030]\\
800 & 32 & 3.143 [3.016, 3.330] & 1.192 [1.147, 1.244] & 0.0156 [0.0153, 0.0159] & 0.0028 [0.0025, 0.0030]\\
800 & 64 & 2.546 [2.440, 2.697] & 1.177 [1.145, 1.214] & 0.0122 [0.0119, 0.0125] & 0.0028 [0.0025, 0.0030]\\
\end{longtable}
\endgroup
\begingroup\small\setlength{\tabcolsep}{3pt}
\begin{longtable}{rrll}
\caption{Natural OfficeHome with fixed PCA-64 geometry. Fit and tilt frequencies use all scheduled draws and are averaged over 12 tasks. Brackets give 95\% pointwise Monte Carlo intervals conditional on the pools and tasks.}\label{tab-e1-fixed-fits}\\
\toprule
$n_T$ & $p$ & Fit success (\%) & Extreme tilt (\%)\\\midrule
\endfirsthead
\toprule
$n_T$ & $p$ & Fit success (\%) & Extreme tilt (\%)\\\midrule
\endhead
\bottomrule\endfoot
50 & 8 & 100.00 [100.00, 100.00] & 0.00 [0.00, 0.00]\\
50 & 16 & 100.00 [100.00, 100.00] & 0.00 [0.00, 0.00]\\
50 & 32 & 99.75 [99.42, 100.00] & 10.75 [9.08, 12.33]\\
50 & 64 & 95.83 [94.67, 96.92] & 60.00 [58.67, 61.33]\\
200 & 8 & 100.00 [100.00, 100.00] & 0.00 [0.00, 0.00]\\
200 & 16 & 100.00 [100.00, 100.00] & 0.00 [0.00, 0.00]\\
200 & 32 & 100.00 [100.00, 100.00] & 0.42 [0.08, 0.83]\\
200 & 64 & 100.00 [100.00, 100.00] & 4.92 [3.75, 6.09]\\
800 & 8 & 100.00 [100.00, 100.00] & 0.00 [0.00, 0.00]\\
800 & 16 & 100.00 [100.00, 100.00] & 0.00 [0.00, 0.00]\\
800 & 32 & 100.00 [100.00, 100.00] & 0.00 [0.00, 0.00]\\
800 & 64 & 100.00 [100.00, 100.00] & 0.00 [0.00, 0.00]\\
\end{longtable}
\endgroup
\begingroup\small\setlength{\tabcolsep}{3pt}
\begin{longtable}{lrrllrrr}
\caption{All 144 task-level natural OfficeHome cells at fixed PCA-64 geometry, using all draws with NP-EOT fallback. A, C, P, and R denote Art, Clipart, Product, and Real World. Arrows point from source to target. The bias column gives DRM and NP absolute reference biases. Ratios divide DRM-EOT by NP-EOT, with 95\% pointwise Monte Carlo intervals conditional on the feature pools.}\label{tab-e1-fixed-tasks}\\
\toprule
Task & $n_T$ & $p$ & RMSE ratio & Variance ratio & $|\mathrm{bias}|$ (DRM / NP) & Fit (\%) & Tilt (\%)\\\midrule
\endfirsthead
\toprule
Task & $n_T$ & $p$ & RMSE ratio & Variance ratio & $|\mathrm{bias}|$ (DRM / NP) & Fit (\%) & Tilt (\%)\\\midrule
\endhead
\bottomrule\endfoot
A$\to$C & 50 & 8 & 1.224 [1.023, 1.478] & 2.077 [1.739, 2.475] & 0.0208 / 0.0185 & 100 & 0\\
A$\to$C & 50 & 16 & 0.933 [0.776, 1.126] & 2.023 [1.727, 2.372] & 0.0119 / 0.0185 & 100 & 0\\
A$\to$C & 50 & 32 & 0.672 [0.590, 0.756] & 1.566 [1.335, 1.837] & 0.0019 / 0.0185 & 100 & 13\\
A$\to$C & 50 & 64 & 0.961 [0.931, 0.986] & 1.290 [1.101, 1.535] & 0.0163 / 0.0185 & 91 & 99\\
A$\to$P & 50 & 8 & 1.255 [1.056, 1.471] & 2.059 [1.717, 2.456] & 0.0306 / 0.0253 & 100 & 0\\
A$\to$P & 50 & 16 & 0.795 [0.647, 0.954] & 1.677 [1.402, 1.990] & 0.0156 / 0.0253 & 100 & 0\\
A$\to$P & 50 & 32 & 0.608 [0.492, 0.729] & 1.549 [1.317, 1.797] & 0.0074 / 0.0253 & 100 & 0\\
A$\to$P & 50 & 64 & 0.631 [0.574, 0.688] & 1.561 [1.312, 1.834] & 0.0087 / 0.0253 & 96 & 53\\
A$\to$R & 50 & 8 & 1.332 [1.170, 1.516] & 1.504 [1.146, 1.901] & 0.0374 / 0.0277 & 100 & 0\\
A$\to$R & 50 & 16 & 1.054 [0.912, 1.218] & 1.422 [1.146, 1.746] & 0.0284 / 0.0277 & 100 & 0\\
A$\to$R & 50 & 32 & 0.780 [0.668, 0.909] & 1.377 [1.090, 1.730] & 0.0188 / 0.0277 & 100 & 0\\
A$\to$R & 50 & 64 & 0.464 [0.397, 0.541] & 1.221 [1.046, 1.414] & 0.0042 / 0.0277 & 100 & 0\\
C$\to$A & 50 & 8 & 1.611 [1.363, 1.937] & 2.081 [1.752, 2.473] & 0.0297 / 0.0177 & 100 & 0\\
C$\to$A & 50 & 16 & 0.820 [0.680, 0.989] & 1.847 [1.613, 2.082] & 0.0067 / 0.0177 & 100 & 0\\
C$\to$A & 50 & 32 & 0.748 [0.679, 0.816] & 1.527 [1.348, 1.694] & 0.0062 / 0.0177 & 97 & 20\\
C$\to$A & 50 & 64 & 1.002 [0.990, 1.016] & 1.095 [0.992, 1.240] & 0.0173 / 0.0177 & 100 & 100\\
C$\to$P & 50 & 8 & 1.244 [1.065, 1.455] & 2.200 [1.812, 2.682] & 0.0290 / 0.0245 & 100 & 0\\
C$\to$P & 50 & 16 & 0.709 [0.593, 0.841] & 1.869 [1.587, 2.198] & 0.0104 / 0.0245 & 100 & 0\\
C$\to$P & 50 & 32 & 0.533 [0.475, 0.590] & 1.467 [1.293, 1.656] & 0.0019 / 0.0245 & 100 & 2\\
C$\to$P & 50 & 64 & 0.827 [0.780, 0.869] & 1.750 [1.550, 1.987] & 0.0161 / 0.0245 & 92 & 92\\
C$\to$R & 50 & 8 & 0.770 [0.646, 0.914] & 1.776 [1.552, 2.080] & 0.0097 / 0.0222 & 100 & 0\\
C$\to$R & 50 & 16 & 0.634 [0.560, 0.720] & 1.568 [1.388, 1.802] & 0.0019 / 0.0222 & 100 & 0\\
C$\to$R & 50 & 32 & 0.717 [0.676, 0.754] & 1.388 [1.245, 1.551] & 0.0103 / 0.0222 & 100 & 28\\
C$\to$R & 50 & 64 & 0.987 [0.972, 0.998] & 1.199 [1.086, 1.329] & 0.0210 / 0.0222 & 98 & 99\\
P$\to$A & 50 & 8 & 3.195 [2.838, 3.623] & 2.932 [2.216, 3.864] & 0.0736 / 0.0211 & 100 & 0\\
P$\to$A & 50 & 16 & 1.085 [0.909, 1.288] & 2.114 [1.710, 2.607] & 0.0204 / 0.0211 & 100 & 0\\
P$\to$A & 50 & 32 & 0.755 [0.623, 0.898] & 2.091 [1.755, 2.546] & 0.0087 / 0.0211 & 100 & 0\\
P$\to$A & 50 & 64 & 0.831 [0.759, 0.900] & 2.694 [2.325, 3.191] & 0.0084 / 0.0211 & 94 & 73\\
P$\to$C & 50 & 8 & 2.084 [1.839, 2.366] & 2.461 [1.849, 3.194] & 0.0584 / 0.0268 & 100 & 0\\
P$\to$C & 50 & 16 & 1.061 [0.904, 1.244] & 1.912 [1.544, 2.372] & 0.0260 / 0.0268 & 100 & 0\\
P$\to$C & 50 & 32 & 0.575 [0.489, 0.680] & 1.625 [1.398, 1.888] & 0.0052 / 0.0268 & 100 & 35\\
P$\to$C & 50 & 64 & 0.864 [0.816, 0.906] & 1.933 [1.658, 2.233] & 0.0185 / 0.0268 & 88 & 97\\
P$\to$R & 50 & 8 & 1.177 [1.053, 1.309] & 2.522 [1.830, 3.555] & 0.0437 / 0.0386 & 100 & 0\\
P$\to$R & 50 & 16 & 0.695 [0.608, 0.787] & 1.958 [1.493, 2.586] & 0.0228 / 0.0386 & 100 & 0\\
P$\to$R & 50 & 32 & 0.384 [0.333, 0.438] & 1.519 [1.191, 1.989] & 0.0061 / 0.0386 & 100 & 0\\
P$\to$R & 50 & 64 & 0.416 [0.366, 0.461] & 1.478 [1.187, 1.829] & 0.0091 / 0.0386 & 100 & 5\\
R$\to$A & 50 & 8 & 1.618 [1.454, 1.808] & 2.058 [1.568, 2.700] & 0.0540 / 0.0329 & 100 & 0\\
R$\to$A & 50 & 16 & 0.846 [0.726, 0.981] & 1.816 [1.410, 2.343] & 0.0250 / 0.0329 & 100 & 0\\
R$\to$A & 50 & 32 & 0.590 [0.491, 0.694] & 1.604 [1.253, 2.018] & 0.0142 / 0.0329 & 100 & 0\\
R$\to$A & 50 & 64 & 0.407 [0.351, 0.465] & 1.441 [1.145, 1.820] & 0.0013 / 0.0329 & 100 & 0\\
R$\to$C & 50 & 8 & 1.072 [0.915, 1.269] & 1.724 [1.378, 2.169] & 0.0255 / 0.0256 & 100 & 0\\
R$\to$C & 50 & 16 & 0.668 [0.567, 0.792] & 1.513 [1.231, 1.864] & 0.0098 / 0.0256 & 100 & 0\\
R$\to$C & 50 & 32 & 0.594 [0.544, 0.644] & 1.302 [1.145, 1.506] & 0.0075 / 0.0256 & 100 & 31\\
R$\to$C & 50 & 64 & 0.951 [0.924, 0.973] & 1.398 [1.265, 1.545] & 0.0224 / 0.0256 & 91 & 99\\
R$\to$P & 50 & 8 & 1.055 [0.932, 1.190] & 2.512 [1.973, 3.237] & 0.0421 / 0.0422 & 100 & 0\\
R$\to$P & 50 & 16 & 0.656 [0.572, 0.750] & 1.949 [1.538, 2.502] & 0.0233 / 0.0422 & 100 & 0\\
R$\to$P & 50 & 32 & 0.449 [0.389, 0.516] & 1.796 [1.436, 2.273] & 0.0112 / 0.0422 & 100 & 0\\
R$\to$P & 50 & 64 & 0.364 [0.322, 0.403] & 1.373 [1.133, 1.668] & 0.0073 / 0.0422 & 100 & 3\\
A$\to$C & 200 & 8 & 2.205 [1.869, 2.605] & 1.476 [1.210, 1.751] & 0.0197 / 0.0073 & 100 & 0\\
A$\to$C & 200 & 16 & 1.693 [1.427, 2.016] & 1.376 [1.204, 1.575] & 0.0145 / 0.0073 & 100 & 0\\
A$\to$C & 200 & 32 & 0.866 [0.716, 1.046] & 1.253 [1.134, 1.395] & 0.0046 / 0.0073 & 100 & 0\\
A$\to$C & 200 & 64 & 0.728 [0.624, 0.858] & 1.185 [1.071, 1.304] & 0.0018 / 0.0073 & 100 & 4\\
A$\to$P & 200 & 8 & 3.471 [2.963, 4.124] & 1.469 [1.234, 1.760] & 0.0341 / 0.0071 & 100 & 0\\
A$\to$P & 200 & 16 & 2.410 [2.028, 2.894] & 1.286 [1.109, 1.519] & 0.0230 / 0.0071 & 100 & 0\\
A$\to$P & 200 & 32 & 2.094 [1.757, 2.527] & 1.179 [1.006, 1.408] & 0.0198 / 0.0071 & 100 & 0\\
A$\to$P & 200 & 64 & 1.336 [1.108, 1.634] & 1.174 [1.041, 1.337] & 0.0110 / 0.0071 & 100 & 0\\
A$\to$R & 200 & 8 & 3.447 [3.026, 4.005] & 1.523 [1.193, 1.951] & 0.0374 / 0.0090 & 100 & 0\\
A$\to$R & 200 & 16 & 3.054 [2.663, 3.564] & 1.542 [1.210, 1.949] & 0.0329 / 0.0090 & 100 & 0\\
A$\to$R & 200 & 32 & 2.763 [2.405, 3.234] & 1.430 [1.151, 1.750] & 0.0297 / 0.0090 & 100 & 0\\
A$\to$R & 200 & 64 & 2.213 [1.905, 2.610] & 1.420 [1.191, 1.663] & 0.0233 / 0.0090 & 100 & 0\\
C$\to$A & 200 & 8 & 3.858 [3.282, 4.597] & 1.825 [1.589, 2.083] & 0.0319 / 0.0047 & 100 & 0\\
C$\to$A & 200 & 16 & 1.915 [1.601, 2.318] & 1.488 [1.311, 1.698] & 0.0140 / 0.0047 & 100 & 0\\
C$\to$A & 200 & 32 & 1.174 [0.988, 1.410] & 1.349 [1.223, 1.500] & 0.0056 / 0.0047 & 100 & 0\\
C$\to$A & 200 & 64 & 0.992 [0.869, 1.146] & 1.313 [1.208, 1.430] & 0.0021 / 0.0047 & 100 & 11\\
C$\to$P & 200 & 8 & 2.824 [2.430, 3.298] & 1.782 [1.496, 2.142] & 0.0303 / 0.0083 & 100 & 0\\
C$\to$P & 200 & 16 & 1.701 [1.437, 2.018] & 1.480 [1.253, 1.727] & 0.0169 / 0.0083 & 100 & 0\\
C$\to$P & 200 & 32 & 1.055 [0.870, 1.278] & 1.252 [1.082, 1.438] & 0.0083 / 0.0083 & 100 & 0\\
C$\to$P & 200 & 64 & 0.809 [0.659, 0.976] & 1.210 [1.096, 1.333] & 0.0036 / 0.0083 & 100 & 15\\
C$\to$R & 200 & 8 & 1.841 [1.526, 2.225] & 1.556 [1.404, 1.708] & 0.0153 / 0.0060 & 100 & 0\\
C$\to$R & 200 & 16 & 1.440 [1.188, 1.757] & 1.420 [1.291, 1.553] & 0.0107 / 0.0060 & 100 & 0\\
C$\to$R & 200 & 32 & 0.979 [0.824, 1.163] & 1.325 [1.222, 1.434] & 0.0034 / 0.0060 & 100 & 0\\
C$\to$R & 200 & 64 & 0.892 [0.784, 1.022] & 1.237 [1.157, 1.317] & 0.0011 / 0.0060 & 100 & 16\\
P$\to$A & 200 & 8 & 8.286 [7.178, 9.672] & 3.160 [2.410, 4.185] & 0.0723 / 0.0064 & 100 & 0\\
P$\to$A & 200 & 16 & 2.978 [2.527, 3.532] & 1.732 [1.369, 2.193] & 0.0250 / 0.0064 & 100 & 0\\
P$\to$A & 200 & 32 & 2.350 [1.978, 2.805] & 1.692 [1.390, 2.060] & 0.0192 / 0.0064 & 100 & 0\\
P$\to$A & 200 & 64 & 1.613 [1.338, 1.953] & 1.621 [1.388, 1.871] & 0.0120 / 0.0064 & 100 & 0\\
P$\to$C & 200 & 8 & 5.188 [4.484, 6.109] & 2.064 [1.638, 2.655] & 0.0564 / 0.0083 & 100 & 0\\
P$\to$C & 200 & 16 & 2.784 [2.364, 3.341] & 1.765 [1.482, 2.146] & 0.0292 / 0.0083 & 100 & 0\\
P$\to$C & 200 & 32 & 1.656 [1.382, 2.013] & 1.315 [1.157, 1.522] & 0.0163 / 0.0083 & 100 & 5\\
P$\to$C & 200 & 64 & 0.930 [0.775, 1.133] & 1.306 [1.180, 1.444] & 0.0059 / 0.0083 & 100 & 6\\
P$\to$R & 200 & 8 & 2.923 [2.623, 3.277] & 1.705 [1.288, 2.258] & 0.0446 / 0.0139 & 100 & 0\\
P$\to$R & 200 & 16 & 2.057 [1.820, 2.342] & 1.693 [1.329, 2.159] & 0.0307 / 0.0139 & 100 & 0\\
P$\to$R & 200 & 32 & 1.430 [1.247, 1.646] & 1.484 [1.180, 1.873] & 0.0206 / 0.0139 & 100 & 0\\
P$\to$R & 200 & 64 & 1.168 [1.014, 1.357] & 1.311 [1.091, 1.593] & 0.0163 / 0.0139 & 100 & 0\\
R$\to$A & 200 & 8 & 4.889 [4.419, 5.458] & 2.291 [1.759, 3.014] & 0.0588 / 0.0108 & 100 & 0\\
R$\to$A & 200 & 16 & 2.871 [2.554, 3.241] & 1.668 [1.288, 2.152] & 0.0341 / 0.0108 & 100 & 0\\
R$\to$A & 200 & 32 & 2.492 [2.210, 2.819] & 1.479 [1.168, 1.852] & 0.0295 / 0.0108 & 100 & 0\\
R$\to$A & 200 & 64 & 1.877 [1.641, 2.150] & 1.558 [1.285, 1.872] & 0.0217 / 0.0108 & 100 & 0\\
R$\to$C & 200 & 8 & 2.681 [2.312, 3.169] & 1.586 [1.257, 1.990] & 0.0279 / 0.0084 & 100 & 0\\
R$\to$C & 200 & 16 & 1.569 [1.312, 1.902] & 1.484 [1.228, 1.783] & 0.0149 / 0.0084 & 100 & 0\\
R$\to$C & 200 & 32 & 0.776 [0.645, 0.939] & 1.267 [1.141, 1.413] & 0.0033 / 0.0084 & 100 & 0\\
R$\to$C & 200 & 64 & 0.681 [0.598, 0.787] & 1.148 [1.048, 1.260] & 0.0004 / 0.0084 & 100 & 7\\
R$\to$P & 200 & 8 & 3.099 [2.759, 3.507] & 1.374 [1.117, 1.730] & 0.0465 / 0.0134 & 100 & 0\\
R$\to$P & 200 & 16 & 2.249 [1.988, 2.567] & 1.225 [1.014, 1.489] & 0.0334 / 0.0134 & 100 & 0\\
R$\to$P & 200 & 32 & 1.906 [1.668, 2.192] & 1.183 [1.003, 1.402] & 0.0280 / 0.0134 & 100 & 0\\
R$\to$P & 200 & 64 & 1.389 [1.194, 1.626] & 1.267 [1.108, 1.469] & 0.0196 / 0.0134 & 100 & 0\\
A$\to$C & 800 & 8 & 4.497 [3.760, 5.400] & 1.456 [1.295, 1.628] & 0.0195 / 0.0019 & 100 & 0\\
A$\to$C & 800 & 16 & 3.486 [2.894, 4.218] & 1.357 [1.235, 1.493] & 0.0149 / 0.0019 & 100 & 0\\
A$\to$C & 800 & 32 & 1.878 [1.545, 2.302] & 1.210 [1.111, 1.314] & 0.0071 / 0.0019 & 100 & 0\\
A$\to$C & 800 & 64 & 1.551 [1.285, 1.899] & 1.161 [1.086, 1.233] & 0.0054 / 0.0019 & 100 & 0\\
A$\to$P & 800 & 8 & 6.653 [5.744, 7.862] & 1.273 [1.042, 1.545] & 0.0307 / 0.0023 & 100 & 0\\
A$\to$P & 800 & 16 & 4.679 [4.002, 5.590] & 1.302 [1.101, 1.529] & 0.0214 / 0.0023 & 100 & 0\\
A$\to$P & 800 & 32 & 4.340 [3.699, 5.190] & 1.253 [1.069, 1.474] & 0.0198 / 0.0023 & 100 & 0\\
A$\to$P & 800 & 64 & 3.104 [2.613, 3.752] & 1.236 [1.101, 1.395] & 0.0138 / 0.0023 & 100 & 0\\
A$\to$R & 800 & 8 & 6.480 [5.723, 7.427] & 1.007 [0.832, 1.230] & 0.0311 / 0.0031 & 100 & 0\\
A$\to$R & 800 & 16 & 5.905 [5.210, 6.781] & 1.049 [0.874, 1.258] & 0.0283 / 0.0031 & 100 & 0\\
A$\to$R & 800 & 32 & 5.482 [4.827, 6.303] & 0.973 [0.818, 1.150] & 0.0262 / 0.0031 & 100 & 0\\
A$\to$R & 800 & 64 & 4.722 [4.133, 5.437] & 0.994 [0.866, 1.147] & 0.0225 / 0.0031 & 100 & 0\\
C$\to$A & 800 & 8 & 6.076 [5.311, 7.036] & 1.568 [1.367, 1.790] & 0.0272 / 0.0013 & 100 & 0\\
C$\to$A & 800 & 16 & 3.056 [2.637, 3.570] & 1.250 [1.124, 1.398] & 0.0131 / 0.0013 & 100 & 0\\
C$\to$A & 800 & 32 & 1.846 [1.567, 2.160] & 1.178 [1.086, 1.285] & 0.0070 / 0.0013 & 100 & 0\\
C$\to$A & 800 & 64 & 1.482 [1.251, 1.733] & 1.163 [1.097, 1.241] & 0.0049 / 0.0013 & 100 & 0\\
C$\to$P & 800 & 8 & 4.488 [3.882, 5.237] & 1.193 [1.044, 1.373] & 0.0249 / 0.0027 & 100 & 0\\
C$\to$P & 800 & 16 & 3.010 [2.588, 3.553] & 1.084 [0.946, 1.232] & 0.0163 / 0.0027 & 100 & 0\\
C$\to$P & 800 & 32 & 1.970 [1.671, 2.341] & 1.091 [0.980, 1.210] & 0.0099 / 0.0027 & 100 & 0\\
C$\to$P & 800 & 64 & 1.573 [1.319, 1.872] & 1.152 [1.072, 1.238] & 0.0071 / 0.0027 & 100 & 0\\
C$\to$R & 800 & 8 & 2.562 [2.148, 3.083] & 1.211 [1.086, 1.362] & 0.0115 / 0.0020 & 100 & 0\\
C$\to$R & 800 & 16 & 2.379 [1.996, 2.867] & 1.181 [1.052, 1.328] & 0.0105 / 0.0020 & 100 & 0\\
C$\to$R & 800 & 32 & 1.450 [1.196, 1.759] & 1.107 [1.006, 1.205] & 0.0053 / 0.0020 & 100 & 0\\
C$\to$R & 800 & 64 & 1.283 [1.064, 1.547] & 1.134 [1.069, 1.211] & 0.0041 / 0.0020 & 100 & 0\\
P$\to$A & 800 & 8 & 12.830 [11.184, 14.953] & 1.665 [1.308, 2.104] & 0.0607 / 0.0021 & 100 & 0\\
P$\to$A & 800 & 16 & 4.392 [3.768, 5.197] & 1.217 [1.047, 1.420] & 0.0203 / 0.0021 & 100 & 0\\
P$\to$A & 800 & 32 & 3.624 [3.089, 4.307] & 1.286 [1.108, 1.489] & 0.0165 / 0.0021 & 100 & 0\\
P$\to$A & 800 & 64 & 2.792 [2.367, 3.328] & 1.262 [1.114, 1.414] & 0.0124 / 0.0021 & 100 & 0\\
P$\to$C & 800 & 8 & 8.879 [7.727, 10.458] & 1.330 [1.093, 1.644] & 0.0506 / 0.0030 & 100 & 0\\
P$\to$C & 800 & 16 & 4.566 [3.912, 5.427] & 1.345 [1.151, 1.580] & 0.0256 / 0.0030 & 100 & 0\\
P$\to$C & 800 & 32 & 2.921 [2.472, 3.509] & 1.193 [1.036, 1.371] & 0.0159 / 0.0030 & 100 & 0\\
P$\to$C & 800 & 64 & 1.746 [1.450, 2.128] & 1.135 [1.039, 1.247] & 0.0086 / 0.0030 & 100 & 0\\
P$\to$R & 800 & 8 & 6.414 [5.714, 7.344] & 1.544 [1.224, 1.909] & 0.0375 / 0.0043 & 100 & 0\\
P$\to$R & 800 & 16 & 4.722 [4.163, 5.451] & 1.399 [1.154, 1.688] & 0.0274 / 0.0043 & 100 & 0\\
P$\to$R & 800 & 32 & 3.647 [3.197, 4.234] & 1.389 [1.159, 1.665] & 0.0210 / 0.0043 & 100 & 0\\
P$\to$R & 800 & 64 & 3.536 [3.087, 4.099] & 1.282 [1.112, 1.463] & 0.0203 / 0.0043 & 100 & 0\\
R$\to$A & 800 & 8 & 8.995 [7.912, 10.309] & 1.279 [0.968, 1.706] & 0.0487 / 0.0037 & 100 & 0\\
R$\to$A & 800 & 16 & 5.299 [4.620, 6.140] & 1.164 [0.958, 1.426] & 0.0285 / 0.0037 & 100 & 0\\
R$\to$A & 800 & 32 & 4.914 [4.272, 5.716] & 1.281 [1.097, 1.515] & 0.0264 / 0.0037 & 100 & 0\\
R$\to$A & 800 & 64 & 4.011 [3.466, 4.690] & 1.260 [1.126, 1.432] & 0.0214 / 0.0037 & 100 & 0\\
R$\to$C & 800 & 8 & 5.124 [4.418, 6.107] & 1.243 [1.084, 1.462] & 0.0264 / 0.0021 & 100 & 0\\
R$\to$C & 800 & 16 & 2.877 [2.464, 3.438] & 1.215 [1.101, 1.350] & 0.0142 / 0.0021 & 100 & 0\\
R$\to$C & 800 & 32 & 1.481 [1.245, 1.760] & 1.127 [1.040, 1.223] & 0.0059 / 0.0021 & 100 & 0\\
R$\to$C & 800 & 64 & 1.279 [1.078, 1.504] & 1.089 [1.019, 1.163] & 0.0044 / 0.0021 & 100 & 0\\
R$\to$P & 800 & 8 & 6.329 [5.641, 7.215] & 1.265 [0.934, 1.704] & 0.0400 / 0.0049 & 100 & 0\\
R$\to$P & 800 & 16 & 4.538 [4.026, 5.211] & 1.190 [0.951, 1.531] & 0.0285 / 0.0049 & 100 & 0\\
R$\to$P & 800 & 32 & 4.166 [3.688, 4.791] & 1.209 [0.994, 1.503] & 0.0261 / 0.0049 & 100 & 0\\
R$\to$P & 800 & 64 & 3.477 [3.054, 4.027] & 1.257 [1.100, 1.450] & 0.0217 / 0.0049 & 100 & 0\\
\end{longtable}
\endgroup

\paragraph{Numerical tolerance check.}
We rerun all four bases on 84 selected draws: the first two repetitions per task and target size, plus 12 additional stress draws with large tilts. Tightening the Sinkhorn tolerance from $10^{-9}$ to $10^{-11}$ leaves all success statuses unchanged and changes the fallback EOT estimates by at most $7.78\times10^{-16}$. Also tightening the DRM score tolerance from $10^{-8}$ to $10^{-10}$ triggers fallback in two additional high-tilt fits at $(n_T,p)=(50,64)$, with a maximum fallback EOT difference of $9.45\times10^{-8}$. These checks concern the selected draws.

\paragraph{Nested bases under exact specification.}
On each fixed PCA-64 base-domain support, take the source to be uniform and the target to be an exact exponential tilt along the first coordinate. Every $Q_p$ includes the true tilt. At allocation $(0.8,0.2)$ and $\varepsilon=0.1$, the oracle variance ratios are nondecreasing in $p$ and at most one in all eight domain--ESS combinations. The four-domain mean ratios in order $p=8,16,32,64$ are $0.891,0.911,0.927,0.945$ at ESS ratio $0.5$ and $0.845,0.878,0.901,0.926$ at ESS ratio $0.8$. This checks the population ordering in Theorem~1 under exact specification; it does not assert a finite-sample ordering for natural shifts.

\clearpage

\section{Unbounded-support discussion}\label{app:I}
\subsection{Remarks on unbounded-support extensions}
\label{unbounded-extensions}

Compact support supplies the envelope bounds and potential stability used in the attainment and inference proofs. The tangent-space calculation, orthogonal projection, and efficiency identity of Appendix Theorem~1 remain formally meaningful on unbounded supports.

On an unbounded domain such as \(K=\mathbb R^d\), consider the multiplier classes
\[
\frac{1}{\omega_{\theta,\rho}}\mathscr F_\varepsilon,
\qquad
\frac{r_\theta}{\omega_{\theta,\rho}}\mathscr F_\varepsilon.
\]
Their scalar weights are bounded by $1/(1-\rho)$ and $1/\rho$, respectively, when $\omega_{\theta,\rho}=1-\rho+\rho r_\theta$ and $\rho$ stays away from zero and one. The difficulty is that the potential class itself need not have a bounded envelope, while derivatives of the weights introduce the possibly unbounded basis $Q$, for example when $q(x)=x$ or contains polynomial terms. Uniform control over parameter neighborhoods therefore requires additional moment and envelope bounds. Extending Appendix Theorems~2--4 would require a weighted empirical-process limit, uniform Taylor-remainder bounds, stability of the EOT potentials in a weighted norm, and the corresponding conditional bootstrap results.

Sub-Gaussian marginals and a suitably controlled DRM basis give a setting in which to investigate these weighted bounds.

\section{Annual cash dividend proposal experiment}\label{app:J}

We compare the finite-sample variance and reference RMSE of scalar EOT estimates for two groups formed from observed annual dividend proposals. Each fiscal year supplies its own empirical pools and reference.

\subsection{Annual pools and group labels}\label{sec:dividend-pools}
The final exchange session of each fiscal year sets the feature cutoff. The sampling frame comprises Shanghai and Shenzhen A-share firms that are still listed on that date and have been listed for at least 365 days. We use Tushare's dividend proposal records\footnote{Field definitions: \url{https://tushare.pro/document/2?doc_id=103}.} for annual periods ending on 31 December. We select proposals announced in the next calendar year and by the delisting date for firms that delist. Firms with a recorded proposal for that period on or before the cutoff are excluded. The recorded pre-tax per-share cash amount must be finite and nonnegative. Positive amounts define group 0, and explicitly recorded zero amounts define group 1. The labels describe proposed amounts for the specified annual period.

We exclude securities with conflicting positive and zero records and use the earliest qualifying record for securities with consistent labels. We use the company identifier when available and the security code otherwise. If multiple eligible securities map to the same issuer in a fiscal year, we exclude those entries. The complete annual pools consist of all remaining firms with valid features. Table~\ref{tab:dividend-pools} gives their sizes. Firms may recur across years. The FY2023 and FY2024 pools share 4,645 issuers.

\begin{table}[H]
\centering\small
\caption{Company counts in the complete annual dividend proposal pools. Each firm appears at most once per fiscal year.}\label{tab:dividend-pools}
\begin{tabular}{lrr}\toprule
Fiscal year & Positive amount (group 0) & Explicit zero (group 1)\\\midrule

FY2018 & 2340 & 907 \\
FY2019 & 2350 & 1066 \\
FY2020 & 2467 & 1170 \\
FY2021 & 2655 & 1378 \\
FY2022 & 2878 & 1580 \\
FY2023 & 3362 & 1423 \\
FY2024 & 3306 & 1628 \\

\bottomrule
\end{tabular}
\end{table}

\subsection{Features and overlap}\label{sec:dividend-overlap}
Each firm's price series must cover the last 61 consecutive exchange sessions of the fiscal year. We multiply closing prices by adjustment factors. The features are the cumulative return over the 60 price changes, the log sample standard deviation of the last 20 simple daily returns, and log market capitalization on the final session. We require positive finite adjusted prices and market capitalization, finite reported daily returns for the final 60 sessions, positive finite volume and turnover in the last 20 sessions, and positive return volatility. We center and scale each coordinate by its pooled mean and population standard deviation over the 18,791 eligible firm-year records in FY2018--FY2022, then divide by $\sqrt{3}$. The same fitted transform defines the geometry throughout, including in the later FY2023 and FY2024 evaluation cohorts.

Let $G$ be the zero-proposal indicator, and let $\pi_g$ be the group proportions before the experiment's fixed-size draws. By Bayes' rule, a model with log odds $a+b^\top z$ gives a feature density ratio $(\pi_0/\pi_1)\exp(a+b^\top z)$. This motivates the linear DRM as a working model. For an overlap diagnostic, we score each annual pool by five-fold stratified cross-validation with a class-balanced, $\ell_2$-regularized logistic model using the same three features. Out-of-fold AUROC is $0.672$ for FY2023 and $0.712$ for FY2024. In each of those years, more than $99\%$ of the positive-proposal scores and all zero-proposal scores are in $[0.05,0.95]$. The overlap diagnostic describes the distribution of fitted scores. Linearity of the population log odds is a separate model specification question. The EOT comparison uses an unpenalized DRM refitted to each draw.

\subsection{Paired evaluation and numerical checks}\label{sec:dividend-evaluation}
In each year and size setting, we make 600 independent draws with replacement within the two annual pools, with $n_0=240$ and $n_1\in\{20,40\}$. These sample sizes define the target-scarce evaluation design. Each pair uses the same two samples for NP-EOT and DRM-EOT. The DRM uses $Q(z)=(1,z^\top)^\top$ and the sample fractions $n_g/(n_0+n_1)$. Both methods use $c(z,z')=\|z-z'\|^2/2$ and product-marginal KL regularization with $\varepsilon=0.1$. The common reference is the NP-EOT value between the uniform empirical measures on the two complete annual pools. It is fixed within each year.

For each method, the sample mean of the 600 errors relative to the reference estimates signed reference bias. The sample variance uses denominator $599$. RMSE is the square root of the mean squared error relative to the same reference. Table~\ref{tab:dividend-errors} reports the reference, signed biases and RMSEs. For each ratio we resample the 600 paired estimates jointly 4,000 times and take the $2.5$th and $97.5$th percentiles. The resulting 95\% pointwise Monte Carlo intervals quantify simulation uncertainty conditional on the corresponding annual empirical pools. 

All 8,400 scheduled pairs produced finite estimates and met the transport marginal residual tolerance of $10^{-9}$. The fitted tilt uses score tolerance $10^{-8}$, and the annual references use transport residual tolerance $10^{-10}$. Transport computations use Newton refinement and a log-domain solver as needed. The numerical audit verified every draw hash, recomputed the seven references and checked 42 paired estimates with tenfold tighter fit and transport tolerances.

\paragraph{Reproduction inputs.}
The archive \texttt{anonymous\_code\_reviewer.zip} contains the saved estimates, table exporter, and refit code. From the extracted \texttt{reproducibility/} directory, install the dependencies in \texttt{requirements.txt} and run \texttt{python} \path{scripts/verify_package.py} to check the archive manifest and recompute the dividend summaries from the 8,400 saved pairs. Run \texttt{python} \path{experiments/dividend/export_tables.py} \texttt{--out} \path{outputs/dividend_tex} to regenerate the main results table, annual counts, and reference-error table. These commands use inputs included in the archive.

A fresh dividend refit uses \path{experiments/dividend/run.py} and a separately obtained, matching company-year feature snapshot. The guide at \path{data/dividend/DATA_PROVENANCE.md} identifies the authorized Tushare records and the feature construction command. The runner checks the snapshot SHA-256 against \path{experiments/dividend/design.json}. The same hash is recorded in \path{data/dividend/pool_manifest.json}. The commands in \path{experiments/dividend/README.md} cover both a short replay and all 14 configurations.

\begin{table}[H]
\centering\small
\caption{Reference bias and RMSE for the dividend proposal experiment. Bias is signed relative to the complete annual NP-EOT reference. Each row summarizes 600 NP-EOT and DRM-EOT pairs conditional on the annual pools.}\label{tab:dividend-errors}
\begin{tabular}{lrrrrrr}\toprule
Fiscal year & $n_1$ & Reference & NP bias & DRM bias & NP RMSE & DRM RMSE\\\midrule

FY2018 & 20 & 0.313 & 0.049 & 0.010 & 0.078 & 0.050 \\
FY2018 & 40 & 0.313 & 0.031 & 0.002 & 0.051 & 0.035 \\
FY2019 & 20 & 0.358 & 0.091 & 0.044 & 0.152 & 0.093 \\
FY2019 & 40 & 0.358 & 0.059 & 0.021 & 0.112 & 0.053 \\
FY2020 & 20 & 0.423 & 0.138 & 0.066 & 0.323 & 0.243 \\
FY2020 & 40 & 0.423 & 0.086 & 0.038 & 0.187 & 0.130 \\
FY2021 & 20 & 0.430 & 0.112 & 0.037 & 0.214 & 0.154 \\
FY2021 & 40 & 0.430 & 0.083 & 0.030 & 0.179 & 0.140 \\
FY2022 & 20 & 0.384 & 0.078 & 0.029 & 0.213 & 0.164 \\
FY2022 & 40 & 0.384 & 0.048 & 0.012 & 0.137 & 0.102 \\
FY2023 & 20 & 0.341 & 0.080 & 0.042 & 0.118 & 0.091 \\
FY2023 & 40 & 0.341 & 0.050 & 0.023 & 0.078 & 0.063 \\
FY2024 & 20 & 0.444 & 0.083 & 0.022 & 0.161 & 0.112 \\
FY2024 & 40 & 0.444 & 0.057 & 0.011 & 0.126 & 0.089 \\

\bottomrule
\end{tabular}
\end{table}

\Needspace{10\baselineskip}

\section{Vanishing target fractions and a change in convergence rate}\label{app:K}\label{app:rare-target}

We prove main-text Proposition~1 for the original unpenalized empirical-likelihood estimator when both samples grow and the target fraction vanishes. The support $K\subset\mathbb R^d$ is compact, the basis $q:K\to\mathbb R^p$ is continuous with fixed $p$, and the regularization $\varepsilon>0$ is fixed. The cost is $c(x,y)=\|x-y\|^2/2$. At a fixed, correctly specified model point $(F,\beta)$, write $F(f)=\int f\,dF$ and define
\[
 R_{F,b}(dx)=\frac{e^{b^\top q(x)}}{F(e^{b^\top q})}\,F(dx),
 \qquad F_1=R_{F,\beta},\qquad
 T(F,b)=\Psi_\varepsilon(F,R_{F,b}).
 \tag{K.1}\label{eq:rare-model}
\]
Write $N=n_0$ and $m=n_1$ for the source and target sample sizes. We observe independent samples $X_1,\ldots,X_N\sim F$ and $Y_1,\ldots,Y_m\sim F_1$, with $N,m\to\infty$ and $\lambda=m/N\to0$. Let $P_N$ and $Q_m$ be their empirical measures. All derivatives and population quantities below are evaluated at this fixed model point.

The proof tracks baseline and tilt estimation errors at their respective rates. Section~\ref{sec:rare-smoothness} establishes joint smoothness of the tilted scalar EOT cost. Section~\ref{sec:rare-fit-expansion} shows that the fitted baseline differs from the source empirical measure by $o_p(N^{-1/2})$ and that the tilt is estimated at rate $m^{-1/2}$. When the first-order tilt derivative vanishes, the expansion in Section~\ref{sec:rare-phase-proof} gives the linear source and quadratic tilt terms that determine the three limits. Section~\ref{sec:rare-example} verifies nondegeneracy and shows how a larger correctly specified tilt space can restore a first-order target term.

\subsection{Estimator and derivatives}\label{sec:rare-estimator}
Set $Q=(1,q^\top)^\top$, $\theta=(\alpha,\beta^\top)^\top$, $\alpha=-\log F(e^{\beta^\top q})$, and $r_\theta=e^{\theta^\top Q}$. The intercept normalizes the tilt in \eqref{eq:rare-model}, as in the main-text DRM. The pooled likelihood and weights below are the main-text empirical-likelihood fit written in terms of $N$, $m$, and $\lambda$. The pooled sum counts each indexed observation, including repeated values. Apart from a parameter-independent constant, the dual likelihood is
\[
 \ell_{N,m}(t)=\sum_{j=1}^m t^\top Q(Y_j)
 -\sum_{z\in\{X_i,Y_j\}}\log\{N+m r_t(z)\}.
 \tag{K.2}\label{eq:rare-likelihood}
\]
At a finite maximum $\widehat\theta=(\widehat\alpha,\widehat\beta^\top)^\top$, its fitted measures are
\[
 \widehat F=\frac{P_N+\lambda Q_m}{1+\lambda r_{\widehat\theta}},
 \qquad \widehat F_1=r_{\widehat\theta}\widehat F,
 \qquad \widehat T=\Psi_\varepsilon(\widehat F,\widehat F_1).
 \tag{K.3}\label{eq:rare-fits}
\]
Division denotes weighting of a measure. The intercept score gives $\widehat F_1(1)=1$, and the identity $\widehat F(1)+\lambda\widehat F_1(1)=1+\lambda$ gives $\widehat F(1)=1$. Consequently,
$\widehat\alpha=-\log\widehat F(e^{\widehat\beta^\top q})$ and $\widehat T=T(\widehat F,\widehat\beta)$. On the event that a finite maximum does not exist, use the NP-EOT fallback specified in Appendix~\ref{app:D}. Section~\ref{sec:rare-fit-expansion} proves that this event has probability tending to zero.

Let $\phi_0,\phi_1$ be the EOT potentials centered under $F,F_1$, respectively, and define
\[
 \begin{gathered}
 q_c=q-E_1q,\qquad J=E_1(q_cq_c^\top),\qquad
 a=E_1(\phi_1q_c),\\
 \psi=\phi_0+r_\theta\phi_1,\qquad
 \sigma^2=E_0\psi^2,\qquad
 \Gamma=\partial_b^2T(F,b)\big|_{b=\beta}.
 \end{gathered}
 \tag{K.4}\label{eq:rare-derivatives}
\]
We assume $J\succ0$. For the second-order conclusion we also assume $a=0$, $\sigma^2>0$, and $\Gamma\ne0$. The condition $a=0$ removes the first-order tilt term in this expansion. In the fixed-proportion experiment, the weighted projection residual $e$ determines the efficiency gain in main-text Theorem~1.

For any mass-zero baseline perturbation $H$, differentiation of the normalized tilt gives
\[
 D_FR[H]=r_\theta H-F_1H(r_\theta),\qquad
 D_bR[v]=(v^\top q_c)F_1.
 \tag{K.5}\label{eq:rare-tilt-derivatives}
\]
Pairing these expressions with the centered EOT potentials yields
\[
 D_FT[H]=H(\psi),\qquad D_bT=a.
 \tag{K.6}\label{eq:rare-first-derivatives}
\]
In particular, $E_0\psi=0$.

\subsection{Joint smoothness in a norm for empirical measures}\label{sec:rare-smoothness}
\citet[Theorem 4]{goldfeld2024limit} establish second-order Hadamard differentiability of EOT potentials with respect to the marginals. For the DRM expansion, we construct a jointly $C^3$ extension of the tilted EOT cost in a norm in which the two empirical measures have their respective square-root rates. To work with a continuous basis in this smooth function space, embed $K$ by $\iota(x)=(x,q(x))$, and let $D\subset\mathbb R^{d+p}$ be a bounded smooth domain containing $\iota(K)$ in its interior. Choose an integer $s>(d+p)/2$, and set
\[
 E=H^s(D),\qquad \mathbb B=E^*.
\]
All measures in this subsection are pushed forward by $\iota$. Sobolev embedding makes point evaluation bounded uniformly on $\iota(K)$, and $E$ is a Banach algebra. The tilt is now multiplication by the smooth function $e^{b^\top t}$ on the augmented coordinates $(x,t)$. The embedding supplies smooth multipliers for the proof. The transport cost continues to use only $x$. Continuity of $q$ keeps the embedded support compact.

For measures $\mu,\nu$ and a fixed $z_*\in\iota(K)$, consider
\[
 \mathcal S(u,v,\tau;\mu,\nu)=
 \begin{pmatrix}
 u+\varepsilon\log\nu[e^{(v-c)/\varepsilon}]\\
 v+\varepsilon\log\mu[e^{(u-c)/\varepsilon}]-\tau\\
 u(z_*)
 \end{pmatrix}.
 \tag{K.7}\label{eq:rare-schrodinger}
\]
Here the first integral is a function of the source coordinate, and the second a function of the target coordinate. At the true probability measures choose the Schr\"odinger potentials with $u(z_*)=0$. Their Schr\"odinger extensions to $\overline D$ are smooth. The kernel is smooth and strictly positive on the bounded domain, so the two integrals have positive lower bounds at this solution.

The map $\mathcal S:E^2\times\mathbb R\times\mathbb B^2\to E^2\times\mathbb R$ is locally $C^3$. To see this, the Banach algebra property makes exponentiation a smooth map on $E$. Pairing with a member of $\mathbb B$ is continuous. Differentiation of the smooth cost kernel in its free coordinate gives bounded operators into $H^s(D)$. The positive lower bounds persist in a sufficiently small $\mathbb B^2$ neighborhood, where the logarithms are smooth. Thus this argument includes signed perturbations in an open neighborhood of the true measures.

For the derivative with respect to the potentials, write
\[
 D_*(x,y)=e^{(u_*(x)+v_*(y)-c(x,y))/\varepsilon},\quad
 Af(x)=\nu_*(D_*(x,\cdot)f),\quad
 Bg(y)=\mu_*(D_*(\cdot,y)g).
\]
The Schr\"odinger equations give $A1=B1=1$, $\nu_*B=\mu_*$, and $\mu_*A=\nu_*$. The derivative is
\[
 \mathcal L(\dot u,\dot v,\dot\tau)
 =(\dot u+A\dot v,\ \dot v+B\dot u-\dot\tau,\ \dot u(z_*)).
 \tag{K.8}\label{eq:rare-inverse-operator}
\]
To solve $\mathcal L(\dot u,\dot v,\dot\tau)=(f,g,d)$, set
\[
 t=\mu_*f-\nu_*g,\qquad k=g-Bf+t.
\]
Then $\nu_*k=0$. The Markov operator $P=BA$ has invariant measure $\nu_*$. Strict positivity and boundedness of $D_*$ give $P(y,dz)\ge\eta\nu_*(dz)$ for some $\eta>0$. On the $\nu_*$-mean-zero subspace it therefore contracts in the uniform norm at a geometric rate. Its smooth kernel also gives
\[
 \|Ph\|_E\le C\|h\|_\infty,\qquad
 \|P^jk\|_E\le C(1-\eta)^{j-1}\|k\|_E\quad(j\ge1).
\]
It follows that $v_0=\sum_{j\ge0}P^jk$ converges in $E$ and solves $(I-P)v_0=k$. With
\[
 c_0=f(z_*)-(Av_0)(z_*)-d,\qquad
 \dot v=v_0+c_0,\quad \dot u=f-A\dot v,\quad\dot\tau=t,
\]
we obtain the unique solution of \eqref{eq:rare-inverse-operator}, with a bounded inverse.

The Banach implicit function theorem now gives a local $C^3$ extension of the normalized potentials. For mass-one probability measures, the first Schr\"odinger equation gives total coupling mass one and the second gives total mass $e^{\tau/\varepsilon}$, so $\tau=0$. On these measures,
$\Psi_\varepsilon(\mu,\nu)=\mu(u)+\nu(v)$. Continuous dual pairing and the smooth normalized tilt therefore give a local $C^3$ extension of $T(F,b)$ to $\mathbb B\times\mathbb R^p$. On a smaller neighborhood its first three derivatives are bounded. This proves the Taylor property used below, including for empirical measures on finite supports.

\subsection{A finite maximum and expansions of the fitted measures}\label{sec:rare-fit-expansion}
\paragraph{Existence of a finite maximum.} The normalized score $m^{-1}\partial_t\ell_{N,m}(t)$ is
\[
 U_{N,m}(t)=Q_m\frac{Q}{1+\lambda r_t}
 -P_N\frac{r_tQ}{1+\lambda r_t}.
 \tag{K.9}\label{eq:rare-score}
\]
At the true parameter, the two population terms cancel exactly. Let $M=E_1(QQ^\top)$, which is positive definite because $J\succ0$. The negative score Jacobian is
\[
 H_{N,m}(t)=\lambda Q_m\frac{r_tQQ^\top}{(1+\lambda r_t)^2}
 +P_N\frac{r_tQQ^\top}{(1+\lambda r_t)^2}.
 \tag{K.10}\label{eq:rare-hessian}
\]
At $t=\theta$, its expectation equals $E_1[QQ^\top/(1+\lambda r_\theta)]\to M$.

Choose a fixed ball $B(\theta,\delta)$. On that ball $Q$, $r_t$ and the first derivatives of the integrands in \eqref{eq:rare-score}--\eqref{eq:rare-hessian} are uniformly bounded for all sufficiently small $\lambda$. First choose $\delta$ so that continuity keeps the expected Hessian uniformly positive definite. Cover the ball by a finite net, apply bounded-variable concentration to each matrix entry at the net points, and extend to the ball with the uniform Lipschitz bound. This gives constants $c,C,c'>0$, independent of $N,m$, such that
\[
 \Pr\!\left\{\inf_{t\in B(\theta,\delta)}
 \lambda_{\min}(H_{N,m}(t))<c\right\}
 \le C(e^{-c'N}+e^{-c'm}).
\]
The same concentration argument applied to the centered score makes
$\|U_{N,m}(\theta)\|<c\delta/4$ outside an event with the same bound. On the intersection of these events, Taylor's formula on the boundary gives
\[
 m^{-1}\{\ell_{N,m}(\theta+h)-\ell_{N,m}(\theta)\}
 \le\|h\|\,\|U_{N,m}(\theta)\|-\tfrac12c\|h\|^2<0,
 \qquad \|h\|=\delta.
\]
The maximum over the closed ball is thus interior and has zero score. Concavity of the full likelihood makes it a global maximum. Local strict concavity makes it unique: a second global maximum would make the segment between the two maxima flat, including a segment in this ball. Denote this event by $\mathcal E_{N,m}$. We have established
\[
 \Pr(\mathcal E_{N,m}^c)\le C(e^{-c'N}+e^{-c'm}),\qquad
 \|\widehat\theta-\theta\|\le c^{-1}\|U_{N,m}(\theta)\|
 \quad\hbox{on }\mathcal E_{N,m}.
 \tag{K.11}\label{eq:rare-good-event}
\]
Since $U_{N,m}(\theta)=O_p(m^{-1/2}+N^{-1/2})$, it follows that $\widehat\theta-\theta=O_p(m^{-1/2})$ on this event. The NP fallback has no effect on the following probability limits because $\Pr(\mathcal E_{N,m}^c)\to0$.

\paragraph{Tilt expansion.} Center the source and target scores separately before taking the limit. Boundedness gives
\begin{align*}
 \sqrt m\,U_{N,m}(\theta)
 &=\sqrt m\,(Q_m-F_1)Q+O_p(\lambda)+O_p(\sqrt{m/N}),\\
 H_{N,m}(\widetilde\theta)&=M+o_p(1)
\end{align*}
for every point $\widetilde\theta$ on the segment between $\theta$ and $\widehat\theta$. The $O_p(\lambda)$ term is the centered target empirical process applied to $Q/(1+\lambda r_\theta)-Q$; the last term is the centered source process. Score linearization and the block inverse of $M$ therefore give
\[
 \sqrt m(\widehat\beta-\beta)=G_m+o_p(1),\qquad
 G_m=J^{-1}\frac1{\sqrt m}\sum_{j=1}^m q_c(Y_j)
 \ \Rightarrow\ G\sim N_p(0,J^{-1}).
 \tag{K.12}\label{eq:rare-beta-expansion}
\]
Indeed, solving $M(x_0,x^\top)^\top=(0,z^\top)^\top$ gives $x=J^{-1}z$.

\paragraph{Baseline expansion.} Because $\mathbb B$ is a Hilbert space and its point masses are uniformly bounded,
\[
 \|P_N-F\|_{\mathbb B}=O_p(N^{-1/2}),\qquad
 \|Q_m-F_1\|_{\mathbb B}=O_p(m^{-1/2}).
 \tag{K.13}\label{eq:rare-empirical-rates}
\]
The fitted baseline satisfies the exact identity
\[
 \widehat F-P_N=\lambda\left\{
 \frac{Q_m}{1+\lambda r_{\widehat\theta}}
 -\frac{r_{\widehat\theta}P_N}{1+\lambda r_{\widehat\theta}}\right\}.
 \tag{K.14}\label{eq:rare-pooled-identity}
\]
Writing $w_t=(1+\lambda r_t)^{-1}$, the expression in braces is exactly
\[
 w_{\widehat\theta}(Q_m-F_1)
 -r_{\widehat\theta}w_{\widehat\theta}(P_N-F)
 +(r_\theta-r_{\widehat\theta})w_{\widehat\theta}F.
\]
On the fixed parameter ball these multipliers have bounded operator norms on $\mathbb B$. The last term is bounded by $C\|\widehat\theta-\theta\|$ because its population measure vanishes at $\theta$ and depends smoothly on the finite-dimensional parameter. Thus
\begin{align*}
 \|\widehat F-P_N\|_{\mathbb B}
 &\le C\lambda\{\|Q_m-F_1\|_{\mathbb B}
 +\|P_N-F\|_{\mathbb B}+\|\widehat\theta-\theta\|\}
 \nonumber\\[-2pt]
 &=O_p(\sqrt m/N)=o_p(N^{-1/2}),
 \qquad \|\widehat F-F\|_{\mathbb B}=O_p(N^{-1/2}).
 \tag{K.15}\label{eq:rare-pooled-rate}
\end{align*}
The exact population cancellation in \eqref{eq:rare-pooled-identity} leaves the factor $\lambda$ multiplying the target-scale fluctuation. This makes the pooled baseline equivalent to the source empirical measure at the source square-root scale.

\subsection{Expansion and the three limiting regimes}\label{sec:rare-phase-proof}
Define
\[
 S_N=\frac1{\sqrt N}\sum_{i=1}^N\psi(X_i).
 \tag{K.16}\label{eq:rare-S}
\]
By the scalar and vector central limit theorems,
$(S_N,G_m)\Rightarrow(S,G)$, where $S\sim N(0,\sigma^2)$ and $G\sim N_p(0,J^{-1})$ are independent. Independence holds already for $S_N$ and $G_m$ because they use different samples.

Suppose first that $a=0$. Apply the joint Taylor expansion established in Section~\ref{sec:rare-smoothness} to $T(\widehat F,\widehat\beta)$. Write $H_N=\widehat F-F$ and $b_m=\widehat\beta-\beta$. The linear term is $H_N(\psi)$. The pure tilt second-order term is $b_m^\top\Gamma b_m/2$. Bounded derivatives and \eqref{eq:rare-beta-expansion}--\eqref{eq:rare-pooled-rate} give
\[
 \begin{aligned}
 H_N(\psi)&=(P_N-F)(\psi)+O_p(\sqrt m/N),\\
 D_{FF}^2T[H_N,H_N]&=O_p(N^{-1}),\\
 D_{Fb}^2T[H_N,b_m]&=O_p((Nm)^{-1/2}),\\
 R_3&=O_p\{(\|H_N\|_{\mathbb B}+\|b_m\|)^3\}
      =O_p(m^{-3/2}),\\
 \tfrac12 b_m^\top\Gamma b_m
 &=\frac1{2m}G_m^\top\Gamma G_m+o_p(m^{-1}).
 \end{aligned}
 \tag{K.17}\label{eq:rare-remainders}
\]
The baseline replacement error, the baseline and mixed quadratic terms, and the cubic remainder are all $o_p(N^{-1/2}+m^{-1})$ for every sequence $m/N\to0$. Specifically, $N^{-1}=o(N^{-1/2})$, $(Nm)^{-1/2}=o(N^{-1/2})$, $m^{-3/2}=o(m^{-1})$, and $(\sqrt m/N)/N^{-1/2}=\sqrt{m/N}\to0$. Hence
\[
 \widehat T-T(F,\beta)
 =\frac{S_N}{\sqrt N}+\frac{G_m^\top\Gamma G_m}{2m}
 +o_p(N^{-1/2}+m^{-1}).
 \tag{K.18}\label{eq:rare-center-expansion}
\]
If $\kappa=m/\sqrt N$, this yields
\[
 \begin{array}{ll}
 \kappa\to\infty:
 &\sqrt N(\widehat T-T)\Rightarrow S,\\[2pt]
 \kappa\to\kappa_0\in(0,\infty):
 &\sqrt N(\widehat T-T)\Rightarrow S+(2\kappa_0)^{-1}G^\top\Gamma G,\\[2pt]
 \kappa\to0:
 &m(\widehat T-T)\Rightarrow\tfrac12G^\top\Gamma G.
 \end{array}
 \tag{K.19}\label{eq:rare-regimes}
\]
The quadratic terms are uncentered. The assumptions $\sigma^2>0$ and $\Gamma\ne0$ ensure the respective source and quadratic limits are nondegenerate.

If $a\ne0$, the first-order tilt term is $a^\top b_m$. The source term has size $O_p(N^{-1/2})$, and the remaining Taylor terms have size $O_p(m^{-1}+N^{-1})$. Since $m/N\to0$, multiplication by $\sqrt m$ gives
\[
 \sqrt m(\widehat T-T(F,\beta))
 \Rightarrow N(0,a^\top J^{-1}a).
 \tag{K.20}\label{eq:rare-nonorthogonal}
\]
Thus first-order tilt sensitivity determines whether target noise enters at order $m^{-1/2}$ or $m^{-1}$.

\subsection{A continuous example with two correctly specified bases}\label{sec:rare-example}
Take $F=F_1=\operatorname{Unif}[-1,1]$, $\beta=0$, and $\varepsilon=1$. First use $q_1(x)=x$. Symmetry of the cost and the two marginals makes the centered potentials equal and even, with $\phi_0=\phi_1=\phi$. Therefore
\[
 a_{Q_1}=0,\qquad J_{Q_1}=1/3,\qquad
 \psi=2\phi,\qquad \sigma^2=4\operatorname{Var}_F(\phi).
 \tag{K.21}\label{eq:rare-symmetric-example}
\]
We next verify that $\sigma^2>0$ and $\Gamma\ne0$ analytically.

Use $A,B$ from Section~\ref{sec:rare-smoothness}, now restricted to the supports and viewed between $L^2(F_1)$ and $L^2(F)$. They are conditional-expectation operators for the same optimal coupling, so $B=A^*$. Consequently $P=BA=A^*A$ is self-adjoint and positive semidefinite on $L^2(F_1)$. For the minorization constant $\eta$, write $P=\eta\Pi+(1-\eta)R$, where $\Pi f=E_1f$ and $R$ is a Markov operator with invariant measure $F_1$. Jensen's inequality makes $R$ an $L^2(F_1)$ contraction. Hence $\|P\|\le1-\eta$ on the mean-zero subspace, which gives the required spectral gap.

For a tilt direction $b$, put $h=b^\top q_c$. Differentiating the Schr\"odinger equations with the source fixed gives
\[
 \dot u+A\dot v+\varepsilon Ah=0,\qquad
 \dot v+B\dot u=0.
\]
Modulo constants, $\dot v=\varepsilon(I-P)^{-1}Ph$. Differentiating $a=E_1(\phi_1q_c)$ then yields
\[
 \Gamma_{ij}=E_1(\phi_1q_{c,i}q_{c,j})
 +\varepsilon\langle q_{c,i},(I-P)^{-1}Pq_{c,j}\rangle_{L^2(F_1)}.
 \tag{K.22}\label{eq:rare-curvature}
\]
The second matrix is positive semidefinite because the spectral multiplier $t/(1-t)$ is nonnegative for $0\le t<1$. The first matrix can have either sign in a general model; the proposition only requires $\Gamma\ne0$.

In the uniform example the target potential satisfies
\[
 v''(y)=1-\operatorname{Var}(X\mid Y=y)>0.
\]
The conditional distribution has positive density on $[-1,1]$, hence variance strictly less than one. Thus $v$ is a strictly convex even function, up to its irrelevant additive constant. It is strictly increasing in $|y|$, and
\[
 E_F[\phi(X)X^2]=\operatorname{Cov}_F(v(X),X^2)>0.
 \tag{K.23}\label{eq:rare-positive-covariance}
\]
For example, strict positivity follows by writing the covariance as half the expectation of
$\{v(X)-v(X')\}(X^2-X'^2)$ for independent uniform $X,X'$. This product is positive whenever $|X|\ne|X'|$. The potential is nonconstant, so $\sigma^2>0$. Equations~\eqref{eq:rare-curvature}--\eqref{eq:rare-positive-covariance} give $\Gamma>0$ for $q_1=x$. All three regimes in \eqref{eq:rare-regimes} therefore apply.

Now enlarge the basis to $q_2(x)=(x,x^2)^\top$ at the same model point. Both $Q_1=(1,x)^\top$ and $Q_2=(1,x,x^2)^\top$ contain the true density ratio $r\equiv1$, and the covariance matrix of $q_2$ is positive definite. The additional tilt coordinate gives
\[
 a_{Q_2}=\bigl(0,\operatorname{Cov}_F(\phi(X),X^2)\bigr)^\top\ne0.
 \tag{K.24}\label{eq:rare-enlarged-basis}
\]
Equation~\eqref{eq:rare-nonorthogonal} therefore applies to $Q_2$, whereas \eqref{eq:rare-center-expansion} applies to $Q_1$. The larger correctly specified model permits a quadratic tilt that changes EOT to first order. The two rates concern two fixed statistical models at the same distribution pair.

The reviewer archive includes a deterministic quadrature check of the derivative and curvature in this example. From the extracted \texttt{reproducibility/} directory, run \texttt{python} \path{scripts/check_rare_target.py} \texttt{--out} \path{outputs/rare_target_quadrature.json}. The script uses the uniform design above and has no external data input. Compare its output with \path{results/rare_target/quadrature.json}.

\end{document}